\documentclass[twocolumn]{svjour3}

\usepackage[latin1]{inputenc}
\usepackage{times}

\usepackage{psfrag, amsmath, amssymb, amscd, amsfonts, latexsym, epsf, graphicx}
\usepackage{natbib}
\usepackage[switch]{lineno}

\def\bbbr{{\mathbb R}} 

\def\bbbs{{\mathbb S}}
\def\bbbc{{\mathbb C}}
\def\diag{{\operatorname{diag}}}
\def\norm{\scriptsize\mbox{norm}}
\def\affnorm{\scriptsize\mbox{affnorm}}
\def\normtiny{\tiny\mbox{norm}}
\def\affnormtiny{\tiny\mbox{affnorm}}

\journalname{arXiv preprint}

\begin{document}

\title{\bf Unified theory for joint covariance properties under geometric image
  transformations for spatio-temporal receptive fields according to
  the generalized Gaussian derivative model for visual receptive fields%
\thanks{The support from the Swedish Research Council 
              (contracts 2018-03586 and 2022-02969) is gratefully acknowledged. }}

\titlerunning{Unified theory for joint geometric covariance properties
  of spatio-temporal receptive fields}

\author{Tony Lindeberg}

\institute{Tony Lindeberg \at
                Computational Brain Science Lab,
              Division of Computational Science and Technology,
              KTH Royal Institute of Technology,
              SE-100 44 Stockholm, Sweden.
              \email{tony@kth.se}}

\date{Received: date / Accepted: date}

\maketitle

\begin{abstract}
\noindent
The influence of natural image transformations on receptive field
responses is crucial for modelling visual operations in computer
vision and biological vision. In this regard, covariance properties with
respect to geometric image transformations in the earliest layers of
the visual hierarchy are essential for expressing robust image
operations, and for formulating invariant visual operations at higher
levels.

This paper defines and proves a set of {\em joint\/} covariance properties
for spatio-temporal receptive fields in terms of spatio-temporal
derivative operators applied to spatio-tempo\-rally smoothed image data
under compositions of spatial scaling transformations, spatial affine
transformations, Gali\-lean transformations and temporal scaling
transformations.
%
Specifically, the derived relations show how the parameters of the
receptive fields need to be
transformed, in order to match the output from spatio-temporal receptive
fields under composed spatio-temporal image transformations.

For this purpose, we also fundamentally extend the notion of scale-normalized
derivatives to affine-normalized derivatives, that are computed
based on spatial smoothing with affine Gaussian kernels, and analyze
the covariance properties of the resulting affine-normalized
derivatives for the affine group as well as for important subgroups thereof.


We conclude with a geometric analysis, showing how the derived joint
covariance properties make it possible to relate or
match spatio-temporal receptive field responses, when observing,
possibly moving, local surface patches from different views, under
locally linearized perspective or projective transformations, as well
as when observing different instances of spatio-temporal events, that may
occur either faster or slower between different views of similar
spatio-temporal events. We do furthermore describe how the parameters
in the studied composed spatio-temporal image transformation models directly relate to
geometric entities in the image formation process and the 3-D scene
structure.

In these ways, this paper presents a {\em unified theory\/} for the interaction
between spatio-temporal receptive field responses and geometric image
transformations, with generic implications for both: (i)~designing computer vision
systems that are to compute image features and image descriptors, 
to be robust under the variabilities in spatio-temporal image
structures as caused by
geometric image transformations, and (ii)~understanding
fundamental geometric constraints for interpreting and constructing models
of biological vision.

\end{abstract}

\begin{figure*}[hbt]
    \begin{center}
    \begin{tabular}{cc}
      {\em\small (a) Non-covariant receptive fields\/}
      $\quad$ & $\quad$
     {\em\small (b) Covariant receptive fields\/} \\
      \includegraphics[width=0.35\textwidth]{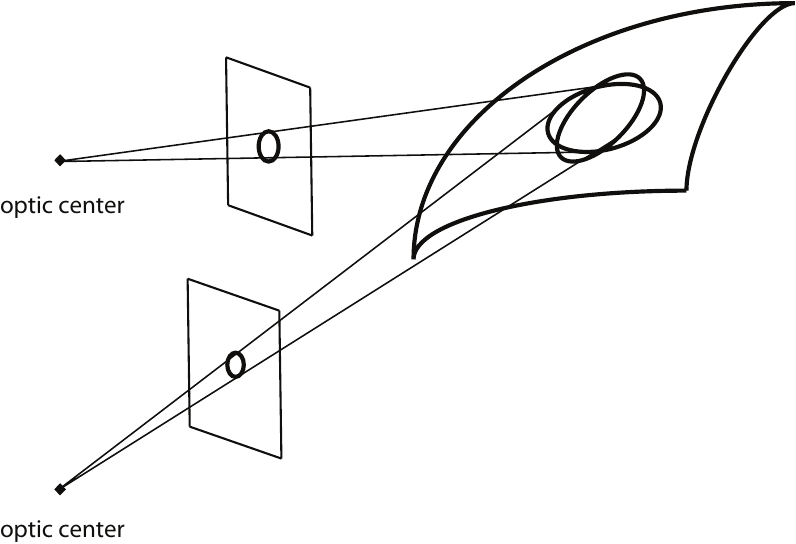}
      $\quad$ & $\quad$
      \includegraphics[width=0.35\textwidth]{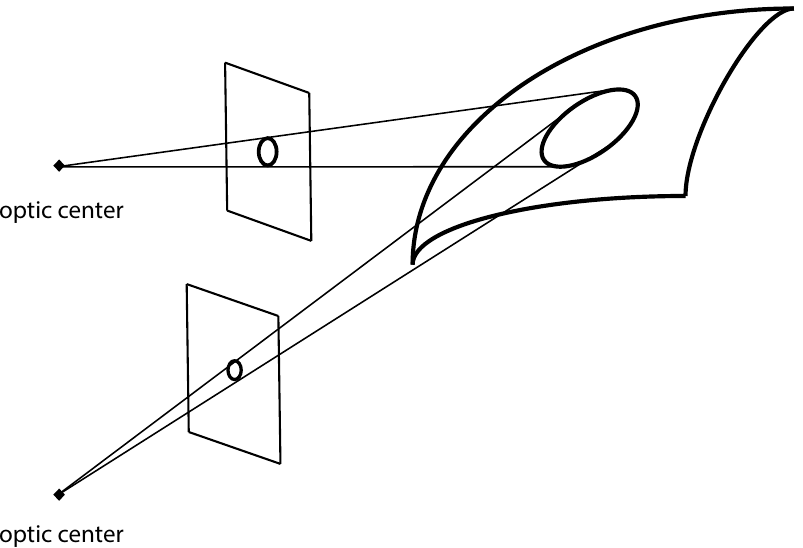} \\
    \end{tabular}
  \end{center}
  \caption{Illustration of the importance of covariance properties for
  the family of receptive fields, for a vision system aimed at
  analyzing the image data that originate from perspective projections
  of a 3-D world.
  (a)~Here, the left figure shows the effect of backprojecting
  non-covariant receptive fields to the tangent
  plane of that surface patch, based on rotationally symmetric
  spatial smoothing operations over the support regions of the
  receptive fields in the image domains, from two different
  image domains that observe local neighbourhoods of the
  same point in the 3-D world. As illustrated in the left figure, this
  will lead to a mismatch between the backprojected receptive fields,
  implying that if the difference in the receptive field measurements
  from the two image domains would be used for deriving cues to the
  3-D structure of the scene, then there would be a large source of
  error because of this mismatch between the receptive fields.
  (b)~If we, on the other hand, use covariant receptive fields, which
  within the family of locally linearized projective transformations
  between the two image domains will correspond to affine anisotropic
  spatial smoothing operations over the support regions of the
  receptive fields in the two image domains, then by appropriately matched
  choices of the parameters of the receptive fields, the backprojected
  receptive fields can be tuned to be equal, thus eliminating the
  mismatch error between the backprojected receptive fields, and, in
  turn, allowing for much more accurate determination of cues to the 3-D
  structure of the scene, compared to what can be achieved based on
  non-covariant receptive fields. Note that while this example illustrates a
  static configuration with purely spatial receptive fields, similar
  effects concerning the backprojected receptive fields arise also for
  a vision system, that observes a dynamic world based on joint spatio-temporal
  receptive fields, implying that covariance properties with respect to Galilean
  transformations as well as with respect to temporal scaling transformations
  constitute essential properties for a vision system, that aims at
  analyzing the image data originating from moving objects and
  spatio-temporal events in a dynamic world.
  (Figures adapted from Lindeberg  (\citeyear{Lin23-FrontCompNeuroSci})
  with permission (OpenAccess).)}
  \label{fig-ill-cov-rfs}
\end{figure*}

\section{Introduction}

When images, video sequences or video streams are acquired from the
real world, they are subject to natural image transformations, as
caused by variations in the positions, the relative orientations and
the motions between the objects in the world and the observer:
\begin{itemize}
\item
  Depending on the distance between the objects in the world and the
  observer, the perspective projections of objects onto the image surfaces may
  become smaller or larger, which to first-order of approximation
  can be modelled as local {\em spatial scaling transformations\/}.
\item
  Depending on the orientations of the surface normals of the objects
  in relation to the viewing direction, the image patterns may be
  compressed by different amounts in different directions (perspective
  foreshortening), which to first-order of approximation can be
  modelled as local {\em spatial affine transformations\/}.
\item
  Depending on how the objects in the world move relative to the
  (possibly time-dependent) viewing direction, the image patterns of
  objects may move in the image plane, which to first-order of
  approximation can be modelled as local {\em Galilean transformations\/}.
\item
  Depending on how fast the perspective projections of objects move in the image plane, or how fast
  spatio-temporal actions occur, the time-line along the temporal
  dimension may be compressed or expanded, which can be modelled as
  {\em temporal scaling transformations\/}.
\end{itemize}
These types of geometric image transformations will have a profound
effect on the spatio-temporal receptive fields, that register and
process the image information at the earliest stages in the visual
hierarchy, in that the output from the receptive fields to particular
image patterns may be strongly dependent on the imaging conditions.
Specifically, if the interaction effects between the geometric image
transformations and the receptive fields are not properly taken into
account, then the robustness of the visual modules can be strongly
affected in a negative manner. If, on the other hand, the interaction effects between the
natural image transformations are properly taken into account, then
the robustness of the visual measurements may be substantially
improved.

A particular way of handling the interaction effects between the
geometric image transformations and the receptive fields, is by
requiring the family of receptive fields to be {\em covariant\/}
under the relevant classes of image transformations
(Lindeberg \citeyear{Lin13-BICY,Lin23-FrontCompNeuroSci}).
Covariance
in this context means that the geometric image transformations essentially commute
with the image operations induced by the receptive fields, and do in
this way provide a way to propagate well-defined relationships between
the geometric image transformations and the receptive
fields. Specifically, covariance properties of the receptive fields at
lower levels in the visual hierarchy make it possible to define
invariant image measurements at higher levels in the visual hierarchy
(Lindeberg \citeyear{Lin13-PONE,Lin21-Heliyon},
Poggio and Anselmi \citeyear{PogAns16-book}).

The subject of this paper is to describe and derive a set of joint covariance
properties of receptive fields according to a specific model for
spatio-temporal receptive fields, in terms of the 
generalized Gaussian derivative model for visual receptive fields
(to be detailed below), under
joint combinations of spatial scaling transformations, spatial affine
transformations, Galilean transformations and temporal scaling
transformations; see Figure~\ref{fig-ill-cov-rfs} for a visualized motivation
regarding the importance of covariance under geometric image
transformations for visual receptive fields.

Then, we will show with a geometric analysis how
these derived joint geometric covariance properties make it possible
to, to first order of approximation, perfectly match the
spatio-temporal receptive field responses between different views of
the same, possibly moving, local surface patch, in relation to a
visual observer. In these ways, the resulting
joint covariance properties make it possible for a vision system, biological or
artificial, to perform more accurate inference to cues of the 3-D
environment, compared to a vision system that does not obey such geometric
covariance properties. Such a possibility, for geometrically accurate
inference to the 3-D structure and motion in the environment,
may, in turn, constitute an essential
desirable property of a vision system for biological agent, who relies
critically on a very well-developed vision system for its survival.

For the purpose of the theoretical analysis to be performed,
we will build upon the regular Gaussian derivative
model for visual receptive fields, proposed by
Koenderink and van Doorn (\citeyear{Koe84,KoeDoo87-BC,KoeDoo92-PAMI}),
which has been used for modelling
biological receptive fields by Young (\citeyear{You87-SV}) as well
used as a component in more developed models of biological vision by
Lowe (\citeyear{Low00-BIO}),
May and Georgeson (\citeyear{MayGeo05-VisRes}),
Hesse and Georgeson (\citeyear{HesGeo05-VisRes}),
Georgeson  {\em et al.\/}\ (\citeyear{GeoMayFreHes07-JVis}),
Wallis and Georgeson (\citeyear{WalGeo09-VisRes}),
Hansen and Neumann (\citeyear{HanNeu09-JVis}),
Wang and Spratling (\citeyear{WanSpra16-CognComp}) and
Pei {\em et al.\/}\ (\citeyear{PeiGaoHaoQiaAi16-NeurRegen}).

In this work, we will, however, consider a more developed generalized Gaussian derivative
model for visual receptive fields, 
extended with a variability over affine image
transformations (Lindeberg and G{\aa}rding \citeyear{LG96-IVC})
as well as furthermore extended
from being applied over purely spatial image domain to being applied
over a joint spatio-temporal image domain
(Lindeberg \citeyear{Lin10-JMIV,Lin16-JMIV,Lin21-Heliyon}).
Compared to the earlier spatio-temporal
modelling work by Young {\em et al.\/} (\citeyear{YouLesMey01-SV,YouLes01-SV}),
we will here specifically consider a more geometric way of parameterizing
the degrees of freedom over the
joint spatio-temporal domain, as will be further described in
Section~\ref{sec-gen-gauss-der-model}.

While Gabor filters have also been commonly used for modelling spatial
receptive fields by Marcelja (\citeyear{Mar80-JOSA}),
Jones and Palmer (\citeyear{JonPal87a,JonPal87b}),
Ringach (\citeyear{Rin01-JNeuroPhys,Rin04-JPhys}),
Serre {\em et al.\/} (\citeyear{SerWolBilRiePog07-PAMI}),
De and Horwitz (\citeyear{DeHor21-JNPhys}) and others, the potential
applicability of Gabor filters for modelling joint spatio-temporal
receptive fields has, however, not been as extensively explored. For this
reason, we will restrict ourselves to modelling visual receptive
fields over the joint spatio-temporal domain in terms of the
generalized Gaussian derivative theory for visual receptive fields in
the following treatment.

Although we will in this treatment be mainly concerned with studying
the properties of the generalized Gaussian derivative model for visual
receptive fields, which has been proposed as a theoretically
principled model for the
simple cells in the primary visual cortex in
Lindeberg (\citeyear{Lin21-Heliyon}), the implications of the
combined analysis of spatio-temporal receptive field responses and
geometric image transformations should also have more general
applications in the area of computer vision.

For example, recent work has explored scale-covariant or
scale-equivariant architectures for deep learning, which have
the ability to properly handle the influence of spatial scaling
transformations on the receptive field responses,
see Worrall and Welling (\citeyear{WorWel19-NeuroIPS}),
Bekkers (\citeyear{Bek20-ICLR}),
Sosnovik {\em et al.\/}
(\citeyear{SosSzmSme20-ICLR,SosMosSme21-BMVC}),
Lindeberg (\citeyear{Lin20-JMIV,Lin22-JMIV}),
Jansson and Lindeberg (\citeyear{JanLin21-ICPR,JanLin22-JMIV}),
Zhu {\em et al.\/} (\citeyear{ZhuQiuCalSapChe22-JMLR}),
Penaud {\em et al.\/} (\citeyear{PenVelAng22-ICIP}),
Sangalli {\em et al.\/} (\citeyear{SanBluVelAng22-BMVC}),
Zhan {\em et al.\/} (\citeyear{ZhaSunLi22-ICCRE}),
Yang  {\em et al.\/} (\citeyear{YanDasMah23-arXiv}),
Wimmer {\em et al.\/}  (\citeyear{WimGolDanMaiCre23-arXiv}),
Barisin  {\em et al.\/}
(\citeyear{BarSchRed24-JMIV,BarAngSchRed24-SIIMS}),
and Perzanowski and Lindeberg (\citeyear{PerLin25-JMIV}).

Based on the theoretical analysis presented in this paper, we propose
that it ought to
be possible to extend similar ideas to more general approaches for
geometric deep learning
(see Bronstein {\em et al.\/} (\citeyear{BroBruCohVel21-arXiv})
and Gerken {\em et al.\/} (\citeyear{GerAroCarLinOhlPetPer23-AIRev}))
that are covariant under wider classes of geometric image transformations.

 \subsection{Contributions and novelty}

The main new contributions in this paper concern:
\begin{itemize}
\item
  the formulation of a {\em joint\/} covariance property of the spatio-temporal
  smoothing transformation under the composition of
  (i)~a spatial scaling transformation,
  (ii)~a spatial affine transformation,
  (iii)~a Galilean transformation and
  (iv)~a temporal scaling transformation
  in Section~\ref{eq-transf-prop-spat-temp-smooth} and
\item
  the formulation of {\em joint\/} transformation properties of
  both regular and scale-normalized spatio-temporal derivative
  responses
  under corresponding
  compositions of the same set of primitive geometric image transformations
  in
  Sections~\ref{eq-transf-prop-spat-temp-ders}--\ref{sec-transf-prop-sc-norm-spat-temp-ders},
  as well as
\item
  the explicit geometric interpretations of the above joint covariance
  and transformation properties between multi-view image observations
  of dynamic scenes 
  in Sections~\ref{sec-geom-interpret-mono}
  and~\ref{sec-geom-interpret-pair}, which extend previous studies of
  multi-view geometry for static scenes to multi-view geometry for
  scenes with relative motions between the objects or events in the environment
  and the observer, with
\item
  the corresponding explicit expressions for the resulting
  transformation properties for the spatio-temporal smoothing operations
  and the spatio-temporal receptive field responses between different
  pairwise views in
  Sections~\ref{sec-joint-cov-prop-spat-temp-smooth}
  and~\ref{sec-joint-cov-prop-spat-temp-spat-temp-ders}.
\end{itemize}
Fundamentally, to be able to express the above covariance and
transformation properties for scale-normalized derivatives under
spatial affine transformations, we also:
\begin{itemize}
\item
  define a set of 
  new notions of scale-normalized derivative operators
  for spatial derivative operators, that are to be computed based on
  an anisotropic
  spatial affine scale-space representation,
  as well as analyze the covariance properties of these
  affine-extended scale-normalized derivatives, for either the
  full affine group or important subgroups of the affine group, in
  Sections~\ref{sec-aff-sc-norm-dir-ders}--\ref{sec-cov-prop-sc-norm-aff-hess-mat}.
\end{itemize}
In this way, we generalize the previous notion of scale-normal\-ized spatial
derivative operators over a more regular isotropic spatial scale-space
representation, as reviewed in Sections~\ref{sec-sc-norm-ders-isotropic}--\ref{sec-scale-cov-isotropic-spat-sc-norm-ders}.

Additionally, to put the presented theoretical results into a wider
perspective of overall theoretical properties regarding an either artificial
or biological vision system, we 
\begin{itemize}
\item
  interpret the presented theoretical results regarding joint
  covariance properties in terms of relationships between the
  variabilities of image structures under the studied classes of
  natural image transformations in relation to the degrees of freedom,
  that are spanned by the spatio-temporal receptive model, in
  Section~\ref{sec-interpret-geom-biol}, and
\item
  relate the parameters in the studied spatio-temporal deformation
  models of image patterns, resulting from the joint image
  transformations to geometric properties of local surface patterns in
  the environment, in Section~\ref{sec-cues-3d-structure}.
\end{itemize}
To be able to present these results in a reasonably self-contained manner for a
reader, for which essential components of the theoretical background
may not be already fully known, we:
\begin{itemize}
\item
  review the underlying spatio-temporal receptive field model in
  Section~\ref{sec-gen-gauss-der-model},
\item
  review the notion of scale-normalized temporal derivatives in
  Section~\ref{sec-sc-norm-temp-ders} with its resulting
  temporal scale covariance property in
  Section~\ref{sec-cov-prop-temp-sc-ders},
  although in a generalized form, with the
    previous use of either non-causal 1-D temporal Gaussian kernels or
    the time-causal limit kernel replaced by a family of more general
    scale-covariant temporal kernels, and
\item
  review the notion of scale-normalized velocity-adapted temporal derivative operators
  in Section~\ref{sec-sc-norm-vel-adapt-temp-ders}, with its associated
  (although not previously explicitly formulated) covariance properties under
  joint spatial and temporal scaling transformations,
  in Sections~\ref{sec-cov-prop-sc-norm-vel-adapt-temp-ders}.
\end{itemize}
Furthermore, to make the motivation clear for the in-depth treatment
of joint covariance and transformation properties in
Section~\ref{sec-joint-cov-props}, for a reader who may not be
already familiar with the material in
(Lindeberg \citeyear{Lin23-FrontCompNeuroSci}), we
\begin{itemize}
\item
  review the covariance properties under the four different types
  of {\em individual\/} geometric image transformations in
  Section~\ref{sec-transf-props-spat-temp-scsp-individ}, while here also extended
  to more explicit transformation properties of spatio-temporal
  derivative operators in
  Section~\ref{sec-transf-props-spat-temp-ders-individ},
  as based on the treatment of the different types of regular
  (not scale-normalized) spatio-temporal derivative operators
  in Section~\ref{sec-sc-norm-spat-temp-ders}.
\end{itemize}
In this context, one more added value of the treatment of the {\em joint\/}
transformation properties in Section~\ref{sec-joint-cov-props},
beyond the fundamental extension from four different types of
individual covariance properties to joint covariance properties,
as well as beyond the also fundamental extensions to algebraically
much simpler covariance and transformation properties in terms
of scale-normalized derivatives, as also performed in
Section~\ref{sec-joint-cov-props},
is that the proof for the joint transformation property
in Section~\ref{eq-transf-prop-spat-temp-smooth} also constitutes a
general proof for
each one of the individual transformation
properties in Section~\ref{sec-transf-props-spat-temp-scsp-individ}.
Such proofs were not provided in the previous treatment in
(Lindeberg \citeyear{Lin23-FrontCompNeuroSci}),
because of the there implied complexity of the need for
then providing four individual proofs,
according to the previous individual treatments for each one of
the different types of geometric covariance
properties. With the unified treatment of the joint covariance
properties developed in this paper, the joint covariance properties
can here instead be proved in an all encompassing single proof.

Beyond the purely review-oriented Section~\ref{sec-gen-gauss-der-model}, the
purpose of the underpinning theory-oriented
Sections~\ref{sec-sc-norm-spat-temp-ders}
and~\ref{sec-individ-cov-props} is thus to provide the conceptual and
theoretical foundations for formulating and deriving the main results regarding
joint covariance and transformation properties in
Sections~\ref{sec-joint-cov-props}--\ref{sec-interpret-geom-biol}.

In summary, the overall aim of this paper is to present a {\em unified theory\/} for
the interaction between spatio-temporal receptive field responses and
geometric image transformations, which comprises several previous
contributions in the area as different special cases, while here also
providing substantial generalizations to: (i)~joint geometric image
transformations, (ii)~receptive fields in terms of richer and more
explicit sets of both regular and scale-normalized spatio-temp\-oral
derivatives, as well as to (iii)~covariant spatial derivatives defined
from an anisotropic affine scale-space representation, to make the
affine-extended notion of scale-normal\-ized spatial derivatives essentially equal under
the influence of spatial affine transformations.

Furthermore, by the presented (iv)~geometric interpretations of the derived theoretical
results, we show how (v)~essential components of early visual
perception can be
expressed in terms of the studied class of composed
geometric image transformations between multiple views,
for either a monocular or a binocular observer, that observes a dynamic
environment from possibly different viewing directions.

\begin{figure*}[hbtp]
  \begin{center}
    \begin{tabular}{ccccc}
      $\sigma_x = 1$ & $\sigma_x = \sqrt{2}$ & $\sigma_x = 2$ & $\sigma_x = 2\sqrt{2}$ & $\sigma_x = 4$ \\
      \includegraphics[width=0.17\textwidth]{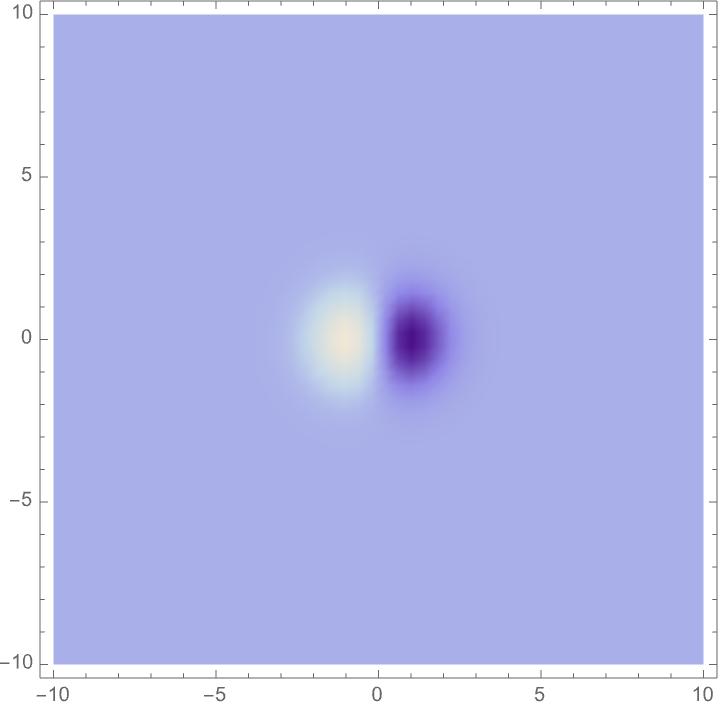}
      & \includegraphics[width=0.17\textwidth]{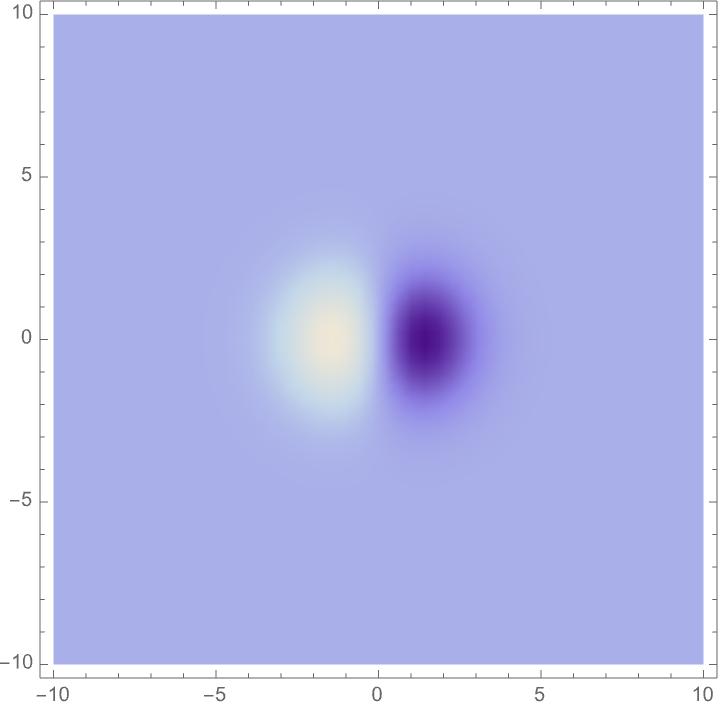}
      & \includegraphics[width=0.17\textwidth]{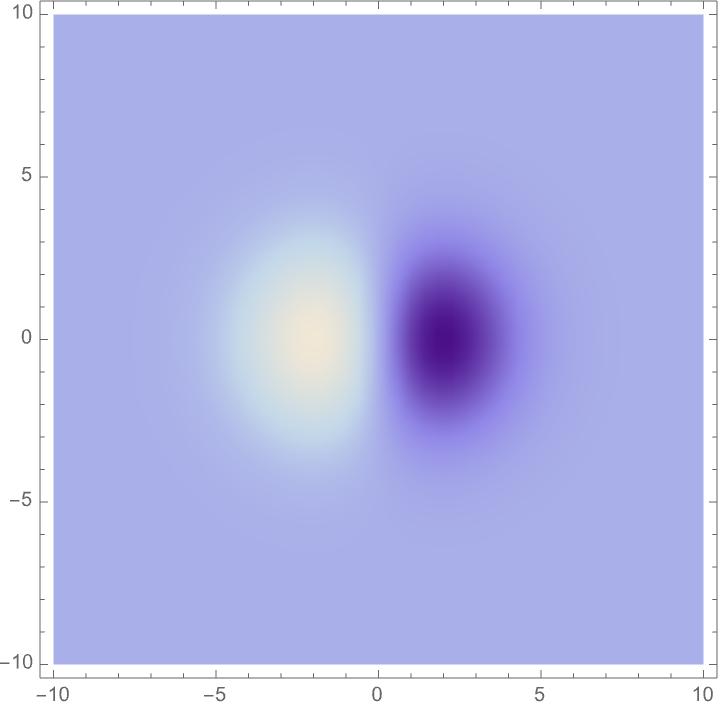}
      & \includegraphics[width=0.17\textwidth]{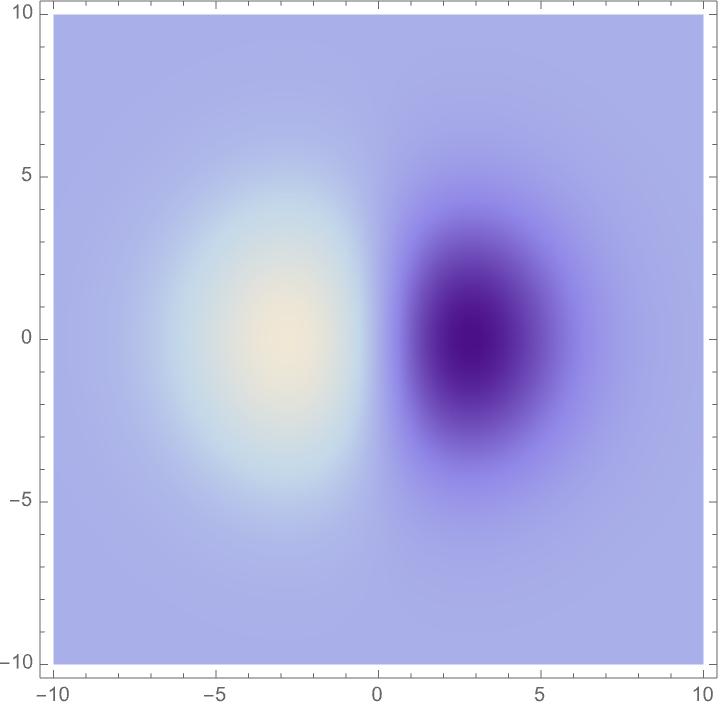}
      & \includegraphics[width=0.17\textwidth]{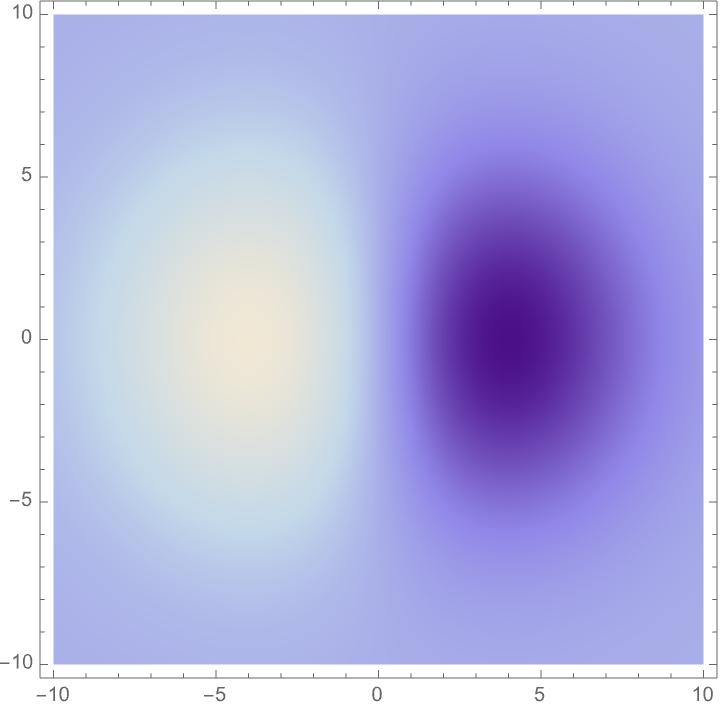}
     \end{tabular}
   \end{center}
   \caption{Variability over scale of the first-order directional
     derivative of the Gaussian kernel
     $T_{\varphi}(x;\; s, \Sigma) = \partial_{\varphi} (g(x;\; s, \Sigma))$
   in the horizontal direction $\varphi = 0$ for an isotropic
   spatial covariance matrix with $\Sigma = I$, and
   corresponding to the variability induced by
     varying the distance between the observed object, while because of
     the requirement of {\em spatial scale covariance\/} requiring the
     sizes of the backprojected receptive fields to be the same in the
     tangent plane of the surface under the resulting spatial scaling
     transformations. The scale parameter $\sigma_x$ used here, in units
     of the standard deviation of the Gaussian kernel, is equal to the
     square root of the variance-based scale parameter $s = \sigma_x^2$
     used elsewhere in this paper.
     (Horizontal axes: spatial coordinate $x_1 \in [-10, 10]$.
     Vertical axes: spatial coordinate $x_2 \in [-10, 10]$.)}
   \label{fig-1spatders-scale-var}
 \end{figure*}

\subsection{Structure of this article}

This paper is organized as follows:
Section~\ref{sec-gen-gauss-der-model} begins by describing the model
for spatio-temporal receptive fields, that we will build upon, in
terms of an underlying joint spatio-temporal smoothing operation
followed by the computation of spatio-temporal derivatives for
different orders of spatial and temporal differentiation. We do also
give a brief summary of how these
spatio-temporal receptive field models can be used for modelling linear
receptive fields in the retina, the lateral geniculate nucleus (LGN)
and the primary visual cortex (V1).

Section~\ref{sec-sc-norm-spat-temp-ders} then describe
the notions of scale-normalized spatial and temporal derivative
operators, with their associated covariance properties under
(individual) spatial
and scaling transformations, which constitute an important concept to
use, when to match receptive field responses that have been computed
for different values of the scale parameters of the receptive fields.
Specifically, we formulate a new notion of affine
scale-normalized directional derivatives, to be applied in connection
with anisotropic affine Gaussian smoothing kernels, and show that this
concept leads to provable covariance properties, for two important
subgroups of the group of more general spatial affine transformations.
More generally, we do also formulate new notions of a scale-normal\-ized
affine gradient operator and a scale-normalized affine Hessian
operator, and show that these concepts, up to possibly unknown low-dimensional
perturbation operators applied to these entities, lead to full affine covariance.

Section~\ref{sec-individ-cov-props} gives an overview of how the
studied model for joint spatio-temporal receptive fields obeys specific (individual)
covariance properties under either spatial scaling transformations, spatial affine
transformations, Galilean transformations or temporal scaling
transformations.
Section~\ref{sec-joint-cov-props} defines the class of joint
compositions of those spatio-temporal image transformations
that we will consider, and does then develop explicit proofs for how
both the underlying spatio-temporal smoothing operation as well as the spatial and
temporal derivative operators, in the spatio-temporal receptive field
model that we study, are transformed under this class of composed
spatio-temporal image transformations. 

Section~\ref{sec-geom-interpret} then gives a geometric interpretation
of the studied class of composed spatio-temporal image transformations,
that we study the covariance properties for, in terms of the scaled orthographic
projection from the tangent plane of a local surface patch, 
complemented by a local translation motion model, to account for
relative motions between the surface patch and the observer, as well
as a temporal scaling transformation, to account for spatio-temporal
events that may occur either faster or slower relative to a reference
view. We do also present extensions, showing how a slight modification
of the composed spatio-temporal transformation model makes it possible
to represent first-order linearized approximations of the projective
transformations between pairwise views of the same local surface
patch.

Section~\ref{sec-cov-props-pairwise-views} then states explicit
covariance properties for the underlying spatio-temporal smoothing
transformation, as well as the underlying spatial and temporal
derivative operators, in the composed model for spatio-temporal
receptive fields, for locally linearized transformations between pairwise
views of the same local surface patch.

In Section~\ref{sec-interpret-geom-biol}, we complement the geometric
interpretation of the model, by describing how the degrees of
freedom in the parameters
in the spatio-temporal receptive field model studied in this treatment
span a similar variability, as the degrees of freedom in the locally
linearized scaled orthographic projection model complemented with a
Galilean motion component, to account for possibly unknown relative
motions between the observed object and the observer, as well a
temporal scaling transformation, to account for similarly looking
spatio-temporal events that may occur either faster or slower relative
to a previous view of a similar spatio-temporal event.
Then, we use this connection for interpreting the functional properties of
the receptive fields of simple cells in the primary visual cortex (V1), to provide
complementary theoretical support for a previously formulated
working hypothesis, that the receptive fields in the
primary visual cortex can be regarded as very well adapted to handling the
variability of image structures caused by observing a dynamic 3-D environment.

Section~\ref{sec-cues-3d-structure} then outlines how the parameters in
the studied spatio-temporal image transformation models can be
interpreted as constituting direct
cues to the 3-D structure of the environment, provided that the
parameters in this image deformation models can be computed
with sufficient accuracy, based on combinations of receptive field responses.

Finally, Section~\ref{sec-summ-disc} gives a summary and conclusions
regarding some of the main results, as well as an outlook concerning
more general applications of the presented theoretical results to computer vision
and biological vision.

\section{The generalized Gaussian derivative model for spatio-temporal
  receptive fields}
\label{sec-gen-gauss-der-model}

Given spatio-temporal image data of the form $f(x, t)$ with
$f \colon \bbbr^2 \times \bbbr \rightarrow \bbbr$ over the
spatial coordinates $x = (x_1, x_2)^T \in \bbbr^2$ and time $t \in \bbbr$, in
Lindeberg (\citeyear{Lin10-JMIV,Lin16-JMIV}) a principled model
$T \colon \bbbr^2 \times \bbbr \times \bbbr_+ \times \bbbs_+^2 \times \bbbr_+
\times \bbbr^2 \rightarrow \bbbr$
for spatio-temporal receptive fields
is derived and applied of the form
(here, however, with slightly modified notation)
\begin{equation}
  \label{eq-spat-temp-RF-model}
  T(x, t;\; s, \Sigma, \tau, v) 
  = g(x - v \, t;\; s, \Sigma) \, h(t;\; \tau),
\end{equation}
where
\begin{itemize}
\item
  $s \in \bbbr_+$ denotes a spatial scale parameter, corresponding to the spatial variance of a
  non-negative spatial smoothing kernel,
\item
  $\Sigma \in \bbbs_+^2$ denotes a symmetric and positive definite $2 \times 2$
  spatial covariance matrix,
  that describes the spatial shape of the spatial smoothing kernel,
\item
  $\tau \in \bbbr_+$ denotes a temporal scale parameter, corresponding to the
  temporal variance of a non-negative temporal smoothing kernel,
\item
  $v = (v_1, v_2)^T \in \bbbr^2$ denotes an image velocity vector,
\item
  $g \colon \bbbr^2 \times \bbbr_+ \times \bbbs_+^2 \rightarrow \bbbr$
  denotes a 2-D affine Gaussian kernel of the form
  \begin{equation}
     \label{eq-gauss-fcn-2D}
     g(x;\; s, \Sigma)
     = \frac{1}{2 \pi \, s \sqrt{\det \Sigma}} \, e^{-x^T  \Sigma^{-1} x/2 s},
  \end{equation}
\item
  $h \colon \bbbr \times \bbbr_+ \rightarrow \bbbr$
  denotes a temporal smoothing kernel, that for any
  temporal scaling factor $S_t \in \bbbr_+$ obeys the {\em temporal scale covariance property\/}
  \begin{equation}
    \label{eq-temp-sc-cov-temp-kernel}
     h(t';\; \tau') = \frac{1}{S_t} \, h(t;\; \tau)
   \end{equation}
   for $t' = S_t \, t$ and $\tau' = S_t^2 \, \tau$.
 \end{itemize}

 \subsection{Temporal smoothing kernels}
 
Based on the treatments in Lindeberg
(\citeyear{Lin10-JMIV,Lin16-JMIV}), the choice of the temporal
smoothing operation as the convolution with a 1-D temporal Gaussian
kernel $g_{1D} \colon \bbbr \times \bbbr_+ \rightarrow \bbbr$ with
\begin{equation}
  \label{eq-def-temp-gauss-kern}
    h(t;\; \tau) = g_{1D}(t;\; \tau) = \frac{1}{\sqrt{2 \pi \tau}} \, e^{-t^2/2 \tau}
\end{equation}
stands out as a canonical choice over a non-causal temporal domain,
where the relative future in relation to any time moment can be
accessed, whereas the choice of the temporal kernel as the
time-causal limit kernel (Lindeberg \citeyear{Lin23-BICY}) 
$\Psi \colon \bbbr \times \bbbr_+ \times \bbbr_{>1} \rightarrow \bbbr$ with
\begin{equation}
  \label{eq-time-caus-limit-kern}
    h(t;\; \tau) = \Psi(t;\; \tau, c),
\end{equation}
defined by having a Fourier transform
$\hat{\Psi} \colon \bbbr \times \bbbr_+ \times \bbbr_{>1} \rightarrow \bbbc$
of the form
\begin{equation}
  \label{eq-FT-comp-kern-log-distr-limit}
     \hat{\Psi}(\omega;\; \tau, c) 
     = \prod_{k=1}^{\infty} \frac{1}{1 + i \, c^{-k} \sqrt{c^2-1} \sqrt{\tau} \, \omega},
\end{equation}
and corresponding to an infinite number of truncated exponential
kernels, with specially chosen time constants to obtain temporal scale covariance,
stands out as a canonical choice over a time-causal temporal domain,
where the future cannot be accessed.
The distribution parameter $c > 1$ of this time-causal limit kernel is for practical purposes often
chosen as $c = \sqrt{2}$ or $c = 2$.

\begin{figure}[hbtp]
  \begin{center}
     \includegraphics[width=0.45\textwidth]{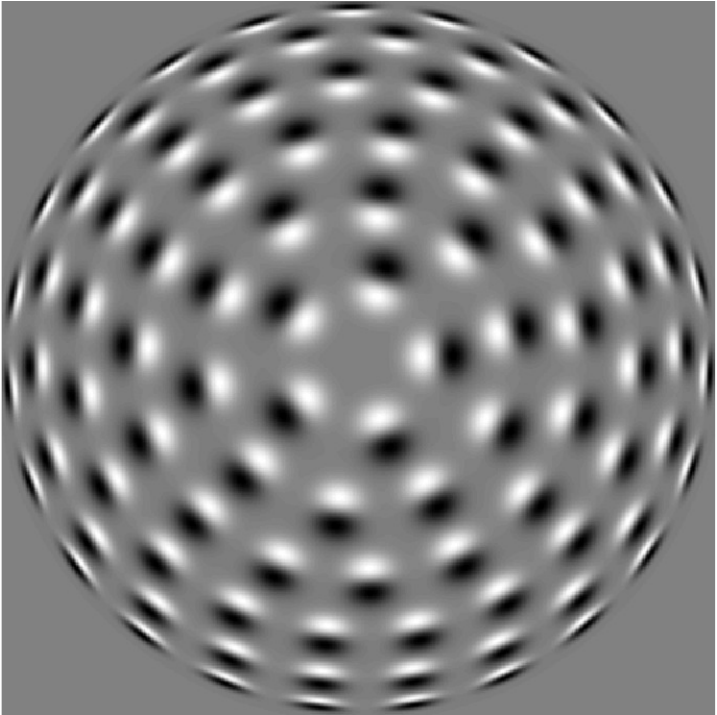} 
   \end{center}
   \caption{Variability of first-order directional spatial derivatives of
     Gaussian kernels
     $T_{\varphi}(x;\; s, \Sigma) = \partial_{\varphi} (g(x;\; s, \Sigma))$
     over a purely spatial domain, here shown in terms of
     a uniform distribution on a
     hemisphere, for different values of the orientation angle
     $\varphi$, the spatial scale parameter $s$ and the spatial
     covariance matrices $\Sigma$,
     and in this way simulating the variability of spatial receptive
     field shapes, that will be the result by interpreting the 
     {\em purely spatial affine covariance property\/}, such that the underlying
     spatial smoothing kernels are required to be rotationally symmetric in the
     tangent plane of a surface patch, while varying the slant and the
     tilt angles of the surface patch over all the angles on the visible
     hemisphere. Similar variabilities will result from directional
     derivatives of higher order. In this figure, the spatial scale
     parameters of the receptive fields have been normalized, such that
   the maximum eigenvalue of the spatial covariance matrix $\Sigma$ is
 the same for all the receptive fields. (Horizontal and vertical
 axes: the spatial coordinates $x_1$ and $x_2$, for multiple spatial receptive
 fields shown within the same frame.)}
   \label{fig-1dir-gaussder}
\end{figure}

\begin{figure*}[hbtp]
  \begin{center}
    \begin{tabular}{ccccc}
      $v = -1$ & $v = -1/2$ & $v = 0$ & $v = 1/2$ & $v = 1$ \\
      \includegraphics[width=0.17\textwidth]{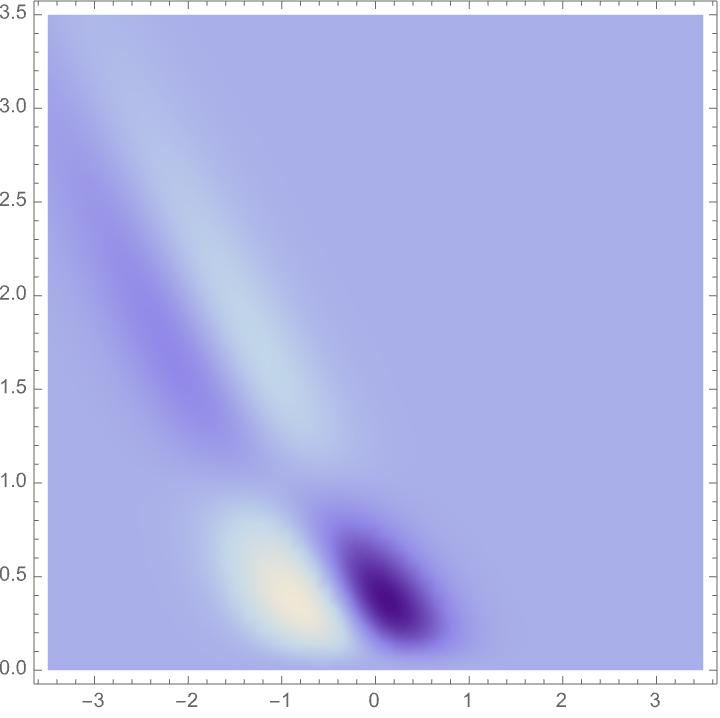}
      & \includegraphics[width=0.17\textwidth]{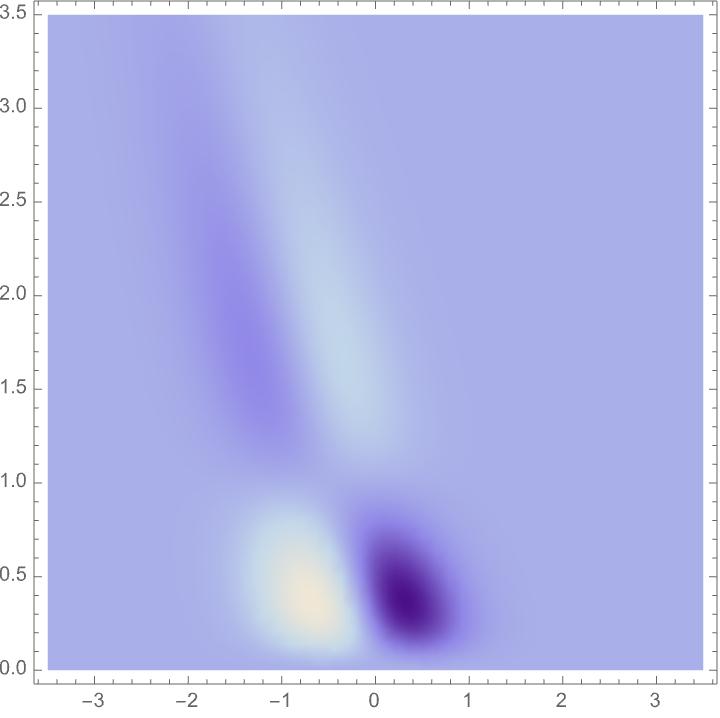}
      & \includegraphics[width=0.17\textwidth]{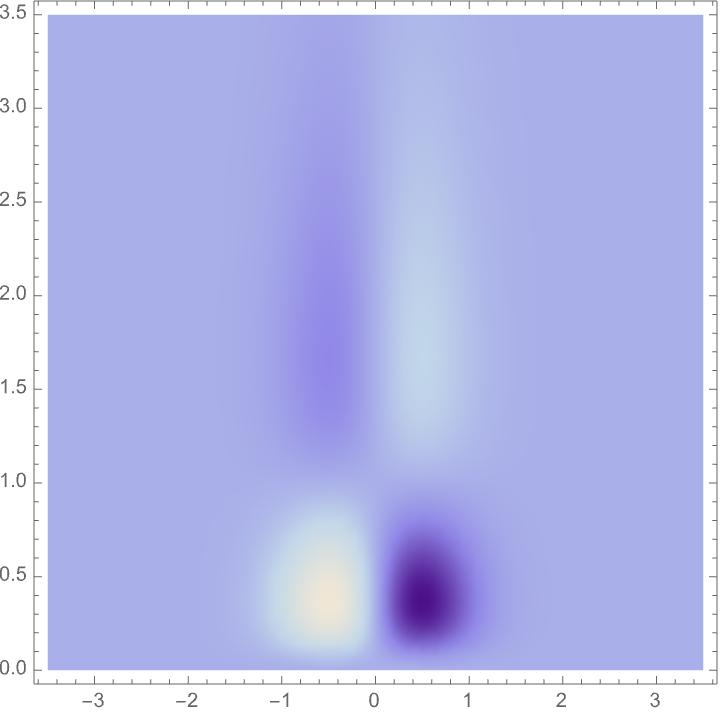}
      & \includegraphics[width=0.17\textwidth]{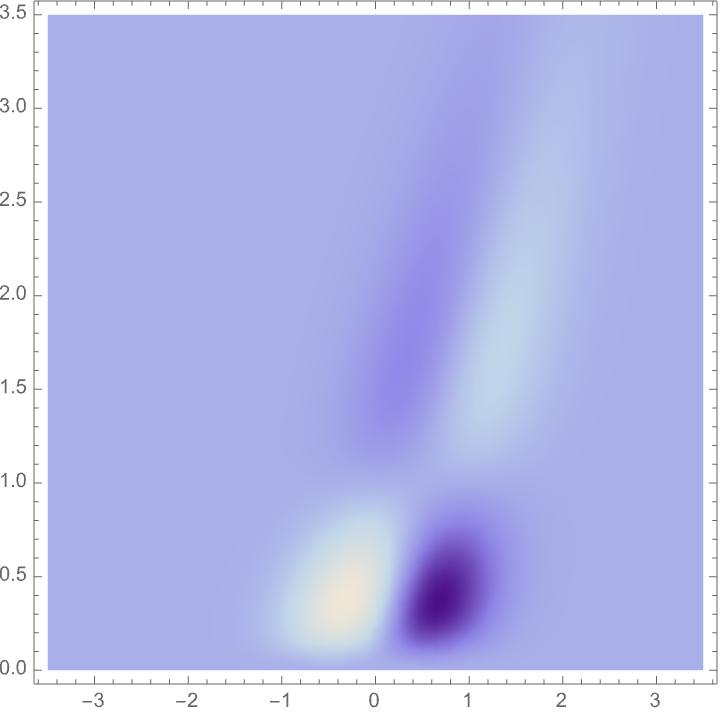}
      & \includegraphics[width=0.17\textwidth]{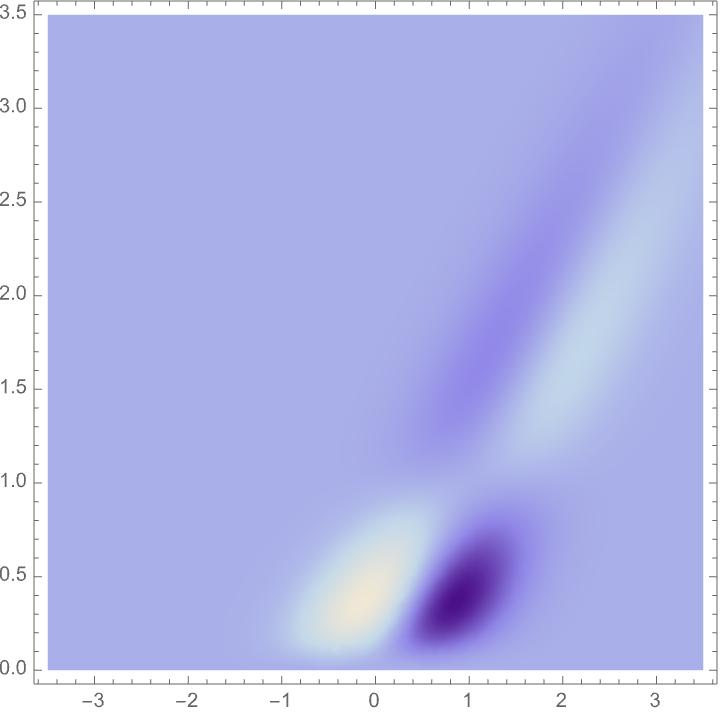}
     \end{tabular}
   \end{center}
   \caption{Variability of mixed spatio-temporal derivatives over a
     1+1-D spatio-temporal domain $T_{x {\bar t}}(x, t;\; s, \tau, v)
   = \partial_x \, \partial_{\bar t} \left( g(x - v \, t;\; s) \, \Psi(t;\; \tau, c) \right)$,
   corresponding to the combination of a
   first-order Gaussian derivative over the spatial domain and
   a first-order derivative of the time-causal limit kernel over the temporal domain,  
   for $s = \sigma_x^2$, $\tau = \sigma_t^2$ and $c =2$
   with $\sigma_x = 1/2$ and $\sigma_t = 1$,
   under variations of the image velocity $v$ of
   the kernels. Geometrically, such a variability will be the result if
   we vary the relative motion between the viewing direction and a
   moving local surface patch, while requiring the backprojected
   receptive fields to, due to requirements of {\em Galilean covariance\/},
   having the same effective effect when backprojected to the tangent
   plane of the surface.
   (Horizontal axes: spatial coordinate $x \in [-3.5, 3.5]$.
   Vertical axes: time $t \in [0, 3.5]$.)}
   \label{fig-1spat1tempdir-timecaus-spattempscsp}
\end{figure*}

\begin{figure*}[hbtp]
  \begin{center}
    \begin{tabular}{ccc}
      $\sigma_t = 1$ & $\sigma_t = 2$ & $\sigma_t = 4$ \\
      \includegraphics[width=0.30\textwidth]{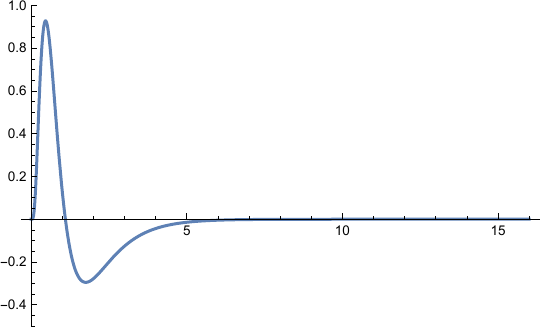}
      & \includegraphics[width=0.30\textwidth]{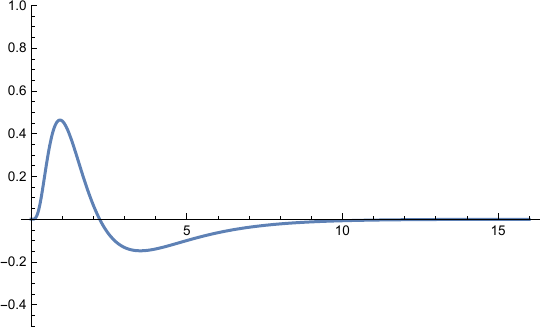}
      & \includegraphics[width=0.30\textwidth]{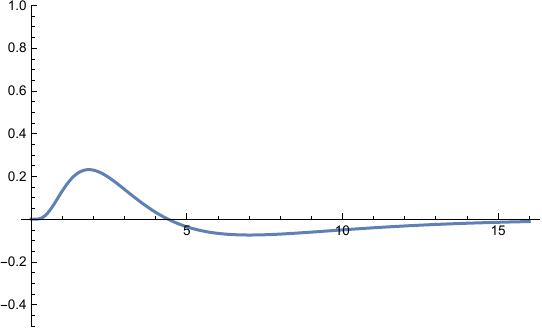}
     \end{tabular}
   \end{center}
   \caption{Variability of the first-order scale-normalized temporal
     derivatives of the time-causal limit kernel
     $T_{t}(t;\; \tau) = \sqrt{\tau} \, \partial_t \left( \Psi(t;\; \tau, c) \right)$
     for $c = 2$ under variations of
     the temporal scale parameter $\sigma_t = \sqrt{\tau}$, in this way
     demonstrating the effect of we, because of a requirement of
     {\em temporal scale covariance\/}, require the backprojected
     temporal receptive fields to correspond to the same structures in
     the temporal signal, when observing temporal structures that
     occur either faster or slower relative to a previously observed
     reference signal.
     (Horizontal axes: time $t \in [0, 16]$.
      Vertical axes: magnitude of the scale-normalized derivative $\in [-0.5, 1]$.)}
   \label{fig-timecaus-1tempders}
\end{figure*}

\subsection{Spatio-temporal derivative operators}
\label{sec-spat-temp-ders-gen-gauss-model}

The above purely spatio-temporal smoothing components of receptive fields
are then to combined with spatial and temporal derivative operations.
Over the spatial domain, we can compute either partial spatial derivatives
\begin{equation}
  \label{eq-def-spat-part-der}
  \partial_{x^{\alpha}} = \partial_{x_1}^{\alpha_1} \, \partial_{x_2}^{\alpha_2} 
\end{equation}
for different orders $\alpha = (\alpha_1, \alpha_2)^T$ of spatial
differentiation, or oriented directional derivatives in any direction
$\varphi \in \bbbr$
\begin{equation}
  \label{eq-dir-der-def}
  \partial_{\varphi}^m
  = (\cos \varphi \, \partial_{x_1} + \sin \varphi \, \partial_{x_2})^m
  = (e_{\varphi}^T \, \nabla_x)^m
\end{equation}
over different orientations $\varphi$ and 
different orders $m$ of spatial differentiation, where
$e_{\varphi} = (\cos \varphi, \sin \varphi)^T$
denotes the unit vector in the direction $\varphi$ and $\nabla_x$
denotes the spatial gradient operator according to
\begin{equation}
  \nabla_x
  = \left (
        \begin{array}{c}
          \partial_{x_1} \\
          \partial_{x_2}
        \end{array}
       \right).
\end{equation}
Over the temporal domain, we can, in turn, compute partial temporal derivatives
\begin{equation}
  \label{eq-temp-der-def}
  \partial_t^n
\end{equation}
for different orders $n$ of temporal differentiation, or
velocity-adapted temporal derivatives
\begin{equation}
  \label{eq-vel-adapt-der-def}
  \partial_{\bar t}^n
  = (v_1 \, \partial_{x_1} + v_2 \, \partial_{x_2} + \partial_t )^n
  = (v^T \, \nabla_x + \partial_t)^n
\end{equation}
for different image velocities $v = (v_1, v_2)^T$ and orders $n$ of
temporal differentiation.

Specifically, in relation to the parameters $\Sigma$ and $v$ of the
purely spatio-temporal smoothing component of the spatio-temporal
receptive fields in (\ref{eq-spat-temp-RF-model}), the image
orientations $\varphi$ in the directional derivative operators
(\ref{eq-dir-der-def}) should preferably be chosen in the directions of the
eigendirections of the spatial covariance matrix $\Sigma$, whereas the
image velocities $v$ in the velocity-adapted derivative operators
$\partial_{\bar t}^n$
should preferably be chosen equal to the image velocity $v$ in the
spatio-temporal smoothing kernel $T(x, t;\; s, \Sigma, \tau, v)$.

\subsection{Relations to biological vision}
\label{sec-rel-biol-vision}

In Lindeberg (\citeyear{Lin21-Heliyon}), it was demonstrated that the
receptive fields of neurons in the lateral geniculate
nucleus (LGN) as well as the receptive fields of simple cells in the
primary visual cortex (V1), as measured by neurophysiological cell
recordings by DeAngelis {\em et al.\/}\ (\citeyear{DeAngOhzFre95-TINS,deAngAnz04-VisNeuroSci}),
Conway and Livingstone (\citeyear{ConLiv06-JNeurSci}) and
Johnson {\em et al.\/}\ (\citeyear{JohHawSha08-JNeuroSci}),
can be well modelled by idealized receptive
fields derived from this generalized Gaussian derivative model for
receptive fields.

According to the treatment in Lindeberg (\citeyear{Lin21-Heliyon}) Sections~4.1--4.2,
the spatio-temporal receptive fields of ``non-lagged neurons'' and ``lagged neurons'',
which have rotationally symmetric response properties over the spatial domain,
can be modelled by idealized receptive fields
$h_{\scriptsize\mbox{LGN}} \colon \bbbr^2 \times \bbbr \times \bbbr_+ \times \bbbr_+
\rightarrow \bbbr$
of the form
\begin{equation}
  \label{eq-lgn-model-1}
  h_{\scriptsize\mbox{LGN}}(x, t;\; s, \tau)  
  = \pm \nabla_x^2 \, g(x;\; s) \, \partial_{t^n} \, h(t;\; \tau),
\end{equation}
where $\nabla_x^2 = \partial_{x_1 x_1} + \partial_{x_2 x_2}$ represents the spatial
Laplacian operator, and $h(t;\; \tau)$ represents a temporal smoothing
kernel, which in the most idealized
situation may correspond to the time-causal limit kernel $\Psi(t;\; \tau, c)$
according to (\ref{eq-FT-comp-kern-log-distr-limit}).

In this context, ``non-lagged neurons'' correspond to first-order
temporal derivatives, whereas ``lagged neurons'' correspond to
second-order temporal derivatives, see also
Ghodrati {\em et al.\/} (\citeyear{GhoKhaLeh17-ProNeurobiol})
for a more extensive treatment of the properties of visual neurons in the lateral
geniculate nucleus (LGN).

The spatio-temporal receptive fields of orientation selective simple
cells in the primary visual cortex (V1) can, in
turn, be modelled by idealized receptive fields
$T_{{\varphi}^{m} {\bar t}^n} \colon \bbbr^2 \times \bbbr \times \bbbr_+ \times \bbbs_+^2 \times \bbbr_+
\times \bbbr^2 \rightarrow \bbbr$
of the form
\begin{multline}
  \label{eq-spat-temp-RF-model-der}
    T_{{\varphi}^{m} {\bar t}^n}(x, t;\; s, \Sigma, \tau, v) 
    =  \\ =
        \partial_{\varphi}^{m} \, \partial_{\bar t}^n 
           \left( g(x - v \, t;\; s, \Sigma) \, h(t;\; \tau) \right),
\end{multline}
where
\begin{itemize}
\item
  $\partial_{\varphi}$ denotes a directional 
  derivative operator in one of the eigendirection of
  the spatial covariance matrix $\Sigma$ according to
  (\ref{eq-dir-der-def}),
\item
  $\partial_{\bar t}$ denotes a velocity-adapted
  temporal derivative operator in the direction $v$
  according to (\ref{eq-vel-adapt-der-def}),
  and
\item
  $h(t;\; \tau)$ again represents a set of first-order truncated
  exponential kernels coupled in cascade, which in the most idealized
  situation may correspond to the time-causal limit kernel $\Psi(t;\; \tau, c)$
  according to (\ref{eq-FT-comp-kern-log-distr-limit});
\end{itemize}
see
Lindeberg (\citeyear{Lin21-Heliyon}) Section~4.3 for explicit
biological modelling results.

Figures~\ref{fig-1spatders-scale-var}--\ref{fig-timecaus-1tempders}
show variabilities of idealized receptive
fields according to this model under (i)~spatial scaling
transformations, (ii)~spatial affine transformations,
(iii)~Galilean transformations and (iv)~temporal scaling
transformations.

This paper addresses the problem of modelling the effects, that joint
compositions of these types of image transformations have upon
receptive field responses, as well as how such joint compositions of
these geometric image image transformations can be interpreted
geometrically, for multi-view observations of dynamic scenes.


\begin{table*}[hbt]
  \begin{tabular}{lll}
    \hline
    Section & Topic & Contribution \\
    \hline
    \ref{sec-sc-norm-ders-isotropic}
            & Spatial scale-normalized derivatives over isotropic scale space
                    & Mainly review of Lindeberg (\citeyear{Lin97-IJCV}) \\
    \ref{sec-scale-cov-isotropic-spat-sc-norm-ders}
            & Covariance properties of isotropic 
              derivatives under pure spatial
              scaling transformations
                    & Mainly review of Lindeberg (\citeyear{Lin97-IJCV}) \\
    \ref{sec-aff-sc-norm-dir-ders}
            & Spatial affine scale-normalized directional derivatives over affine
              scale space
                    & New \\
    \ref{sec-cov-prop-aff-sc-norm-dir-ders}
            & Covariance properties of affine directional derivatives
              under spatial affine transformations
                    & New \\
    \ref{sec-sc-norm-aff-grad-op}
            & Spatial scale-normalized affine gradient operator over affine
              scale space
                    & New \\
    \ref{sec-cov-prop-sc-norm-aff-grad-op}
            & Covariance properties of scale-normalized affine gradient under
              spatial affine transformations
                    & New \\
    \ref{sec-sc-norm-aff-hess-op}
            &  Spatial scale-normalized affine Hessian operator over affine
              scale space
                    & New \\
    \ref{sec-cov-prop-sc-norm-aff-hess-mat}
            & Covariance properties of scale-normalized affine Hessian under
              spatial affine transformations
                    & New \\
    \ref{sec-sc-norm-temp-ders}
            & Temporal scale-normalized derivatives over temporal scale space
                    & Generalization of Lindeberg (\citeyear{Lin17-JMIV}) \\
    \ref{sec-cov-prop-temp-sc-ders}
            &  Covariance properties of temporal derivatives under temporal
              scaling transformations
                    & Generalization of Lindeberg (\citeyear{Lin17-JMIV}) \\
    \ref{sec-sc-norm-vel-adapt-temp-ders}
            & Spatio-temporal scale-normalized velocity-adapted 
              derivatives over spatio-temporal scale space
              & Extension of Lindeberg (\citeyear{Lin13-BICY}) \\
    \ref{sec-cov-prop-sc-norm-vel-adapt-temp-ders}
            & Covariance property of velocity-adapted derivatives under joint scaling
              transformations
              & New \\
              \hline
  \end{tabular}
  \caption{Overview of the conceptual contributions regarding
    essentially {\em individual\/} covariance properties in the different
    subsections in Section~\ref{sec-sc-norm-spat-temp-ders}, which
    then constitute the theoretical foundation for the in-depth studies
    of {\em joint\/} covariance properties, that will follow in
    Sections~\ref{sec-individ-cov-props}--\ref{sec-cues-3d-structure}.}
  \label{tab-contribs-sec3}
\end{table*}

\section{Scale-normalized spatial and temporal derivative operators}
\label{sec-sc-norm-spat-temp-ders}

When computing spatial and temporal derivatives from 
spatio-temporally smoothed video data, as obtained by convolution of
the video data with
the spatio-temporal smoothing kernel (\ref{eq-spat-temp-RF-model}), a
basic observation is that the magnitude of the computed
spatio-temporal derivatives will decrease in magnitude
with increasing values of the spatial and the temporal scale
parameters. To handle this problem, and to enable the definition of
spatial and temporal derivative operators, that are truly covariant
with regard to variations of the spatial and the temporal scale
parameters, that occur as parameters in the models of the
spatio-temporal receptive fields, we
will make use of scale-normalized derivative operators.

In this section, we will state the definitions of such scale-normalized derivative
operators, regarding both spatial and temporal derivatives over
different types of spatial, temporal or spatio-temporal domains, and show how
this notion leads to basic covariance properties under different types
of individual spatial and temporal scaling transformations.

For the purely spatial scale-normalized derivatives of the {\em isotropic\/}
scale-space representation, based on convolutions with rotationally
symmetric Gaussian kernels, as well as for the purely temporal
scale-space representations, defined from convolutions with either 1-D
non-causal temporal Gaussian kernels or the time-causal limit kernel,
the corresponding treatments in
Sections~\ref{sec-sc-norm-ders-isotropic}--\ref{sec-scale-cov-isotropic-spat-sc-norm-ders}
and~\ref{sec-sc-norm-temp-ders}--\ref{sec-cov-prop-temp-sc-ders}
will largely be reviews of previously existing results,
although with explicit formulations of scale-normalized directional
derivatives and gradient vectors over the
spatial domain in
Sections~\ref{sec-sc-norm-ders-isotropic}--\ref{sec-scale-cov-isotropic-spat-sc-norm-ders},
as covered by the general theory in Lindeberg (\citeyear{Lin97-IJCV})
but not explicitly stated there, as well as
to scale-normalized temporal derivates for more general
families of scale-covariant temporal smoothing kernels
in Sections~\ref{sec-sc-norm-temp-ders}--\ref{sec-cov-prop-temp-sc-ders},
restated as well as extended here to constitute foundations for further developments
building on these results, as well as to provide a balance in regard
to the new theoretical formulations, as detailed further below.%
\footnote{Except for the fact that
  Sections~\ref{sec-sc-norm-temp-ders}--\ref{sec-cov-prop-temp-sc-ders}
  generalize the previous
treatments for scale-normalized derivatives, computed based on
convolution with either non-causal 1-D temporal kernels or the
time-causal limit kernel, to temporal derivatives computed based on
temporal smoothing with an arbitrary
scale-covariant temporal smoothing kernel.}

For the purely spatial scale-normalized derivative operators, based on
convolutions with {\em anisotropic\/} Gaussian kernels,
the treatments in
Sections~\ref{sec-aff-sc-norm-dir-ders}--\ref{sec-cov-prop-sc-norm-aff-hess-mat} will, on the other
hand present a set of novel theoretical constructions regarding
affine-extended scale-normal\-ized derivative operators, which will
then allow for provable covariance
properties over either different subgroups of the affine group, or the
full affine group, depending on the different formulations of affine-extended
scale-normalized derivative operators.

Additionally, regarding the notion of scale-normalized
velocity-adapted derivatives over a joint spatio-temporal domain,
Sections~\ref{sec-sc-norm-vel-adapt-temp-ders}--\ref{sec-cov-prop-sc-norm-vel-adapt-temp-ders}
will state corresponding covariance properties of such spatio-temporal
derivative operations, which have not been sufficiently formulated in
previous work.

The scale-normalized spatial and temporal
derivative operators, defined in these ways, will then be essential to
obtain true covariance properties with associated transformation
properties of a particularly simple form,
under the different classes of
locally linearized joint spatio-temporal image transformations,
that we will consider 
between the image data obtained from different views of the
same scene.

Later in Section~\ref{sec-joint-cov-props}, we will then consider
specific compositions of these 
primitive individual types of geometric image transformations studied in this section.

Table~\ref{tab-contribs-sec3} provides a comprehensive overview of the
different types of contributions, that will follow in this section.

\subsection{Scale-normalized spatial derivative operators for a
  regular scale-space representation based on smoothing with
  rotationally symmetric Gaussian kernels}
\label{sec-sc-norm-ders-isotropic}

Given 2-D spatial image data $f \colon \bbbr^2 \rightarrow \bbbr$,
for a regular (isotropic) Gaussian scale-space representation
$L \colon \bbbr^2 \times \bbbr_+ \rightarrow \bbbr$,
defined by
convolution with rotationally symmetric Gaussian kernels
$g(\cdot;\; s, I)$ at scale $s \in \bbbr_+$, for which the
covariance matrix $\Sigma \in \bbbs_+^2$ in (\ref{eq-gauss-fcn-2D}) is a unit
matrix $\Sigma = I$,
\begin{equation}
  L(\cdot;\; s) = g(\cdot;\; s, I) * f(\cdot),
\end{equation}
the notion of
scale-normalized derivative operators corresponding to
the regular partial derivative operators (\ref{eq-def-spat-part-der}),
to be used at the spatial scale level $s \in \bbbr_+$ in the corresponding
spatial scale-space representation, can be defined according to
Lindeberg (\citeyear{Lin97-IJCV})
\begin{equation}
  \label{eq-def-spat-part-der-sc-norm-basic}
  \partial_{x^{\alpha},\norm}
  = s^{(\alpha_1 + \alpha_2)/2} \, \partial_{x_1}^{\alpha_1} \, \partial_{x_2}^{\alpha_2}.
\end{equation}
The corresponding scale-normalized analogues of the directional
derivative operators (\ref{eq-dir-der-def}) in the direction $\varphi$
will then be of the form
\begin{equation}
   \label{eq-dir-der-def-sc-norm-basic}
   \partial_{\varphi,\norm}^m
    = s^{m/2} \,  (\cos \varphi \, \partial_{x_1} + \, \sin \varphi \, \partial_{x_2})^m
    = s^{m/2} \, (e_{\varphi}^T \, \nabla_x)^m,
\end{equation}
and the corresponding scale-normalized spatial gradient operator will be 
\begin{equation}
  \label{eq-nabla-op}  
  \nabla_{x,\norm} = s^{1/2} \, \nabla_x.
\end{equation}
By multiplying the regular spatial derivative operators by the scale
parameter raised to a suitable power, proportional to the order of
spatial differentiation, the scale-normalized spatial derivative concept will
in this way compensate for the otherwise general decrease in the
magnitude of the spatially smoothed spatial derivatives with
increasing spatial scales, to enable truly scale-covariant spatial
derivative operators, whose magnitudes can be perfectly matched
under spatial scaling transformations, as will be detailed in the next section.

\subsection{Covariance property for scale-normalized
  spatial derivatives under spatial scaling transformations}
\label{sec-scale-cov-isotropic-spat-sc-norm-ders}

Consider next a spatial scaling transformation
\begin{equation}
  f'(x') = f(x)  \quad\quad\mbox{for}\quad\quad  x' = S_x \, x,
\end{equation}
with the spatial scaling factor $S_x \in \bbbr_+$, applied to the
corresponding isotropic Gaussian purely spatial scale-space
representations $L \colon \bbbr^2 \times \bbbr_+ \rightarrow \bbbr$
and $L' \colon \bbbr^2 \times \bbbr_+ \rightarrow \bbbr$ of
$f \colon \bbbr^2 \rightarrow \bbbr$ and
$f' \colon \bbbr^2 \rightarrow \bbbr$, respectively, defined according to
\begin{align}
  \begin{split}
    L(\cdot;\; s) = g(\cdot;\; s, I) * f(\cdot),
  \end{split}\\
  \begin{split}
    L'(\cdot;\; s') = g(\cdot;\; s', I) * f'(\cdot).
  \end{split}
\end{align}
As shown in (Lindeberg \citeyear{Lin97-IJCV} Equation~(16)),
these scale-space representations
obey spatial scale covariance over the
underlying spatial smoothing transformation, such that
\begin{equation}
  L'(x';\; s') = L(x;\; s)
\end{equation}
holds for matching values of
the spatial scale parameters according to
\begin{equation}
  s' = S_x^2 \, s.
\end{equation}
Given the definitions of the scale-normalized partial derivatives
(\ref{eq-def-spat-part-der-sc-norm-basic}) and the directional
derivatives (\ref{eq-dir-der-def-sc-norm-basic}) in the direction
$\varphi$ in the original domain, 
let us define corresponding scale-normalized spatial derivatives over the
transformed spatial domain according to
\begin{align}
  \begin{split}
     \partial_{{x'}^{\alpha},\norm}
     = {s'}^{(\alpha_1 + \alpha_2)/2} \, \partial_{x'_1}^{\alpha_1} \, \partial_{x'_2}^{\alpha_2},
  \end{split}\\
  \begin{split}
    \partial_{\varphi',\norm}^m =
    {s'}^{m/2} \, (e_{\varphi'}^T \, \nabla_{x'})^m,
   \end{split}
\end{align}
where
\begin{itemize}
\item
  $e_{\varphi'} = (\cos \varphi', \sin \varphi')^T$ denotes the
  corresponding unit vector in the transformed direction
  $\varphi' \in \bbbr$ after the image transformation,
\item
  the entity
  \begin{equation}
    \nabla_{x'}
    = \left (
      \begin{array}{c}
        \partial_{x'_1} \\
        \partial_{x'_2}
      \end{array}
    \right)
  \end{equation}
  denotes the transformed gradient operator, and
\item
  here the angles for the directional derivatives are not affected
  by the uniform spatial scaling transformation, such that
  \begin{equation}
    \varphi' = \varphi.
  \end{equation}
\end{itemize}
Let us also define the corresponding scale-normalized gradient
operator over the transformed domain according to
\begin{equation}
  \nabla_{x',\norm} = {s'}^{1/2} \, \nabla_{x'}.
\end{equation}
Then, since the scale-normalized spatial derivative operators over the
two respective domains will be related according to
(see Lindeberg (\citeyear{Lin97-IJCV}) Equation~(20) for the specific
choice of the scale normalization power $\gamma = 1$ in that paper)
\begin{equation}
  \partial_{x'_i,\norm} = \partial_{x_i,\norm},
\end{equation}
it follows that
the scale-normalized spatial derivatives of the
transformed spatial scale-space representation $L'$
will be equal to the scale-normalized spatial derivatives of the
transformed spatial scale-space representation $L$, such that
\begin{align}
  \begin{split}
     L'_{{x'}^{\alpha},\norm}(x';\; s') & = L_{{x}^{\alpha},\norm}(x;\; s),
  \end{split}\\
  \begin{split}
    \label{eq-eq-sc-norm-nabla-spat-scsp-spat-sc-transf}
     (\nabla_{x',\norm} L')(x';\; s') & = (\nabla_{x,\norm} L)(x;\; s),
  \end{split}\\
  \begin{split}
    \label{eq-eq-sc-norm-dirders-spat-scsp-spat-sc-transf}    
     L'_{{\varphi'}^m,\norm}(x';\; s') & = L_{\varphi^m,\norm}(x;\; s),
  \end{split}
\end{align}
which thus constitute covariance properties for
scale-normal\-ized spatial derivatives of an isotropic purely
spatial scale-space representation under spatial scaling
transformations.

When interpreted geometrically, these spatial scale covariance
properties mean that, if we observe the same scene from different
distances, while keeping the viewing direction constant, then the
scale-normalized spatial derivative responses can, to first order of
approximation, be perfectly matched, when viewing the same local surface
patch from different distances, along the same viewing direction.

\subsection{Scale-normalized directional derivative operators for an
  affine scale-space representation based on smoothing with
  anisotropic affine Gaussian kernels}
\label{sec-aff-sc-norm-dir-ders}

For the later developments in this paper, we do, in addition to
the above isotropic scale-normalized derivative concept, for spatial
scale-space representations based on convolutions with rotationally
symmetric Gaussian kernels, also need
to define scale-normalized spatial derivative operators for spatial
derivatives that are to be computed based on an affine Gaussian
scale-space representation
$L \colon \bbbr^2 \times \bbbr_+ \times \bbbs_+^2 \rightarrow \bbbr$
of any 2-D image $f \colon \bbbr^2 \rightarrow \bbbr$,
obtained by spatial smoothing with
anisotropic affine Gaussian kernels
$g \colon \bbbr^2 \times \bbbr_+ \times \bbbs_+^2 \rightarrow \bbbr$,
according to (\ref{eq-gauss-fcn-2D})
\begin{equation}
  \label{eq-def-aff-scsp}
  L(\cdot;\; s, \Sigma) = g(\cdot;\; s, \Sigma) * f(\cdot),
\end{equation}
that is, based on (symmetric and positive definite) $2 \times 2$ spatial
covariance matrices $\Sigma \in \bbbs_+^2$ that are not generally
equal to a unit matrix $I$.
For this reason, we will in the following extend the above scale-normalized derivative
concept from an isotropic Gaussian scale-space representation to an
affine Gaussian scale-space representation in different ways.

In this section, we will first develop such a notion of
scale-normalized derivatives for directional derivatives defined
from an affine Gaussian scale-space representation.
Later, we will then develop the other notions of the
scale-normalized affine gradient operator
(see Section~\ref{sec-sc-norm-aff-grad-op})
and the scale-normalized affine Hessian operator
(see Section~\ref{sec-sc-norm-aff-hess-op}).

\subsubsection{Definition of the affine scale-normalized directional
  derivative operator}

\begin{figure}[hbt]
  \begin{center}
    \includegraphics[width=0.70\textwidth]{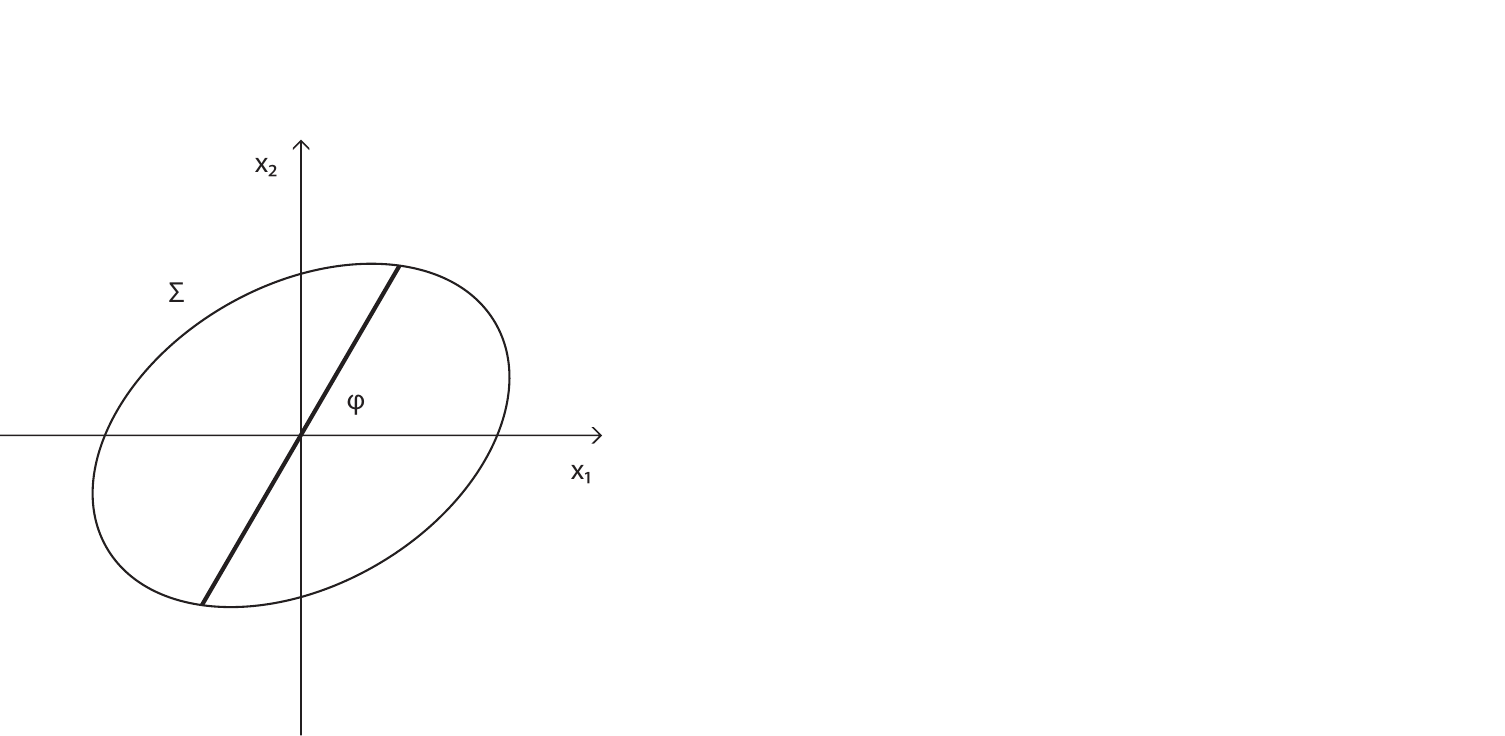}
  \end{center}
  \caption{The definition of the affine scale-normalized derivative
    operator according to (\ref{eq-aff-sc-norm-dir-der}) is based on
    extracting the amount of spatial smoothing in the direction
    $e_{\varphi} = (\cos \varphi, \sin \varphi)$ of the spatial
    covariance matrix $\Sigma$, here visualized in terms of an
    intersection of an ellipse representation of the spatial covariance matrix
  $\Sigma$.}
  \label{fig-geom-ill-aff-sc-norm-dir-der}
\end{figure}

Given an affine spatial scale-space representation $L(x;\; s, \Sigma)$, that has been
computed according to (\ref{eq-def-aff-scsp}), we define the affine
scale-normalized directional derivative operator in the direction
$\varphi$, with the unit vector in this direction denoted $e_{\varphi}$, according to
\begin{equation}
  \label{eq-aff-sc-norm-dir-der}
  \partial_{\varphi,\norm}^m
  = s^{m/2} \, (e_{\varphi}^T \, \Sigma \, e_{\varphi})^{m/2} \, \partial_{\varphi}^m.
\end{equation}
A general motivation for this definition, is that the entity
$e_{\varphi}^T \, \Sigma \, e_{\varphi}$ should, disregarding the
effect of the complementary scalar scale parameter $s$, reflect the amount of
spatial smoothing, that convolution with an affine Gaussian kernel with
spatial covariance matrix $\Sigma$ corresponds to, when measured in
the spatial direction $e_{\varphi}$ only;
see Figure~\ref{fig-geom-ill-aff-sc-norm-dir-der} for an illustration.

Specifically, if the
coordinate system is rotated in such a way that the covariance matrix becomes
a diagonal matrix $\Sigma = \diag(\lambda_1, \lambda_2)$
with $\lambda_1 \in \bbbr_+$ and $\lambda_2 \in \bbbr_+$,
then if the unit vector $e_{\varphi}$ for computing the
directional derivative is selected equal to one of the eigenvectors
$e_i$ of the spatial covariance matrix $\Sigma$, then
the entity $e_{\varphi}^T \, \Sigma \, e_{\varphi}$ will
select the eigenvalue $\lambda_i$ of the spatial covariance matrix
corresponding to that eigenvector $e_i$. Since the affine Gaussian
convolution operation in such a configuration will reduce to separable smoothing
with 1-D Gaussian kernels with spatial scale parameters $\lambda_1$ and
$\lambda_2$ along the spatial eigendirections $e_1$ and $e_2$,
respectively, this
implies that the resulting spatial scale normalization factor
$\lambda_i^{m/2}$, for spatial derivatives of orders $m$ in the
eigendirection $e_i$ of the spatial covariance matrix $\Sigma$, will, for the
stated definition, correspond precisely to the regular spatial
scale normalization factor $\lambda_i^{m/2}$ for the separated 1-D scale-space
representations along the respective orthogonal eigendirections of the
spatial covariance matrix.

Furthermore, we can note that in the special case of choosing the spatial
covariance matrix $\Sigma$ equal to the unit matrix $I$,
the affine scale-normalized directional
derivative operator $\partial_{\varphi,\norm}^m$ according to
(\ref{eq-aff-sc-norm-dir-der}) reduces to the
previously defined isotropic scale-normalized directional
derivative operator $\partial_{\varphi,\norm}^m$ according to
(\ref{eq-dir-der-def-sc-norm-basic}).

For the special case of a regular Gaussian scale-space representation,
defined by convolution with rotationally symmetric Gaussian kernels
for $\Sigma = I$,
this new definition of affine scale-normalized Gaussian directional derivatives
is therefore consistent with the previous formulated
notion of isotropic scale-normalized directional derivatives.

In the following, we will analyze and formulate explicit covariance
properties for this notion of affine scale-normal\-ized directional
derivatives, for the cases of two important subgroups of the affine
group. By necessity, the mathematical
details may be somewhat technical. The reader more interested in the
general overall results than the details of the derivations
should however, without major loss of continuity, be able to skip this treatment,
to then continue with Section~\ref{sec-sc-norm-temp-ders}, while 
noting the three below main theoretical results, summarized
under the below boldface headers ``Summary of main result'',
with the corresponding geometric illustrations of the first two main
results in
Figures~\ref{fig-geom-int-sim-transf-aff-norm-dir-der-cov-prop}
and~\ref{fig-geom-int-cov-prop-aff-sc-norm-ders-coupl-eigendecomp}.

\subsection{Covariance properties for affine scale-normalized
  directional derivatives under special subgroups of spatial affine transformations}
\label{sec-cov-prop-aff-sc-norm-dir-ders}

Consider a spatial affine transformation between two 2-D images
$f \colon \bbbr^2 \rightarrow \bbbr$ and $f' \colon \bbbr^2 \rightarrow \bbbr$
of the form
\begin{equation}
  \label{eq-spat-aff-transf-def-aff-sc-norm-ders}
  f'(x') = f(x) \quad\quad\mbox{for}\quad\quad x' = A \, x,
\end{equation}
where $A$ is a non-singular $2 \times 2$ affine transformation matrix,
and with the corresponding affine Gaussian scale-space
representations
$L \colon \bbbr^2 \times \bbbr_+ \times \bbbs_+^2 \rightarrow \bbbr$
and
$L' \colon \bbbr^2 \times \bbbr_+ \times \bbbs_+^2 \rightarrow \bbbr$
of $f$ and $f'$, respectively, defined according to
\begin{align}
  \begin{split}
    \label{eq-def-aff-scsp-again}
    L(\cdot;\; s, \Sigma) = g(\cdot;\; s, \Sigma) * f(\cdot),
  \end{split}\\
  \begin{split}
    \label{eq-def-aff-scsp-again-prim}
    L'(\cdot;\; s', \Sigma') = g(\cdot;\; s', \Sigma') * f'(\cdot),
  \end{split}
\end{align}
with matching values of the spatial scale parameters
$s \in \bbbr_+$ and $s' \in \bbbr_+$ as
well as the $2 \times 2$ spatial covariance matrices
$\Sigma \in \bbbs_+^2$ and $\Sigma' \in \bbbs_+^2$ over
the two domains, such that%
\footnote{To derive this expression, we can
  set $B = A$, $\Sigma_L = s \, \Sigma$ and $\Sigma_R = s' \, \Sigma'$
  in the transformation property of the spatial covariance matrices
  $\Sigma_R = B \, \Sigma_L \, B^T$ under a spatial affine
  transformation of the form $x_R = B \, x_L$ according to Equation~(30)
  in Lindeberg and G{\aa}rding (\citeyear{LG96-IVC}).
  In this paper, we do, however, on the other hand, overparameterize the degrees of freedom
  in the affine Gaussian convolution kernels, in order to later more clearly
  separate the degree of freedom in uniform scaling transformations from
  the degrees of freedom in non-isotropic affine image
  transformations, see also Lindeberg (\citeyear{Lin25-BICY}) for a canonical
  parameterization of the degrees of freedom in 2-D spatial affine image
  transformations, based on a singular value decomposition of the
  affine transformation matrix $A$. The underlying reasons for this
  aim are: (i)~to prepare for the degrees of freedom in uniform scaling
  transformations and more general non-isotropic affine
  transformations to be studied in Section~\ref{sec-joint-cov-props}, and also
  (ii)~to prepare for possible different ways of normalizing the
  spatial covariance matrices $\Sigma$ with regard to specific
  geometric interpretations of the imaging situation.}
\begin{equation}
  \label{eq-transf-prop-sc-par-spat-cov-mat-pure-aff-scsp}
  s' \Sigma' = s \, A \, \Sigma \, A^T,
\end{equation}
which then implies that the affine scale-space representations will be
equal for these matching values of the scale parameters and the
covariance matrices (Lindeberg and G{\aa}rding (\citeyear{LG96-IVC}) Equation~(29))
\begin{equation}
  \label{eq-equal-aff-scsp-repr-def-aff-sc-norm-ders}
  L'(x';\; s', \Sigma') = L(x;\; s, \Sigma).
\end{equation}
Let us, in analogy with the definition of affine scale-normalized
derivatives along the direction $\varphi$ in the original domain according to
(\ref{eq-aff-sc-norm-dir-der}),
define affine scale-normalized directional derivatives
along the direction $\varphi'$ in the
transformed domain according to
\begin{equation}
  \label{eq-aff-sc-norm-dir-der-prim}
  \partial_{\varphi',\norm}^m
  = {s'}^{m/2} \, (e_{\varphi'}^T \, \Sigma' \, e_{\varphi'})^{m/2} \, \partial_{\varphi'}^m.
\end{equation}
Replacing $s' \, \Sigma'$ in this expression by
$s' \Sigma' = s \, A \, \Sigma \, A^T$ according to
(\ref{eq-transf-prop-sc-par-spat-cov-mat-pure-aff-scsp}),
and using the following transformation property of the unit vectors
\begin{equation}
  \label{eq-match-unit-vectors-spat-aff}
  e_{\varphi'} = \frac{A \, e_{\varphi}}{\| A \, e_{\varphi} \| },
\end{equation}
while additionally noting that
\begin{equation}
  \label{eq-norm-aff-transf-unit-vector}
   \| A \, e_{\varphi} \|^2 = e_{\varphi}^T \, A^T A \, e_{\varphi},
\end{equation}
implies that we can rewrite (\ref{eq-aff-sc-norm-dir-der-prim}) as
\begin{equation}
  \label{eq-aff-sc-norm-dir-der-prim-rewrite}
  \partial_{\varphi',\norm}^m
  = {s}^{m/2} \,
     \left(
        \frac{e_{\varphi}^T \, A^T A \, \Sigma \,  A^T A \, e_{\varphi}}
                {e_{\varphi}^T \, A^T A \, e_{\varphi}}
      \right)^{m/2}
      \partial_{\varphi'}^m.
\end{equation}
From the definitions of the regular directional derivative operators
over the respective image domains
\begin{align}
 \begin{split}
     \partial_{\varphi} = e_{\varphi}^T \, \nabla_{x},
  \end{split}\\
  \begin{split}
     \partial_{\varphi'} = e_{\varphi'}^T \, \nabla_{x'},
  \end{split}
\end{align}
and the transformation property%
\footnote{Concerning the notation, we throughout this paper denote the
  transpose of an inverse matrix as $A^{-T} = (A^{-1})^T$.}
of the spatial gradient operator under
the spatial affine transformation (\ref{eq-spat-aff-transf-def-aff-sc-norm-ders})
\begin{equation}
  \nabla_x = A^T \, \nabla_{x'} \quad\quad\mbox{implying that}\quad\quad
  \nabla_{x'} = A^{-T} \, \nabla_{x},
\end{equation}
we can after some simplifications obtain that the
scale-normal\-ized directional derivative operators in the two
domains are related according to
\begin{equation}
  \partial_{\varphi',\norm} = \frac{\partial_{\varphi,\norm}}{\| A \, e_{\varphi} \| }.
\end{equation}
To make a preliminary summary, while again making use of the
relationship (\ref{eq-norm-aff-transf-unit-vector}), this means that the affine
scale-normalized directional derivative operators over the two image domains
can be written as
\begin{align}
  \begin{split}
    \label{eq-aff-sc-norm-dir-der-reduced}
     \partial_{\varphi,\norm}^m
     & = s^{m/2} \, (e_{\varphi}^T \, \Sigma \, e_{\varphi})^{m/2} \, \partial_{\varphi}^m,
   \end{split}\\
  \begin{split}
    \label{eq-aff-sc-norm-dir-der-reduced-prim}    
       \partial_{\varphi',\norm}^m
       & = {s}^{m/2} \,
              \left(
                 \frac{e_{\varphi}^T \, A^T A \, \Sigma \,  A^T A \, e_{\varphi}}
                          {(e_{\varphi}^T \, A^T A \, e_{\varphi})^2}
              \right)^{m/2}
              \partial_{\varphi}^m.
     \end{split}
\end{align}

\subsubsection{Analysis for the special case with spatial similarity
  transformations}
\label{sec-aff-anal-dir-der-sim-group}

To analyse the possibility of the expressions
(\ref{eq-aff-sc-norm-dir-der-reduced}) and
(\ref{eq-aff-sc-norm-dir-der-reduced-prim})
leading to the same result, for the special case of spatial similarity
transformations, let us insert
\begin{equation}
  A = S_x \, R,
\end{equation}
where $S_x \in \bbbr_+$ is a uniform spatial scaling factor and $R$ is a $2 \times 2$
rotation matrix with $R^T \, R = R \, R^T = I$, such that $A^T A = S_x^2 \, I$,
where $I$ is the $2 \times 2$ unit matrix, into the above expressions.
Then, after simplification,
(\ref{eq-aff-sc-norm-dir-der-reduced}) and
(\ref{eq-aff-sc-norm-dir-der-reduced-prim}) reduce to the similar expressions
\begin{align}
  \begin{split}
    \label{eq-aff-sc-norm-dir-der-reduced-similarity}
     \partial_{\varphi,\norm}^m
     & = s^{m/2} \, (e_{\varphi}^T \, \Sigma \, e_{\varphi})^{m/2} \, \partial_{\varphi}^m,
   \end{split}\\
  \begin{split}
    \label{eq-aff-sc-norm-dir-der-reduced-prim-similarity}    
       \partial_{\varphi',\norm}^m
       & = {s}^{m/2} \, (e_{\varphi}^T \, \Sigma \, e_{\varphi})^{m/2} \, \partial_{\varphi}^m,
     \end{split}
\end{align}
implying that when applied over their respective image domains, these
two underlying affine scale-normalized directional operators
will lead to the same result
\begin{equation}
  \label{eq-aff-sc-norm-cov-sim-transf}
  \partial_{\varphi',\norm}^m L'(x';\; s', \Sigma')
  = \partial_{\varphi,\norm}^m L(x;\; s, \Sigma).
\end{equation}

\noindent
{\bf Summary of main result:}
To conclude, this result shows that, when applied to an affine Gaussian scale-space
representation (\ref{eq-def-aff-scsp}),
defined by convolutions with arbitrary affine Gaussian
kernels, the affine scale-normalized directional derivative concept,
defined according to (\ref{eq-aff-sc-norm-dir-der-prim}) 
\begin{equation}
  \partial_{\varphi,\norm}^m
  = s^{m/2} \, (e_{\varphi}^T \, \Sigma \, e_{\varphi})^{m/2} \,
  \partial_{\varphi}^m,
\end{equation}
is for every direction $e_{\varphi}$ in the image plane covariant under
arbitrary combinations of uniform scaling transformations and
rotations of the form
\begin{equation}
  f'(x') = f(x) \quad\quad\mbox{for}\quad\quad x' = S_x \, R \, x,
\end{equation}
such that for the matching image orientation
$e_{\varphi'} = R \, e_{\varphi}$, and
with the transformed
affine scale-normalized directional derivative operator in this
direction defined according to
\begin{equation}
\partial_{\varphi,'norm}^m
  = {s'}^{m/2} \, (e_{\varphi'}^T \, \Sigma' \, e_{\varphi'})^{m/2} \,
  \partial_{\varphi'}^m,
\end{equation}
then the relationship
\begin{equation}
  \label{eq-aff-norm-dir-der-sim-transf}
  \partial_{\varphi',\norm}^m L'(x';\; s', \Sigma')
  = \partial_{\varphi,\norm}^m L(x;\; s, \Sigma)
\end{equation}
will hold
for all values of the spatial scale parameter $s \in \bbbr_+$ and
the $2 \times 2$  spatial covariance matrix $\Sigma$ in the original domain,
provided that the spatial scale parameter $s' \in \bbbr_+$ and
the spatial covariance matrix $\Sigma'$ over the transformed domain
are related according to
\begin{equation}
  \label{eq-rel-sc-pars-cov-mats-cov-prop-aff-sc-norm-dir-der-sim-transf}
  s' \Sigma'  = s \, A \, \Sigma \, A^T= s \, S_x^2 \, R \, \Sigma \, R^T.
\end{equation}

\begin{figure}[hbt]
  \begin{center}
    \includegraphics[width=0.35\textwidth]{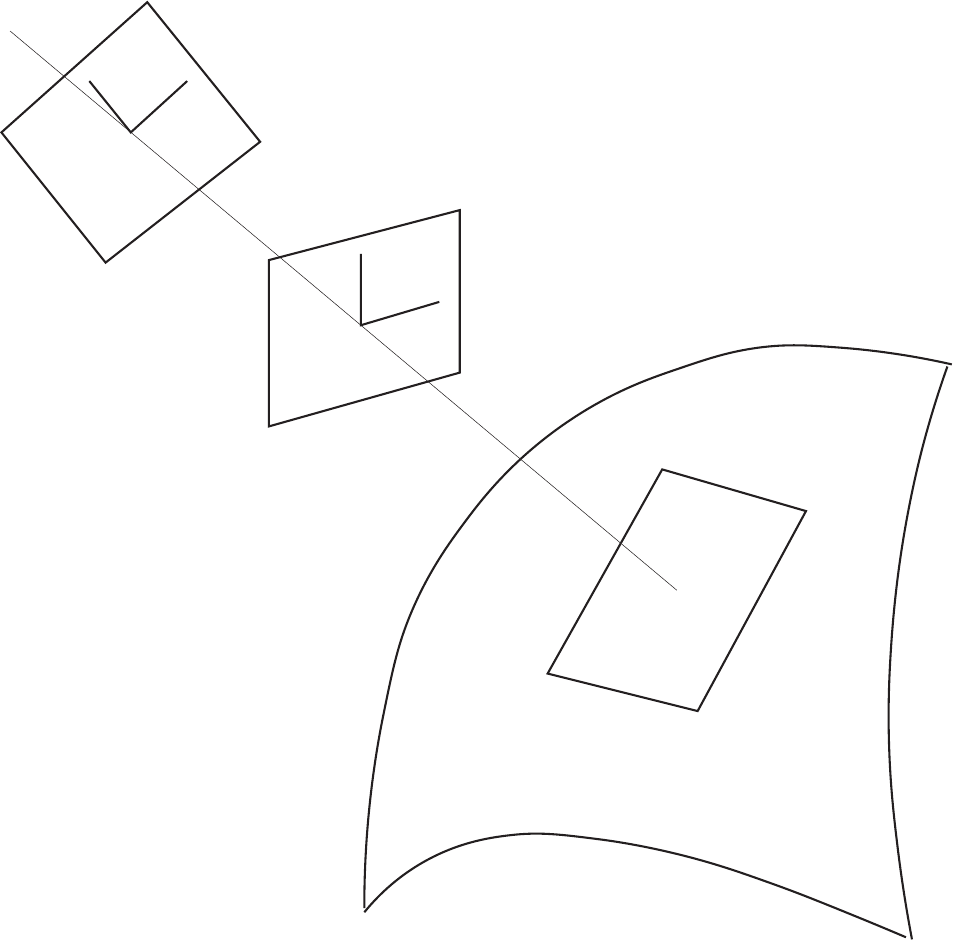}
  \end{center}  
   \caption{In terms of geometric interpretation, the covariance
    property (\ref{eq-aff-norm-dir-der-sim-transf}) for the affine
    scale-normalized derivative operator
    (\ref{eq-aff-sc-norm-dir-der-prim}) under similarity
    transformations means that, if two
    cameras view the same local surface patch along the same optical
    axis, while being at possibly difference distances from the object and also
    having possibly rotated orientations of the image planes relative
    to each other around the optical axis, then
    the affine scale-normalized derivatives according to
    (\ref{eq-aff-sc-norm-dir-der-reduced-similarity}) and
    (\ref{eq-aff-sc-norm-dir-der-reduced-prim-similarity})
    can, to first order of
    approximation, be perfectly matched between such different views of
    the same local surface patch, provided that the scale parameters
    and the covariance matrices of the resulting spatial receptive fields are
    matched according to (\ref{eq-rel-sc-pars-cov-mats-cov-prop-aff-sc-norm-dir-der-sim-transf}).}
  \label{fig-geom-int-sim-transf-aff-norm-dir-der-cov-prop}
\end{figure}

\subsubsection{Geometric interpretation of the analysis for the
  special case with spatial similarity
  transformations}

With regard to geometric interpretation of the above result for the
special case of the similarity group,
if we interpret the composed
spatial image transformation as a locally linear approximation of the
perspective mapping from the tangent plane of a local surface patch to
the image plane, this covariance property under spatial similarity
transformations implies that the affine scale-normalized directional
derivative responses can, to first order of approximation,
be perfectly matched between different views of
the same local surface patch, when varying the distance between the
viewed object and the observer, as well as when rotating either the camera or the
object around the optical axis;
see Figure~\ref{fig-geom-int-sim-transf-aff-norm-dir-der-cov-prop} for an illustration.

Concerning the dimensionality of the manifold spanned by this
covariance property, we have that the spatial scale parameter $s$ has
dimensionality~1 and the spatial covariance matrix $\Sigma$ has
dimensionality~3. Due to the coupling of these parameters of the form
$s \, \Sigma$, these parameters together do, however, only correspond to a variability over
3~effective dimensions in the effective parameter space. The affine
transformation matrices according to the similarity group
$A = S_x \, R$ span 2~out of the 4~dimensions in the variability of
the 2-D affine group.
The degree of freedom in the direction $\varphi$ adds 1~dimension to
this space. Thus, we have that this covariance result under
spatial similarity transformations spans 6~out of the totally 8~dimensions
in the variability of computing affine scale-normalized derivatives
from an affine scale-space representation in different directions in
the image plane, that are possible under the different types of image
transformations that are spanned by the full affine group,
as well as under the different
parameter settings that can be performed for the affine Gaussian
spatial smoothing component in the spatial receptive fields.

\subsubsection{Analysis for the special case with coupled eigenvalue
  decompositions of the spatial covariance matrix $\Sigma$ and the
  affine transformation matrix $A$}
\label{sec-aff-anal-dir-der-coupled-eigen-decomp}

Let us next assume that the eigenvalue decompositions of the spatial
covariance matrix $\Sigma$ and the affine transformation matrix $A$
are coupled, in such a way that
\begin{align}
  \begin{split}
    \label{eq-cov-mat-eigen-decomp-coupled-proof}
    \Sigma & = U \, \diag(\lambda_1, \lambda_2) \, U^T,
  \end{split}\\
  \begin{split}
    \label{eq-transf-mat-eigen-decomp-coupled-proof}    
    A & = U \, \diag(S_1, S_2) \, U^T,
  \end{split}           
\end{align}
where
\begin{itemize}
\item
  $\diag(\lambda_1, \lambda_2)$ is a $2 \times 2$ diagonal matrix with the
    eigenvalues $\lambda_1 \in \bbbr_+$ and $\lambda_2 \in \bbbr_+$ of the spatial covariance
    matrix $\Sigma$,
\item
  $\diag(S_1, S_2)$ is a $2 \times 2$ diagonal matrix with the
    eigenvalues $S_1 \in \bbbr_+$ and $S_2 \in \bbbr_+$ of the affine transformation
    matrix $A$, and thus representing the two spatial scaling factors
    $S_1$ and $S_2$ of a non-uniform spatial scaling transformation, and
\item
  $U$ is a $2 \times 2$ real unitary matrix, such that $U^T U = U\, U^T = I$,
  with its two columns $e_1$ and $e_2$
    constituting the eigenvectors of both the spatial covariance
    matrix $\Sigma$ and the affine transformation matrix $A$.
\end{itemize}
Note, however, that we do not here assume the eigenvalues to be ordered with
respect to magnitude. Instead, we assume that the eigenvalues are
ordered in such a way that, the order of the eigenvectors is the same
for both the spatial covariance matrix $\Sigma$ and the affine
transformation matrix $A$.

Given the expressions for the spatial covariance matrix $\Sigma$ according
to (\ref{eq-cov-mat-eigen-decomp-coupled-proof}) and the affine
transformation matrix $A$ according to
(\ref{eq-transf-mat-eigen-decomp-coupled-proof}), the expression for
the transformed transformation matrix $\Sigma'$ according to
(\ref{eq-transf-prop-sc-par-spat-cov-mat-pure-aff-scsp}) does after
simplification assume the form
\begin{equation}
  s' \, \Sigma'
  = s \, A \, \Sigma \, A^T
  = s \, U \, \diag(S_1^2 \, \lambda_1, S_2^2 \, \lambda_2) \, U^T,
\end{equation}
which does thus also share a coupled eigenvalue decomposition, in
relation to the
original spatial covariance matrix $\Sigma$ and the affine
transformation matrix $A$.

If we next choose the unit vector $e_{\varphi}$, for defining the
affine scale-normalized directional derivative operator
$\partial_{\varphi,\norm}$ in the original domain
according to (\ref{eq-aff-sc-norm-dir-der}),
equal to $e_i$, where $e_i$ is one
of the unit vectors $e_1$ or $e_2$ in the above unitary matrix $U$,
which in turn for $e_{\varphi} = e_i$ implies that also the
transformed unit vector becomes
\begin{equation}
  e_{\varphi'}
  = \frac{A \, e_{\varphi}}{\| A \, e_{\varphi} \|}
  = \frac{U \, \diag(S_1, S_2) \, U^T e_i}
             {\| U \, \diag(S_1, S_2) \, U^T e_i \|}
  = e_i,
\end{equation}
and then make use of the result that
\begin{equation}
  A^T A = U \, \diag(S_1^2, S_2^2) \ U^T,
\end{equation}
we then obtain that the expression
\begin{equation}
  e_{\varphi}^T \, \Sigma \, e_{\varphi}
  = e_i^T \, U \, \diag(\lambda_1, \lambda_2) \ U^T e_i
\end{equation}
in the expression for the affine scale-normalized derivative operator
$\partial_{\varphi,\norm}$ in (\ref{eq-aff-sc-norm-dir-der-reduced})
for the unit direction $e_{\varphi} = e_i$ reduces to
\begin{equation}
  e_{\varphi}^T \, \Sigma \, e_{\varphi} = \lambda_i.
\end{equation}
It also follows that the expression%
\footnote{In the expression below, as well as in following
  mathematical equations in this paper, the notation ``$\times$'' means means a mere
  multiplication, however, used here to separate different components in a
  product, or later also as a binding symbol between multiplications of
  multiple components in more complex expressions over multiple lines.}
\begin{equation}
  e_{\varphi'}^T \, \Sigma' \, e_{\varphi'}
  = \frac{s}{s'}  \times
      \frac{e_{\varphi}^T \, A^T A \, \Sigma \,  A^T A \, e_{\varphi}}
              {(e_{\varphi}^T \, A^T A \, e_{\varphi})^2}
\end{equation}
in the expression for the affine scale-normalized derivative operator
$\partial_{\varphi',\norm}$ in (\ref{eq-aff-sc-norm-dir-der-reduced-prim})
for the transformed unit direction $e_{\varphi'} = e_i$ reduces to
\begin{equation}
  e_{\varphi'}^T \, \Sigma' \, e_{\varphi'}
  = \frac{s}{s'}  \times \lambda_i.
\end{equation}
Thus, we therefore, for $e_{\varphi} = e_{\varphi'} = e_i$, have that
\begin{align}
  \begin{split}
    \label{eq-aff-sc-norm-dir-der-reduced-coupled}
     \partial_{\varphi,\norm}^m
     & = s^{m/2} \, \lambda_i^{m/2} \, \partial_{\varphi}^m,
   \end{split}\\
  \begin{split}
    \label{eq-aff-sc-norm-dir-der-reduced-prim-coupled}    
       \partial_{\varphi',\norm}^m
       & = {s}^{m/2} \, \lambda_i^{m/2} \, \partial_{\varphi}^m,
     \end{split}
\end{align}
implying that
when applied their respective original domains, these
two underlying affine scale-normalized directional operators
will, for $e_{\varphi} = e_{\varphi'} = e_i$, lead to the same result
\begin{equation}
  \partial_{\varphi',\norm}^m L'(x';\; s', \Sigma')
  = \partial_{\varphi,\norm}^m L(x;\; s, \Sigma).
\end{equation}

\noindent
{\bf Summary of main result:}
To summarize, this result shows that provided that we assume that
the eigenvalue decompositions of the $2 \times 2$ spatial covariance matrix
$\Sigma$ and the $2 \times 2$ affine transformation matrix $A$ are coupled
according to (\ref{eq-cov-mat-eigen-decomp-coupled-proof}) and
(\ref{eq-transf-mat-eigen-decomp-coupled-proof})
\begin{align}
  \begin{split}
    \label{eq-cov-mat-eigen-decomp-coupled-proof-2}
    \Sigma & = U \, \diag(\lambda_1, \lambda_2) \, U^T,
  \end{split}\\
  \begin{split}
    \label{eq-transf-mat-eigen-decomp-coupled-proof-2}    
    A & = U \, \diag(S_1, S_2) \, U^T,
  \end{split}           
\end{align}
where $U$ is a some real $2 \times 2$ unitary matrix,
and provided that we then apply the
affine scale-normalized directional derivative operator
according to (\ref{eq-aff-sc-norm-dir-der-prim}) 
\begin{equation}
  \partial_{\varphi,\norm}^m
  = s^{m/2} \, (e_{\varphi}^T \, \Sigma \, e_{\varphi})^{m/2} \,
  \partial_{\varphi}^m
\end{equation}
in a unit direction $e_{\varphi} = e_i$, chosen as either of the two
columns $\{e_1, e_2 \}$ in the unitary matrix $U$ in the above
eigendecompositions, to the affine Gaussian scale-space
representation $L(x;\; s, \Sigma)$ in the original domain,
then for arbitrary choices of the non-uniform
scaling matrix $\diag(S_1, S_2)$ in spatial image transformations of
the form
\begin{equation}
  f'(x') = f(x) \quad\quad\mbox{for}\quad\quad
  x' = U \, \diag(S_1, S_2) \, U^T x,
\end{equation}
such that for the matched image orientation $e_{\varphi}' = e_i$ corresponding
to the same unit vector in the unitary matrix $U$,
with the transformed
affine scale-normalized directional derivative operator in this
direction defined according to
\begin{equation}
  \label{eq-def-aff-sc-norm-ders-cov-prop-coupl-eigendecomp}
  \partial_{\varphi,'norm}^m
  = {s'}^{m/2} \, (e_{\varphi'}^T \, \Sigma' \, e_{\varphi'})^{m/2} \,
     \partial_{\varphi'}^m,
\end{equation}
then the relationship
\begin{equation}
  \label{eq-cov-prop-aff-sc-norm-dir-ders-coupl-eigen-decomp}
  \partial_{\varphi',\norm}^m L'(x';\; s', \Sigma')
  = \partial_{\varphi,\norm}^m L(x;\; s, \Sigma)
\end{equation}
will hold over the affine Gaussian scale-space representations
$L(x;\; s, \Sigma)$ and $L(x';\; s', \Sigma')$ of $f(x)$ and $f'(x')$, respectively,
for all values of the spatial scale parameter $s \in \bbbr_+$ and
the $2 \times 2$ spatial covariance matrix $\Sigma$ in the original domain,
provided that the spatial scale parameter $s' \in \bbbr_+$ and
the $2 \times 2$ spatial covariance matrix $\Sigma'$ in the transformed domain
are related according to
\begin{equation}
  s' \, \Sigma' = s \, U \, \diag(S_1^2 \, \lambda_1, S_2^2 \, \lambda_2) \, U^T.
\end{equation}

\subsubsection{Geometric interpretation of the analysis for
  the special case with coupled eigenvalue
  decompositions of the spatial covariance matrix $\Sigma$ and the
  affine transformation matrix $A$}

Interpreted geometrically, the above result from the analysis for the
special case with coupled eigenvalue
  decompositions has a special meaning, when
the spatial affine spatial transformations
constitute locally linearized projections from the tangent plane of
a local surface patch to the image domain, for different viewing
directions in relation to the local surface normal. Then, the special
form $x' = U \, \diag(S_1, S_2) \, U^T x$
of the image transformation corresponds to the
unitary matrix $U^T$ transforming the original coordinate frame to a
new coordinate frame,  where the spatial affine image transformation $A$
reduces to a pure non-uniform scaling transformation $\diag(S_1, S_2)$.
The preferred choice of image
orientation $e_{\varphi} = e_{\varphi'}$ will then specifically
correspond to either the
tilt direction%
\footnote{The tilt direction in a monocular projection model
  is the projection of the surface normal, at
the observation point, to the image plane. The slant angle is, in
turn, the angle between the surface normal and the viewing direction.}
or its perpendicular direction, and with the
variability spanned by varying the spatial scaling factor $S_i$
specifically corresponding to the tilt direction,
then corresponding to varying the slant angle
between the direction of the local surface normal at the
observation point and the viewing direction.

\begin{figure}[hbt]
  \begin{center}
    \includegraphics[width=0.35\textwidth]{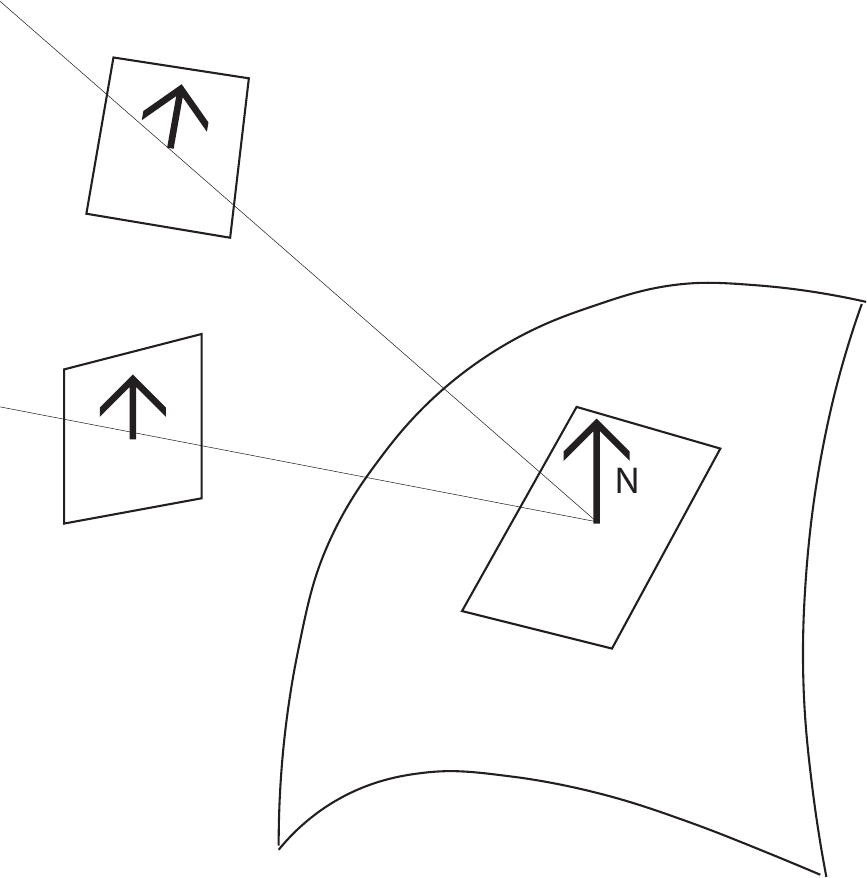}
  \end{center}  
  \caption{In terms of geometric interpretation, the covariance property
    (\ref{eq-cov-prop-aff-sc-norm-dir-ders-coupl-eigen-decomp})
    of affine scale-normalized derivatives according to
    (\ref{eq-def-aff-sc-norm-ders-cov-prop-coupl-eigendecomp}),
    in the case of coupled eigendecompositions of the spatial covariance
    matrix $\Sigma$ and the affine transformation matrix $A$ of the forms
    (\ref{eq-cov-mat-eigen-decomp-coupled-proof-2}) and
    (\ref{eq-transf-mat-eigen-decomp-coupled-proof-2}), means
    that if we consider a camera that views a smooth local surface patch, and
    then move the camera in the plane spanned by the backprojected tilt
    direction and the optical axis, then the affine scale-normalized
    directional derivative responses can, to first order of
    approximation, be perfectly matched between the resulting different
    views of the same local surface patch. In this way, we can thus, to
    first order of approximation, perfectly match the receptive field responses
    obtained for different slant angles relative to the viewing
    direction, provided that that one of the eigendirections of the
    spatial covariance matrix $\Sigma$, used for defining the spatial
    receptive fields, is parallel to the tilt
    direction, and that the relative motion of the observer between the
    different views also coincides with the backprojected tilt direction.}
  \label{fig-geom-int-cov-prop-aff-sc-norm-ders-coupl-eigendecomp}
\end{figure}

This derived covariance property does, thus, mean that the affine scale-normalized
directional derivative responses computed
in the tilt direction can, to
first order of approximation, be perfectly matched, when varying the slant
angle of the surface, as well as when varying the distance between the
object and the observer along the viewing direction;
see Figure~\ref{fig-geom-int-cov-prop-aff-sc-norm-ders-coupl-eigendecomp} for an illustration.

Concerning the dimensionality of the manifold spanned by this
covariance property, with the coupled eigenvalue decompositions of
the spatial covariance matrix $\Sigma$ and the affine transformation
matrix $A$ according to
(\ref{eq-cov-mat-eigen-decomp-coupled-proof-2})
and (\ref{eq-transf-mat-eigen-decomp-coupled-proof-2}),
the diagonal matrix $\diag(\lambda_1, \lambda_2)$ and the
diagonal matrix $\diag(S_1, S_2)$ do both each span variabilities of
dimensionality~2, to which the unitary matrix $U$ adds another variability
of dimensionality~1. The spatial scale parameter $s$ does not add any
effective dimensionality to the parameter space, because of its
coupled occurrence in the product $s \, \Sigma$. Similarly, the
direction $\varphi$ in the image plane does not add any dimensionality to the variability
of the resulting manifold either, since the unit vector $e_{\varphi}$
is determined from the unitary matrix $U$. Thus, we have that this
covariance result, under coupled eigenvalue decompositions of the
spatial covariance matrix $\Sigma$ and the affine transformation
matrix $A$, spans 5~out of the totally 8~dimensions in the variability
of computing affine scale-normalized derivatives $\partial_{\varphi,\norm}$
from an affine scale-space representation $L(x;\; s, \Sigma)$ in
different directions $\varphi$ in
image space, that are possible under the different types of image
transformations that are spanned by the full affine group,
as well as the different
parameter settings of the composed spatial covariance matrix
$s \, \Sigma$ that can be performed for the affine Gaussian
spatial smoothing component in the spatial receptive fields.

Out of these 5~dimensions, one of these dimensions,
corresponding to the ratio between the singular values of the affine
transformation matrix $A$, here manifested in terms of the ratio
$S_1/S_2$ between the two spatial scaling factors, is different, compared to the previously
treated covariance result under spatial similarity transformations.
In that respect, this covariance result provides important added value
in relation to the previously formulated covariance result under spatial
similarity transformations in Section~\ref{sec-aff-anal-dir-der-sim-group}.

%

\subsubsection{Lack of full affine-covariant properties
  for the affine scale-normalized
  directional derivative operators}
\label{app-lack-full-aff-cov-aff-scnorm-dir-der}

In this section, we will analyze the abilities of
the affine scale-normalized derivative operators according to
(\ref{eq-aff-sc-norm-dir-der})
\begin{equation}
  \label{eq-aff-sc-norm-dir-der-again}
  \partial_{\varphi,\norm}^m
  = s^{m/2} \, (e_{\varphi}^T \, \Sigma \, e_{\varphi})^{m/2} \, \partial_{\varphi}^m
\end{equation}
and
(\ref{eq-aff-sc-norm-dir-der-prim})
\begin{equation}
  \label{eq-aff-sc-norm-dir-der-prim-again}
  \partial_{\varphi',\norm}^m
  = {s'}^{m/2} \, (e_{\varphi'}^T \, \Sigma' \, e_{\varphi'})^{m/2} \, \partial_{\varphi'}^m
\end{equation}
in the matching image directions $\varphi$ and $\varphi'$, respectively,
according to the following relationship between their corresponding unit
vectors
(\ref{eq-match-unit-vectors-spat-aff})
\begin{equation}
  \label{eq-match-unit-vectors-spat-aff-again}
  e_{\varphi'} = \frac{A \, e_{\varphi}}{\| A \, e_{\varphi} \| },
\end{equation}
to allow 
for covariance under general affine transformations,
where $s \in \bbbr_+$ and $s' \in \bbbr_+$
as well as $\Sigma \in \bbbs_+^2$ and $\Sigma' \in \bbbs_+^2$
are matching scale parameters and spatial covariance matrices, respectively,
according to the relationship (\ref{eq-transf-prop-sc-par-spat-cov-mat-pure-aff-scsp})
\begin{equation}
  \label{eq-transf-prop-sc-par-spat-cov-mat-pure-aff-scsp-again}
  s' \Sigma' = s \, A \, \Sigma \, A^T.
\end{equation}
To perform this analysis, we will start from the intermediate results
in Equations~(\ref{eq-aff-sc-norm-dir-der-reduced}) and
(\ref{eq-aff-sc-norm-dir-der-reduced-prim}),
where we reduced the expressions for the affine
scale-normalized directional derivative operator
$\partial_{\varphi,\norm}^m$ along the direction $\varphi$ in the
original domain according to
(\ref{eq-aff-sc-norm-dir-der}) as well as
the directional derivative operator $\partial_{\varphi',\norm}^m$
along the direction $\varphi'$ in
the transformed domain according to
(\ref{eq-aff-sc-norm-dir-der-prim}) to the following forms:
\begin{align}
  \begin{split}
    \label{eq-aff-sc-norm-dir-der-reduced-app}
     \partial_{\varphi,\norm}^m
     & = s^{m/2} \, (e_{\varphi}^T \, \Sigma \, e_{\varphi})^{m/2} \, \partial_{\varphi}^m,
   \end{split}\\
  \begin{split}
    \label{eq-aff-sc-norm-dir-der-reduced-prim-app}    
       \partial_{\varphi',\norm}^m
       & = {s}^{m/2} \,
              \left(
                 \frac{e_{\varphi}^T \, A^T A \, \Sigma \,  A^T A \, e_{\varphi}}
                          {(e_{\varphi}^T \, A^T A \, e_{\varphi})^2}
              \right)^{m/2}
              \partial_{\varphi}^m,
     \end{split}
\end{align}
as defined over the affine Gaussian scale-space representations
$L \colon \bbbr^2 \times \bbbr_+ \times \bbbs_+^2 \rightarrow \bbbr$
and
$L' \colon \bbbr^2 \times \bbbr_+ \times \bbbs_+^2 \rightarrow \bbbr$, respectively,
obtained by convolution with affine Gaussian kernels
according to (\ref{eq-def-aff-scsp}) of two 2-D images
$f \colon \bbbr^2 \rightarrow \bbbr$ and $f' \colon \bbbr^2
\rightarrow \bbbr$, respectively,
that are related according to a spatial affine transformation
of the form (\ref{eq-spat-aff-transf-def-aff-sc-norm-ders})
\begin{equation}
  \label{eq-spat-aff-transf-def-aff-sc-norm-ders-again}
  f'(x') = f(x) \quad\quad\mbox{for}\quad\quad x' = A \, x,
\end{equation}
where $\Sigma$ denotes any $2 \times 2$ symmetric and positive definite
covariance matrix and $A$ denotes any non-singular $2 \times 2$ affine transformation matrix.

From the expressions in
Equations~(\ref{eq-aff-sc-norm-dir-der-reduced-app}) and
(\ref{eq-aff-sc-norm-dir-der-reduced-prim-app}),
we can therefore see that a both necessary and
sufficient condition, for these affine scale-normalized derivatives to
be fully affine covariant, such that
\begin{equation}
  \partial_{\varphi',\norm}^m L'(x';\; s', \Sigma')
  = \partial_{\varphi,\norm}^m L(x;\; s, \Sigma)
\end{equation}
would hold for all matching image points $x' = A \, x$ according to
(\ref{eq-spat-aff-transf-def-aff-sc-norm-ders}), for all matching
parameters of the receptive fields $s' \Sigma' = s \, A \, \Sigma \, A^T$
according to (\ref{eq-transf-prop-sc-par-spat-cov-mat-pure-aff-scsp}),
for all matching unit vectors
$e_{\varphi'} = e_{\varphi}/\| A e_{\varphi} \|$ as well as for all orders $m$ of
spatial differentiation, would be that
the relationship
\begin{equation}
  \label{eq-cond-aff-cov-general-in-matrices}
    \frac{e_{\varphi}^T \, A^T A \, \Sigma \,  A^T A \, e_{\varphi}}
            {(e_{\varphi}^T \, A^T A \, e_{\varphi})^2}
  = e_{\varphi}^T \, \Sigma \, e_{\varphi}
\end{equation}
would hold for all $2 \times 2$ affine transformation matrices
$A$, for all $2 \times 2$ symmetric and positive definite matrices
$\Sigma$, as well as for
all 2-D unit vectors $e_{\varphi}$.

In the following, we will explicitly
show that such a relationship does not, however, hold generally, although that
we have in Sections~\ref{sec-aff-anal-dir-der-sim-group}
and~\ref{sec-aff-anal-dir-der-coupled-eigen-decomp} previously
shown that such a relationship holds for two, for our purposes very important,
subgroups of the full affine group.

Since both the matrices $\Sigma$ and $A^T A$ are symmetric and
positive definite, let us start by replacing these matrices with their
eigenvalue decompositions
\begin{align}
  \begin{split}
     \label{eq-Sigma-factorization-proof}
     \Sigma = U \Lambda \, U^T,
  \end{split}\\
  \begin{split}
    \label{eq-ATA-factorization-proof}
     A^T A = V D \, V^T,
  \end{split}
\end{align}
where $U$ and $V$ are real unitary $2 \times 2$  matrices,
and $\Lambda$ and $D$ are $2 \times 2$ diagonal matrices with strictly
positive elements.
Then, the question of whether the relation
(\ref{eq-cond-aff-cov-general-in-matrices}) holds for all non-singular
$2 \times 2$ affine
transformation matrices $A$, all $2 \times 2$ affine covariance matrices $\Sigma$
and all 2-D unit vectors $e_{\varphi}$, or not, can
reformulated into investigating whether
\begin{equation}
  \frac{e_{\varphi}^T \, V \, D \, V^T U  \Lambda \, U^T V \,
    D \, V^T e_{\varphi}}{(e_{\varphi}^T \, V D \, V^T e_{\varphi})^2}
  = e_{\varphi}^T \, U \Lambda \, U^T e_{\varphi}
\end{equation}
would hold for all $2 \times 2$ unitary matrices $U$ and $V$, all $2
\times 2$ diagonal matrices
$\Lambda$ and $D$ and all 2-D unit vectors $e_{\varphi}$, or not.
By, in turn, setting
\begin{align}
  \begin{split}
    V^T e_{\varphi} = e_{\psi},
  \end{split}\\
  \begin{split}
    U^T V = W,
  \end{split}
\end{align}
which gives
\begin{align}
  \begin{split}
    e_{\varphi} = V^{-T} e_{\psi},
  \end{split}\\
  \begin{split}
    U= V^{-T} W^T,
  \end{split}
\end{align}
this expression can after a few simplifications be reformulated into
investigating whether the expression
\begin{equation}
  \frac{e_{\psi}^T \, D \, W^T \Lambda \, W \, D \, e_{\psi}}
          {(e_{\psi}^T \, D \, e_{\psi})^2}
  = e_{\psi}^T \, W^T \Lambda \, W \, e_{\psi}
\end{equation}
would hold for all $2 \times 2$ unitary vectors $W$, all $2 \times 2$ diagonal matrices $\Lambda$ and
$D$ and all 2-D unit vectors $e_{\psi}$, or not. By further setting
\begin{equation}
   C = W^T \Lambda \, W,
 \end{equation}
where $C$ then will become an arbitrary $2 \times 2$ symmetric and positive
definite matrix, we thus have
that the necessary and sufficient relationship for
affine covariance (\ref{eq-cond-aff-cov-general-in-matrices}) can then
be reformulated as whether the relationship
\begin{equation}
  \label{eq-cond-aff-cov-general-in-matrices-transformed}
  e_{\psi}^T \, D \, C \, D \, e_{\psi}
  = (e_{\psi}^T \, C \, e_{\psi}) \, (e_{\psi}^T \, D \, e_{\psi})^2
\end{equation}
would hold for all symmetric and positive semi-definite $2 \times 2$ matrices $C$, all
$2 \times 2$ diagonal matrices $D$ and all 2-D unit vectors $e_{\psi}$.

By further parameterizing these entities as
\begin{align}
  \begin{split}
    C = \left(
              \begin{array}{cc}
                c_{11} & c_{12} \\
                c_{12} & c_{22}
              \end{array}
           \right),
  \end{split}\\
  \begin{split}
    D = \left(
              \begin{array}{cc}
                d_1 & 0 \\
                0     & d_2
              \end{array}
           \right),
  \end{split}\\
  \begin{split}
    e_{\psi} =
           \left(
              \begin{array}{c}
                \cos \psi \\
                \sin \psi
               \end{array}
           \right),
  \end{split}                     
\end{align}
and then expanding
(\ref{eq-cond-aff-cov-general-in-matrices-transformed}) with respect
to this parameterization,
we can thus reduce the problem of investigating whether the necessary
and sufficient condition for affine covariance
(\ref{eq-cond-aff-cov-general-in-matrices}) would hold, to
investigating whether the following equation would hold for all
combinations of $c_{ij} \in \bbbr$, $d_k \in \bbbr_+$ and $\psi \in \bbbr$:
\begin{align}
 \begin{split}
   c_{11} \, \cos^2 \psi
   & \left(d_1^2 \left(1 -\cos ^4\psi\right)
   \right.
  \end{split}\nonumber\\
  \begin{split}
    & \left.
      \quad -2 \, d_1 \, d_2 \, \sin ^2\psi \, \cos ^2\psi
  \right.
  \end{split}\nonumber\\
  \begin{split}
    & \left.
      \quad -d_2^2 \, \sin ^4\psi \right)
  \end{split}\nonumber\\
  \begin{split}
    +c_{12} \, \cos \psi \, \sin \psi
    & \left(-2 \, d_1^2 \, \cos ^4\psi
  \right.
  \end{split}\nonumber\\
  \begin{split}
    & \left.
      \quad +d_1 d_2 \left(2 -4 \, \sin ^2\psi \, \cos ^2\psi\right)
  \right.
  \end{split}\nonumber\\
  \begin{split}
    & \left.
      \quad -2 \, d_2^2 \, \sin ^4\psi \right)
  \end{split}\nonumber\\
  \begin{split}
    +c_{22} \, \sin^2 \psi \, 
    & \left(-d_1^2 \, \cos ^4\psi
  \right.
  \end{split}\nonumber\\
  \begin{split}
    & \left.
      \quad -2 \, d_1 \, d_2 \, \sin ^2 \psi \, \cos ^2 \psi 
  \right.
  \end{split}\nonumber\\
  \begin{split}
    & \left.
      \quad + d_2^2 \left(1 -\sin ^4\psi\right)\right)
  = 0.
  \end{split}
\end{align}
Disregarding the singular cases when either $\cos \psi = 0$ or
$\sin \psi = 0$, if
this expression is to hold for all combinations of the parameters
$c_{ij}$, $d_k$ and $\psi$,
then this specifically implies that the coefficients for each one of the matrix elements
$c_{ij}$ must be zero, implying that the following relations must hold
for all combinations of $d_k \in \bbbr_+$ and $\psi \in \bbbr$:
\begin{align}
  \begin{split}
    \label{eq-red-rel-c11}
    d_1^2 \left(1 -\cos ^4\psi\right)
    -2 \, d_1 \, d_2 \, \sin ^2\psi \, \cos ^2\psi
    - d_2^2 \, \sin ^4\psi  = 0,
  \end{split}\\
  \begin{split}
    -2 \, d_1^2 \, \cos ^4\psi
    +d_1 d_2 \left(2 -4 \sin ^2\psi \, \cos ^2\psi\right)
    -2 \, d_2^2 \, \sin ^4\psi = 0,
  \end{split}\\
  \begin{split}
    \label{eq-red-rel-c22}    
    -d_1^2 \, \cos ^4\psi
    -2 \, d_1 \, d_2 \, \sin ^2 \psi \, \cos ^2 \psi
    +d_2^2 \left(1 -\sin ^4\psi\right) = 0.
  \end{split}
\end{align}
Subtracting Equation~(\ref{eq-red-rel-c22}) from
Equation~(\ref{eq-red-rel-c11}),
then leads to the following necessary condition for the affine
scale-normalized directional derivatives
(\ref{eq-aff-sc-norm-dir-der-reduced-app}) and
(\ref{eq-aff-sc-norm-dir-der-reduced-prim-app}) to be equal:
\begin{equation}
  d_1^2 - d_2^2 = 0.
\end{equation}
Given the restriction that the elements $d_k$ of the
diagonal matrix $D = \diag(d_1, d_2)$
are to be non-negative, we thus have that the requirement, for the
affine scale-normal\-ized
directional derivative operators according to
(\ref{eq-aff-sc-norm-dir-der-reduced-app}) and
(\ref{eq-aff-sc-norm-dir-der-reduced-prim-app}) to be equal,
implies that the affine covariance matrix
$\Sigma = U \Lambda \, U^T$ according to
(\ref{eq-Sigma-factorization-proof}) must be an isotropic matrix,
and that the matrix $A^T A = V D \, V^T$ formed from the affine
transformation matrix $A$ according to
(\ref{eq-ATA-factorization-proof}) must also be an isotropic matrix,
thus, in turn, implying that the affine transformation matrix $A$ must be in the
similarity group, corresponding to $A = S_x \, R$, where $S_x$ is a
uniform spatial scaling factor and $R$ is a rotation matrix.

In this way, we have formally shown that the affine scale-normalized
directional derivative operators according to
(\ref{eq-aff-sc-norm-dir-der-again}) and
(\ref{eq-aff-sc-norm-dir-der-prim-again})
do not allow for
covariance properties under general affine transformations.

\medskip

\noindent
{\bf Summary of main result:}
The
affine scale-normalized derivative operators according to
(\ref{eq-aff-sc-norm-dir-der-again}) and
(\ref{eq-aff-sc-norm-dir-der-prim-again}) 
do not allow
for covariance under general affine transformations.

\medskip
\noindent
Thus, if we want to aim at full affine covariance, we have to consider
some other definition of a scale-normalized derivative concept based
on the affine scale-space representation, which we will now develop
in the next section.

\subsection{Scale-normalized affine gradient operator for an affine scale-space representation based on smoothing with anisotropic affine Gaussian kernels}
\label{sec-sc-norm-aff-grad-op}

In this section, we will define another new type of scale-normalized
spatial derivatives for an affine Gaussian scale-space representation,
which, in contrast to the previous definition of affine
scale-normalized derivatives, will, however, lead to true covariance
covariance properties over the full group of spatial affine transformations.

Given any 2-D image $f \colon \bbbr^2 \rightarrow \bbbr$,
let us again consider its affine Gaussian scale-space representation
$L \colon \bbbr^2 \times \bbbr_+ \times \bbbs_+^2 \rightarrow \bbbr$,
of the
form
\begin{equation}
  \label{eq-def-aff-scsp-again}
  L(\cdot;\; s, \Sigma) = g(\cdot;\; s, \Sigma) * f(\cdot),
\end{equation}
generated by convolutions with anisotropic affine Gaussian kernels
(\ref{eq-gauss-fcn-2D}), based on $2 \times 2$ spatial
covariance matrices $\Sigma \in \bbbs_+^2$ that are not generally equal to a unit matrix.

\subsubsection{Specialized definition of the square root of the
  covariance matrix $\Sigma$}

Given an eigenvalue decomposition of the $2 \times 2$ symmetric and positive definite spatial
covariance matrix $\Sigma$ of the form
\begin{equation}
  \label{eq-eigen-decomp-Sigma-aff-grad}
  \Sigma
  = U \Lambda \, U^T,
\end{equation}
where $\Lambda = \diag(\lambda_1, \lambda_2)$ is a $2 \times 2$ diagonal matrix
with positive elements, and $U$ is a $2 \times 2$  real unitary matrix, let us
first define the square root of the diagonal matrix $\Lambda$ as
$\Lambda^{1/2} = \diag(\lambda_1^{1/2}, \lambda_2^{1/2})$,
to rewrite (\ref{eq-eigen-decomp-Sigma-aff-grad}) as
\begin{equation}
  \Sigma
  = U \, \Lambda^{1/2} \,  \Lambda^{1/2}\, U^T
  = (U \, \Lambda^{1/2}) \,  (U \, \Lambda^{1/2})^T.
\end{equation}
From this expression, let us then define the square root
$\Sigma^{1/2}$ of $\Sigma$ as
\begin{equation}
  \label{eq-def-sqrt-of-Sigma}
  \Sigma^{1/2} = \Lambda^{1/2}\, U^T,
\end{equation}
such that
\begin{equation}
  \label{eq-Sigma-from-sqrt}
  \Sigma = (\Sigma^{1/2})^T (\Sigma^{1/2}).
\end{equation}
Note, however, that this definition is not unique. For a
general square root of a matrix, also the matrix
\begin{equation}
  \label{eq-general-sq-root-mat}
  \Sigma^{1/2} = \rho \, \Lambda^{1/2}\, U^T,
\end{equation}
where $\rho$ is an arbitrary $2 \times 2$ unitary matrix, would also%
\footnote{To understand the origin of the indeterminacy of this form,
  consider first a singular-value-type decomposition of a general
  square root matrix $\Sigma^{1/2}$ of the covariance matrix $\Sigma$
  of the form $\Sigma^{1/2} = U_{1/2} \, \Lambda^{1/2} \, V_{1/2}^T$, where
  $\Lambda^{1/2}$ is a diagonal matrix with not necessarily ordered
  eigenvalues. With regard to the expression
  (\ref{eq-general-sq-root-mat}), this corresponds to
  setting $U_{1/2} = \rho$ and $V_{1/2}^T = U^T$. For our convention for the
  square root matrix $\Sigma^{1/2}$ according to
  (\ref{eq-def-sqrt-of-Sigma}), the chosen form of the principal square
  root thus corresponds to choosing $U_{1/2} = I$.}
satisfy
(\ref{eq-Sigma-from-sqrt}), since then
\begin{equation}
  (\Sigma^{1/2})^T (\Sigma^{1/2})
  = U \, \Lambda^{1/2} \, \rho^T \, \rho \, \Lambda^{1/2} \, U^T
  = U \Lambda \, U^T.
\end{equation}

\subsubsection{Definition of the scale-normalized affine gradient
  operator}

Given the above specially chosen definition of the square root
$\Sigma^{1/2}$ of the spatial covariance matrix $\Sigma$, and given that
we have applied the regular spatial gradient operator $\nabla_x$ to the
affine Gaussian scale-space representation $L(x;\; s, \Sigma)$ for the 
spatial scale parameter $s \in \bbbr_+$ and the $2 \times 2$ spatial
covariance matrix $\Sigma$, let us define the scale-normalized affine
gradient operator as
\begin{equation}
  \label{eq-def-sc-norm-aff-grad-op}
  \nabla_{x,\affnorm} = s^{1/2} \, \Sigma^{1/2} \, \nabla_x.
\end{equation}
The motivation for this definition of the scale-normalized directional
derivative operator is to, as for the
previous definition of the affine scale-normalized directional
derivative operator $\partial_{\varphi,\norm}$ according to
(\ref{eq-aff-sc-norm-dir-der}), first of all
compensate for the general decrease in the magnitude of spatial
derivatives with increasing amount of spatial smoothing, and
specifically also take into account the different amounts of spatial smoothing in
the different spatial directions in the image plane,
as resulting from using anisotropic affine
Gaussian kernels $g(x;\; s, \Sigma)$, as opposed to rotationally symmetric Gaussian
kernels $g(x;\; s, I)$ for the spatial smoothing operation in the
spatial receptive fields.

In contrast to the previous definition of the affine scale-normalized directional
derivative operator $\partial_{\varphi,\norm}$ according to (\ref{eq-aff-sc-norm-dir-der}),
which only considers a single orientation in the image plane, and then
takes into account the amount of spatial smoothing in that 
direction, for which the directional derivative is computed,
the definition of the affine scale-normalized gradient operator
$\nabla_{x,\affnorm}$ according to (\ref{eq-def-sc-norm-aff-grad-op}) does instead
consider the genuine 2-D image gradient as the conceptual object, and
does then also take into account all the information about the spatial
covariance matrix $\Sigma$, when defining the corresponding
scale-normalized object. As we will see in the next section, this new
definition will therefore allow for full affine covariance, as opposed
to covariance properties over smaller specific subgroups of the affine
group, as obtained for the previously defined affine scale-normalized directional
derivative operator $\partial_{\varphi,\norm}$ according to (\ref{eq-aff-sc-norm-dir-der}).

Before turning into the details of the derivation of the general full
affine covariance property, let us, however, note that in the special
case when the spatial covariance matrix $\Sigma$ is a diagonal matrix
$\Sigma = \diag(\lambda_1, \lambda_2)$, we obtain
$\Sigma^{1/2} = \diag(\lambda_1^{1/2}, \lambda_2^{1/2})$,
which implies that the definition of the scale-normalized affine
gradient operator (\ref{eq-def-sc-norm-aff-grad-op}) then reduces to the relation
\begin{equation}
  \left(
     \begin{array}{c}
        \partial_{x_1,\affnorm} \\
        \partial_{x_2,\affnorm}
      \end{array}
   \right)
   =
  \left(
    \begin{array}{c}
       s^{1/2} \, \lambda_1^{1/2} \, \partial_{x_1} \\
       s^{1/2} \, \lambda_2^{1/2} \, \partial_{x_2}    
      \end{array}
   \right),
\end{equation}
which is the same result as we then obtain for the previously defined
affine scale-normalized directional derivative operator $\partial_{\varphi,\norm}$ according to
(\ref{eq-aff-sc-norm-dir-der}), if we choose the directions $\varphi$ for
computing the affine scale-normalized directional derivatives
as the two coordinate directions $e_1$ for $\varphi_1 = 0$ and
$e_2$ for $\varphi_2 = \pi/2$, for the spatial differentiation order
$m = 1$,
such that
\begin{align}
  \begin{split}
    \partial_{\varphi_1,\norm}
    = s^{1/2} \, \lambda_1^{1/2} \, \partial_{\varphi_1}
    = s^{1/2} \, \lambda_1^{1/2} \, \partial_{x_1},
  \end{split}\\
   \begin{split}
    \partial_{\varphi_2,\norm}
    = s^{1/2} \, \lambda_2^{1/2} \, \partial_{\varphi_2}
    = s^{1/2} \, \lambda_2^{1/2} \, \partial_{x_2}.
  \end{split}
\end{align}
Thus, in this very special case of the spatial covariance matrix
$\Sigma$ being a diagonal matrix $\diag(\lambda_1, \lambda_2)$, 
and then choosing the directions for the directional derivative
operators along the coordinate directions, the definition of the
scale-normalized affine gradient operator $\nabla_{x,\affnorm}$ according to
(\ref{eq-def-sc-norm-aff-grad-op}) is consistent with the previous
definition of the affine scale-normal\-ized directional derivative
operator $\partial_{\varphi,\norm}$ according to (\ref{eq-aff-sc-norm-dir-der}).

Furthermore, in the case when the spatial covariance matrix $\Sigma$
is a unit matrix $I$, the
scale-normalized affine gradient operator $\nabla_{x,\affnorm}$ according to
(\ref{eq-def-sc-norm-aff-grad-op}) then
reduces to the isotropic scale-normalized
gradient operator $\nabla_{x,\norm}$  according to (\ref{eq-nabla-op}).

\subsection{Full covariance property of the scale-normalized affine
  gradient operator under general spatial affine transformations}
\label{sec-cov-prop-sc-norm-aff-grad-op}

Consider a spatial affine transformation of the form
\begin{equation}
  \label{eq-spat-aff-transf-def-aff-sc-norm-grad-op}
  f'(x') = f(x) \quad\quad\mbox{for}\quad\quad x' = S_x \, A \, x,
\end{equation}
where $S_x \in \bbbr_+$ is a spatial scaling factor,
$A$ is a non-singular $2 \times 2$ affine transformation matrix,
from which we define the respective affine Gaussian scale-space
representations
$L \colon \bbbr^2 \times \bbbr_+ \times \bbbs_+^2 \rightarrow \bbbr$
and
$L' \colon \bbbr^2 \times \bbbr_+ \times \bbbs_+^2 \rightarrow \bbbr$
according to
\begin{align}
  \begin{split}
    \label{eq-aff-scsp-def-sc-norm-aff-grad}
    L(\cdot;\; s, \Sigma) = g(\cdot;\; s, \Sigma)* f(\cdot),
  \end{split}\\
  \begin{split}
    \label{eq-aff-scsp-def-sc-norm-aff-grad-prim}
    L'(\cdot;\; s', \Sigma') = g(\cdot;\; s', \Sigma')* f'(\cdot),
  \end{split}
\end{align}
for matching values of the spatial scale parameters
$s \in \bbbr_+$ and $s' \in \bbbr_+$ as
well as the $2 \times 2$ spatial covariance matrices
$\Sigma \in \bbbs_+^2$ and $\Sigma' \in \bbbs_+^2$ over
the two domains, such that (\ref{eq-transf-prop-sc-par-spat-cov-mat-pure-aff-scsp})
\begin{equation}
  \label{eq-transf-prop-sc-par-spat-cov-mat-pure-aff-scsp-full}
  s' \, \Sigma' = s \, (S_x \, A) \, \Sigma \, (S_x A)^T = s \, S_x^2 \, A \, \Sigma \, A^T,
\end{equation}
which then implies that the affine scale-space representations for
these parameter values of the affine Gaussian smoothing kernels are
equal (\ref{eq-equal-aff-scsp-repr-def-aff-sc-norm-ders})
\begin{equation}
  \label{eq-equal-aff-scsp-repr-aff-cov-proof}
  L'(x';\; s', \Sigma') = L(x;\; s, \Sigma).
\end{equation}
Given an eigenvalue decomposition of the spatial covariance matrix
$\Sigma'$ in the transformed domain
\begin{equation}
   \Sigma' = U' \Lambda' \, {U'}^T,
 \end{equation}
where $U'$ is a $2 \times 2$ unitary matrix and
$\Lambda'$ is a $2 \times 2$ diagonal matrix with positive elements,
let us in a similar way as in (\ref{eq-general-sq-root-mat}) 
define the square root of $\Sigma'$ as
\begin{equation}
  \label{eq-def-sqrt-of-Sigma-prim}
  {\Sigma'}^{1/2} = {\Lambda'}^{1/2}\, {U'}^T,
\end{equation}
while noting that any other definition of the square root of $\Sigma'$  according to
\begin{equation}
  \label{eq-def-sqrt-of-Sigma-prim-perm}
  {\Sigma'}^{1/2} = \rho' \, {\Lambda'}^{1/2}\, {U'}^T,
\end{equation}
where $\rho'$ is an arbitrary $2 \times 2$ rotation matrix, would also satisfy
\begin{equation}
   \Sigma' = ( {\Sigma'}^{1/2})^T  ({\Sigma'}^{1/2}).
\end{equation}
Inserting this expression, as well as
\begin{equation}
   \Sigma = ( \Sigma^{1/2})^T  (\Sigma^{1/2})
\end{equation}
according to (\ref{eq-Sigma-from-sqrt}), into the coupled relationship
(\ref{eq-transf-prop-sc-par-spat-cov-mat-pure-aff-scsp-full}) between the
spatial scale parameters $s$ and $s'$ as well as the spatial
covariance matrices $\Sigma$ and $\Sigma'$,
then, with the added degree
of freedom corresponding to different possible rotation matrices in 
(\ref{eq-general-sq-root-mat}) and
(\ref{eq-def-sqrt-of-Sigma-prim-perm}), gives
\begin{multline}
  s' \, ({\Sigma'}^{1/2})^T \, {\rho'}^T \, {\rho'} \,  ({\Sigma'}^{1/2}) = \\
   = s \, S_x^2 \, A \, (\Sigma^{1/2})^T \, \rho^T \, \rho \, (\Sigma^{1/2}) \, A^T.
\end{multline}
This relationship does then imply that square roots $\Sigma^{1/2}$ and
${\Sigma'}^{1/2}$ of the spatial
covariance matrices $\Sigma$ and $\Sigma'$ must be related according
to
\begin{equation}
   {s'}^{1/2} \, \rho' \, {\Sigma'}^{1/2} =  s^{1/2} \, S_x \, \rho \, \Sigma^{1/2} \, A^T,
 \end{equation}
which in turn implies that the following relationship must hold for
some, possibly other, $2 \times 2$ rotation matrix $\tilde{\rho} = {\rho'}^T \rho$:
\begin{equation}
  \label{eq-rel-sqrt-cov-mat-aff-cov-proof}
   {s'}^{1/2} \, {\Sigma'}^{1/2} =  \tilde{\rho} \, s^{1/2} \, S_x \, \Sigma^{1/2} \, A^T.
\end{equation}
Under the image transformation
(\ref{eq-spat-aff-transf-def-aff-sc-norm-grad-op}),
the spatial gradient operators are
related according to
\begin{equation}
  \label{eq-nabla-x-transf-aff-cov-proof}
  \nabla_x = (S_x \, A)^T \, \nabla_{x'},
\end{equation}
implying that
\begin{equation}
  \label{eq-nabla-x-transf-aff-cov-inv-proof}  
  \nabla_{x'} = \frac{1}{S_x} \times A^{-T} \nabla_x.
\end{equation}
Inserting this expression (\ref{eq-nabla-x-transf-aff-cov-inv-proof}), as
well as the relationship (\ref{eq-rel-sqrt-cov-mat-aff-cov-proof})
between the square roots of the spatial covariance matrices between
the two domains, into the corresponding definition of the
scale-normalized affine gradient operator over the transformed doman
\begin{equation}
   \nabla_{x',\affnorm} = \tilde{\rho'} \, {s'}^{1/2} \, {\Sigma'}^{1/2} \, \nabla_{x'},
\end{equation}
then implies that the expression for the
scale-normalized affine gradient operator over the transformed domain
reduces to
\begin{equation}
   \nabla_{x',\affnorm} = \tilde{\rho} \, s^{1/2} \, \Sigma^{1/2} \, \nabla_x.
\end{equation}
After comparison with the scale-normalized affine
gradient operator over the original domain
(\ref{eq-def-sc-norm-aff-grad-op}), it therefore holds that
the scale-normalized affine gradient operators over the two different
domains must be related according to
\begin{equation}
   \nabla_{x',\affnorm} = \tilde{\rho} \, \nabla_{x,\affnorm},
\end{equation}
for some rotation matrix $\tilde{\rho}$. Thus, by, in turn, applying these
operators to the affine scale-space representations $L(x;\; s, \Sigma)$
and $L'(x';\; s', \Sigma')$ over their respective domains,
for matching values of the parameter
values of the affine Gaussian smoothing kernel according to
(\ref{eq-transf-prop-sc-par-spat-cov-mat-pure-aff-scsp-full}),
this implies that
the scale-normalized affine gradient vectors
$\nabla_{x,\affnorm} L$
and $\nabla_{x',\affnorm}  L'$ over the two domains must be
related according to
\begin{equation}
  \label{eq-equal-aff-scsp-repr-aff-cov-proof-again}
  (\nabla_{x',\affnorm}  L')(x';\; s', \Sigma') =
  \tilde{\rho} \,  (\nabla_{x,\affnorm} L)(x;\; s, \Sigma)
\end{equation}
for some rotation matrix $\tilde{\rho}$.
Specifically, if the affine transformation $A$ is in the similarity
group, {\em i.e.\/} if $A = S_x \, R$ for some positive scaling
factor $S_x \in {\mathbb R}$ and some rotation matrix $R$,
then the rotation matrix $\tilde{\rho}$
in (\ref{eq-equal-aff-scsp-repr-aff-cov-proof-again})
can be shown to be
restricted to a unit matrix $\tilde{\rho} = I$.

To realize why the rotation matrix $\tilde{\rho}$ reduces to a unit
matrix for the case of similarity transformations, let us insert
$A = S_x \, R$ into the relationship
(\ref{eq-transf-prop-sc-par-spat-cov-mat-pure-aff-scsp-full}),
which then gives 
\begin{equation}
  \label{eq-transf-prop-sc-par-spat-cov-mat-pure-aff-scsp-full-sim}
  s' \, \Sigma' = s \, (S_x \, R) \, \Sigma \, (S_x R)^T = s \, S_x^2 \, R \, \Sigma \, R^T
\end{equation}
and into (\ref{eq-nabla-x-transf-aff-cov-inv-proof}), which gives 
\begin{equation}
  \label{eq-nabla-x-transf-aff-cov-inv-proof}  
  \nabla_{x'} L' = \frac{1}{S_x} \times R^{-T} \nabla_x L.
\end{equation}
From the definition of the scale-normalized affine gradient vector of
the transformed domain
\begin{equation}
  \nabla_{x',\affnorm} L' = {s'}^{1/2} \, {\Sigma'}^{1/2} \, \nabla_{x'} L',
\end{equation}
and inserting $\Sigma = U \Lambda \, U^T$ into
(\ref{eq-transf-prop-sc-par-spat-cov-mat-pure-aff-scsp-full-sim})
\begin{equation}
  \Sigma'  = \frac{s}{s'} \times S_x^2 \, R \, U \Lambda \, U^T R^T,
\end{equation}
which gives
\begin{equation}
  {\Sigma'}^{1/2} = \left( \frac{s}{s'} \right)^{1/2} S_x \, \Lambda^{1/2} \, U^T R^T,
\end{equation}
we then obtain
\begin{multline}
  \nabla_{x',\affnorm} L'
  = {s'}^{1/2} \, \left( \frac{s}{s'} \right)^{1/2}  \, \Lambda^{1/2} \, U^T
  R^T  R^{-T} \nabla_x L = \\
  = s^{1/2} \, \Sigma^{1/2} \, \nabla_x L = \nabla_{x,\affnorm} L,
\end{multline}
thus showing that $\tilde{\rho}$ in this case reduces to a unit matrix
$I$.

\medskip

\noindent
{\bf Summary of main result.} To summarize, this result shows that, if
we under an arbitrary non-singular spatial affine transformation of the form
\begin{equation}
  \label{eq-spat-aff-transf-def-aff-sc-norm-ders-main-result}
  f'(x') = f(x) \quad\quad\mbox{for}\quad\quad x' = S_x A \, x,
\end{equation}
define the affine Gaussian scale-space representations
$L(x;\; s, \Sigma)$ and $L'(x';\; s', \Sigma')$ of
the images
$f$ and $f'$, respectively,
according to (\ref{eq-aff-scsp-def-sc-norm-aff-grad}) and
(\ref{eq-aff-scsp-def-sc-norm-aff-grad-prim}),
with the spatial scale parameters
$s \in \bbbr_+$ and $s' \in \bbbr_+$ as
well as the $2 \times 2$ spatial covariance matrices $\Sigma$
and $\Sigma'$, respectively, and then
define the scale-normalized affine gradient operator over the original
domain as
\begin{equation}
  \label{eq-def-sc-norm-aff-grad-op-aff-cov-main-result}
  \nabla_{x,\affnorm} = s^{1/2} \, \Sigma^{1/2} \, \nabla_x,
\end{equation}
as well as define the corresponding scale-normalized affine gradient operator
over the transformed domain as
\begin{equation}
  \label{eq-def-sc-norm-aff-grad-op-aff-cov-prim-main-result}
  \nabla_{x',\affnorm} = {s'}^{1/2} \, {\Sigma'}^{1/2} \, \nabla_{x'},
\end{equation}
then the corresponding scale-normalized affine gradient vectors over
the two domains will, up to some rotation matrix $\rho$, be equal, such that
\begin{equation}
  \label{eq-equal-aff-scsp-repr-aff-cov-proof-again-again}
  (\nabla_{x',\affnorm}  L')(x';\; s', \Sigma') =
  \tilde{\rho} \,  (\nabla_{x,\affnorm} L)(x;\; s, \Sigma)
\end{equation}
holds for some rotation matrix $\tilde{\rho}$,  provided that the
parameters of the underlying affine Gaussian
smoothing kernels are related according to
\begin{equation}
  \label{eq-transf-prop-sc-par-spat-cov-mat-pure-aff-scsp-main-result}
  s' \, \Sigma' = s \, (S_x \, A) \, \Sigma \, (S_x \, A)^T = s \, S_x^2 \, A \, \Sigma \, A^T.
\end{equation}
In the case of similarity transformations $A = S_x \, R$, the
rotation matrix in
(\ref{eq-equal-aff-scsp-repr-aff-cov-proof-again-again})
does specifically degenerate to an identity matrix $\tilde{\rho} = I$.

\begin{figure}[hbt]
 \begin{center}
    \includegraphics[width=0.40\textwidth]{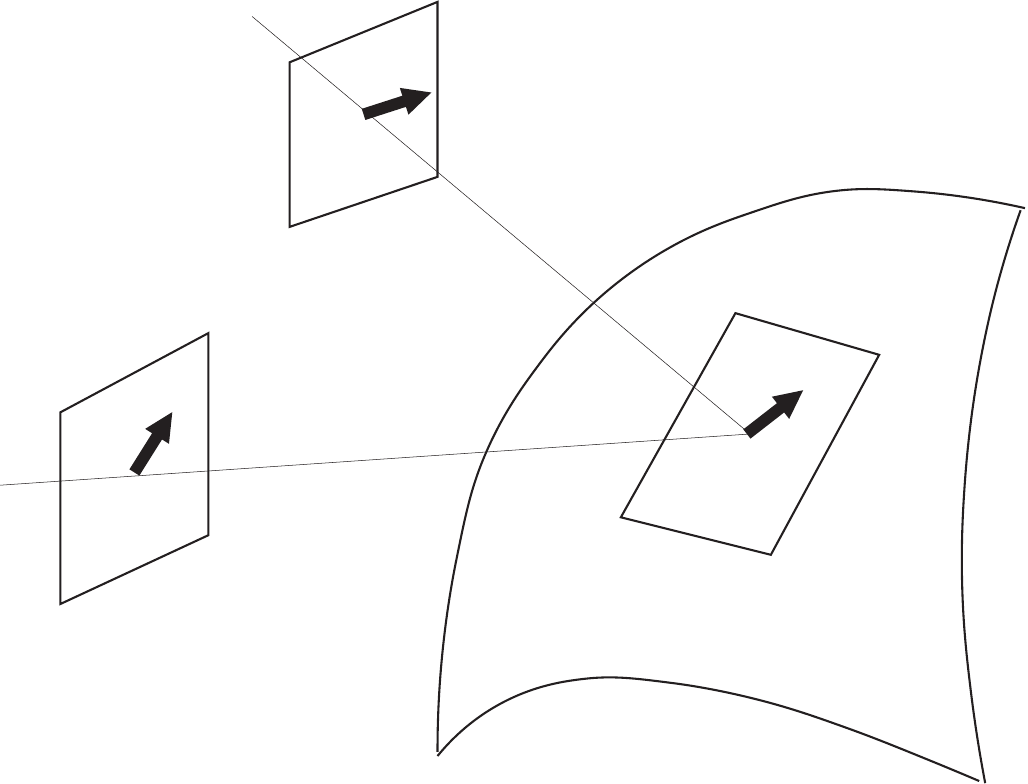}
  \end{center}
  \caption{The covariance property
    (\ref{eq-equal-aff-scsp-repr-aff-cov-proof-again-again}) of the
    scale-normalized affine gradient operator
    (\ref{eq-def-sc-norm-aff-grad-op-aff-cov-main-result}) under
    general (non-singular) affine transformations means that, if we
    consider two cameras, that view the same local surface patch from
    general (non-degenerate) viewing conditions, then, to first order
    of approximation, the resulting affine gradient responses for the
    different views, here illustrated as arrows before the affine
    scale normalization, can, up
    to a rotation transformation $\tilde{\rho}$, be perfectly
    matched, provided that the scale parameters and the covariance
    matrices of the receptive fields are properly matched according to
    (\ref{eq-transf-prop-sc-par-spat-cov-mat-pure-aff-scsp-main-result}).}
  \label{fig-geom-int-cov-prop-sc-norm-aff-grad}
\end{figure}

Interpreted geometrically, this result means that if we interpret the
composed affine transformation $S_x \, A$ as constituting a local
linearization of the perspective mapping from the tangent plane of a
local surface patch, or as a local linearization of the projective mapping
between different views of the same local surface patch,
then this result means that the scale-normalized
affine gradient vectors $\nabla_{x,\affnorm} L$ and $\nabla_{x',\affnorm} L'$
can, up to an arbitrary rotation of the elements, and to first order of approximation, be
perfectly matched between different views of the same local surface patch;
see Figure~\ref{fig-geom-int-cov-prop-sc-norm-aff-grad} for a geometric
illustration and Figure~\ref{fig-comm-diag-spat-ders-sc-norm} for a commutative
diagram.

\begin{figure*}[hbt]
  \[
    \begin{CD}
       \hspace{0mm}\nabla_{x,\affnormtiny} L(x;\; s, \Sigma)
       @>{\footnotesize\begin{array}{c} x' = S_x A \, x  \\ s' \,
                         \Sigma' = s \, S_x^2 \, A \, \Sigma \, A^{T} \\
                         \nabla_{x,\affnormtiny} = s^{1/2} \, \Sigma^{1/2} \, \nabla_x \\
                         \nabla_{x',\affnormtiny} = {s'}^{1/2} \, {\Sigma'}^{1/2} \, \nabla_{x'} \\                         
                         \nabla_{x',\affnormtiny} =
    \tilde{\rho} \, \nabla_{x,\affnormtiny} \end{array}}>>
                     \nabla_{x',\affnormtiny} L'(x';\; s', \Sigma') \\
       \Big\uparrow\vcenter{\rlap{$\scriptstyle{{*
               (\nabla_{x,\affnormtiny} g)(x;\; s,\Sigma)}}$}} & &
       \Big\uparrow\vcenter{\rlap{$\scriptstyle{{*(
               \nabla_{x',\affnormtiny} g)(x';\; s', \Sigma')}}$}} \\
       f(x) @>{\footnotesize \begin{array}{c} x' = S_x A \, x
                                  \\ \end{array}}>> f'(x')
    \end{CD}
  \]
\caption{Commutative diagram for scale-normalized affine gradient
  operators under spatial affine transformations of the form $x' = S_x \, A$.
  This commutative diagram, which should be read from the lower left corner to the
  upper right corner, means that irrespective of whether the input image 
  $f(x)$ is first subject to the affine transformation
  $x' = S_x \, A$ 
  and then filtered with a scale-normalized affine gradient kernel
    $(\nabla_{x',\affnorm} \, g)(x';\; s', \Sigma')$,
  or instead directly convolved
  with the scale-normalized affine gradient kernel
  $(\nabla_{x,\affnorm} \, g)(x;\; s, \Sigma)$ and then
  subject to the same affine transformation, we do
  then, up to a possibly unknown rotation transformation $\tilde{\rho}$, get the same
  result, provided that the parameters of the spatial
  smoothing kernels $g(x;\; s, \Sigma)$ and $g(x';\; s', \Sigma')$ are related
  according to $s' \, \Sigma' = s \, S_x^2 \, A \, \Sigma \, A^{T}$.}
\label{fig-comm-diag-spat-ders-sc-norm}
\end{figure*}

By counting the number of dimensions involved in the the variabilities
spanned by this covariance result, we have that the spatial scale
parameter $s$ and the spatial covariance matrix $\Sigma$ together span
an effective dimensionality of~3, because their joint occurrence in
the product $s \, \Sigma$. The space of the full%
\footnote{Because of physical constraints, as arising when viewing a
  3-D surface patch in terms of 2-D images, we do, however, restrict the spatial
  scaling factors $\sigma_1$ and $\sigma_2$
  in a decomposition of the 2-D affine transformation matrix according to
  $A = {\cal R}_{\psi/2} \, {\cal R}_{\varphi/2}
  \, \diag(\sigma_1, \sigma_2) \, {\cal R}_{\varphi/2} \, {\cal R}_{-\psi/2}$,
  where ${\cal R}_{\psi/2}$ and ${\cal R}_{\varphi/2}$ are rotation
  matrices (see Equation~(15) in Lindeberg (\citeyear{Lin95-ICCV})),
  to be positive.}
2-D affine
transformations (with the spatial translation offset not considered here)
spans a variability over 4~dimensions. Thus, the stated covariance
results holds over independent variabilities over all the totally
7~dimensions of the resulting manifold.

\subsection{Scale-normalized affine Hessian operator for an affine
  scale-space representation based on smoothing with 
  anisotropic affine Gaussian kernels}
\label{sec-sc-norm-aff-hess-op}

Given the affine Gaussian scale-space representation
$L \colon \bbbr^2 \times \bbbr_+ \times  \, \bbbs_+^2 \rightarrow \bbbr$
according to (\ref{eq-aff-scsp-def-sc-norm-aff-grad})
of any 2-D image $f \colon \bbbr^2 \rightarrow \bbbr$,
we can define the regular Hessian matrix as
\begin{equation}
  {\cal H}_x L = \nabla_x \, \nabla_x^T L
  = \left(
        \begin{array}{cc}
           L_{x_1x_1} & L_{x_1x_2} \\
           L_{x_1x_2} & L_{x_2x_2}
         \end{array}
      \right).
\end{equation}
For an affine Gaussian scale-space representation $L(x;\, s, \Sigma)$
computed for
spatial scale parameter $s \in \bbbr_+$ and spatial covariance matrix
$\Sigma$, with the previously defined affine gradient operator
$\nabla_{x,\affnorm}$ according to (\ref{eq-def-sc-norm-aff-grad-op}),
it is therefore natural to define a corresponding scale-normalized
affine Hessian operator ${\cal H}_{x,\affnorm}$ according to
\begin{equation}
  {\cal H}_{x,\affnorm}
  = \nabla_{x,\affnorm} \, \nabla^T_{x,\affnorm},
\end{equation}
which, when expanded from the definition, then assumes the form
\begin{align}
  \begin{split}
    {\cal H}_{x,\affnorm}
    & = s \, (\Sigma^{1/2}) \, \nabla_x \, \nabla_x^T (\Sigma^{1/2})^T
  \end{split}\\
  \begin{split}
    \label{eq-def-sc-norm-aff-hess-mat}
     & = s \, (\Sigma^{1/2}) \, {\cal H}_x \, (\Sigma^{1/2})^T,
  \end{split}    
\end{align}
with the interpretation that, since the matrix $(\Sigma^{1/2})^T$ is to
be regarded as a constant with regard to the spatial differentiation
operators $\nabla_x$ and ${\cal H}_x$, these operators can be applied
through this matrix.

\subsection{Full covariance property of the scale-normalized affine
  Hessian operator under general spatial affine transformations}
\label{sec-cov-prop-sc-norm-aff-hess-mat}

Consider again a spatial affine transformation of the form
\begin{equation}
  \label{eq-spat-aff-transf-def-aff-sc-norm-ders-hess}
  f'(x') = f(x) \quad\quad\mbox{for}\quad\quad x' = S_x \, A \, x,
\end{equation}
where $S_x \in \bbbr_+$ represents an overall spatial scaling factor,
$A$ is a $2 \times 2$ affine transformation matrix,
with the affine Gaussian scale-space representations $L(x;\; s, \Sigma)$
and $L'(x';\; s', \Sigma')$
according to (\ref{eq-aff-scsp-def-sc-norm-aff-grad}) and
(\ref{eq-aff-scsp-def-sc-norm-aff-grad-prim}),
for matching values of the spatial scale parameters
$s \in \bbbr_+$ and $s' \in \bbbr_+$ as
well as the $2 \times 2$ spatial covariance matrices $\Sigma$ and $\Sigma'$ over
the two domains according to
(\ref{eq-transf-prop-sc-par-spat-cov-mat-pure-aff-scsp-main-result})
\begin{equation}
  \label{eq-transf-prop-sc-par-spat-cov-mat-pure-aff-scsp-hess}
  s' \Sigma' = s \, S_x^2 \, A \, \Sigma \, A^T,
\end{equation}
such that the affine scale-space representations
$L(x;\; s, \Sigma)$ and $L'(x';\; s', \Sigma')$,
for these parameter values of the affine Gaussian smoothing kernels, are
equal (\ref{eq-equal-aff-scsp-repr-def-aff-sc-norm-ders})
\begin{equation}
  \label{eq-equal-aff-scsp-repr-aff-cov-proof-hess}
  L'(x';\; s', \Sigma') = L(x;\; s, \Sigma).
\end{equation}
Let us then define the scale-normalized affine Hessian operator over the
two domain as
\begin{align}
  \begin{split}
    \label{eq-def-sc-norm-aff-hess-op-aff-cov-main-result}
     {\cal H}_{x,\affnorm}
     & = \nabla_{x,\affnorm} \, \nabla^T_{x,\affnorm},
   \end{split}\\
   \begin{split}
     {\cal H}_{x',\affnorm}
    & = \nabla_{x',\affnorm} \, \nabla^T_{x',\affnorm},
    \end{split}
\end{align}
with the underlying scale-normalized affine gradient operators of the
forms (\ref{eq-def-sc-norm-aff-grad-op-aff-cov-main-result}) and
(\ref{eq-def-sc-norm-aff-grad-op-aff-cov-prim-main-result}),
\begin{align}
  \begin{split}
      \nabla_{x,\affnorm}
      & = s^{1/2} \, \Sigma^{1/2} \, \nabla_x,
   \end{split}\\
   \begin{split}
     \nabla_{x',\affnorm}
     & = {s'}^{1/2} \, {\Sigma'}^{1/2} \, \nabla_{x'},
   \end{split}
\end{align}
which when combining these expressions over the transformed domain gives
\begin{align}
  \begin{split}
    {\cal H}_{x',\affnorm}
    & = s' \, ({\Sigma'}^{1/2}) \, \nabla_{x'} \, \nabla_{x'}^T \, ({\Sigma'}^{1/2})^T
  \end{split}\\
  \begin{split}
     & = s' \, ({\Sigma'}^{1/2}) \, {\cal H}_{x'} \, ({\Sigma'}^{1/2})^T.
  \end{split}    
\end{align}
By additionally taking the indeterminacy with respect to a possible
rotation matrix $\tilde{\rho}$ into account, we obtain
\begin{equation}
  \label{eq-equal-aff-scsp-repr-aff-cov-proof-hess-again}
  {\cal H}_{x',\affnorm} = \tilde{\rho} \,  {\cal H}_{x,\affnorm} \, \tilde{\rho}^T.
\end{equation}
This result thereby implies that, when these scale-normalized affine Hessian
operators are applied to the affine Gaussian
scale-space representations $L(x;\; s, \Sigma)$ and $L'(x';\; s', \Sigma')$
over their respective domains, we obtain that
\begin{equation}
  ({\cal H}_{x',\affnorm} L')(x';\; s', \Sigma')
  = \tilde{\rho} \, ({\cal H}_{x,\affnorm} L)(x;\; s, \Sigma) \, \tilde{\rho}^T
\end{equation}
will hold for some rotation matrix $\tilde{\rho}$, provided that the
parameters of the underlying affine Gaussian
smoothing kernels are related according to
\begin{equation}
  \label{eq-match-sc-pars-cov-mats-cov-prop-sc-norm-aff-hess}
  s' \, \Sigma' = s \, (S_x \, A) \, \Sigma \, (S_x A)^T = s \, S_x^2 \, A \, \Sigma \, A^T.
\end{equation}
Thus, this definition of the scale-normalized affine Hessian matrix is
also covariant under the full group of non-singular spatial affine
transformations.
Again, the rotation matrix $\tilde{\rho}$ is restricted to a
unit matrix in the case of similarity transformations.

\begin{figure}[hbt]
 \begin{center}
    \includegraphics[width=0.40\textwidth]{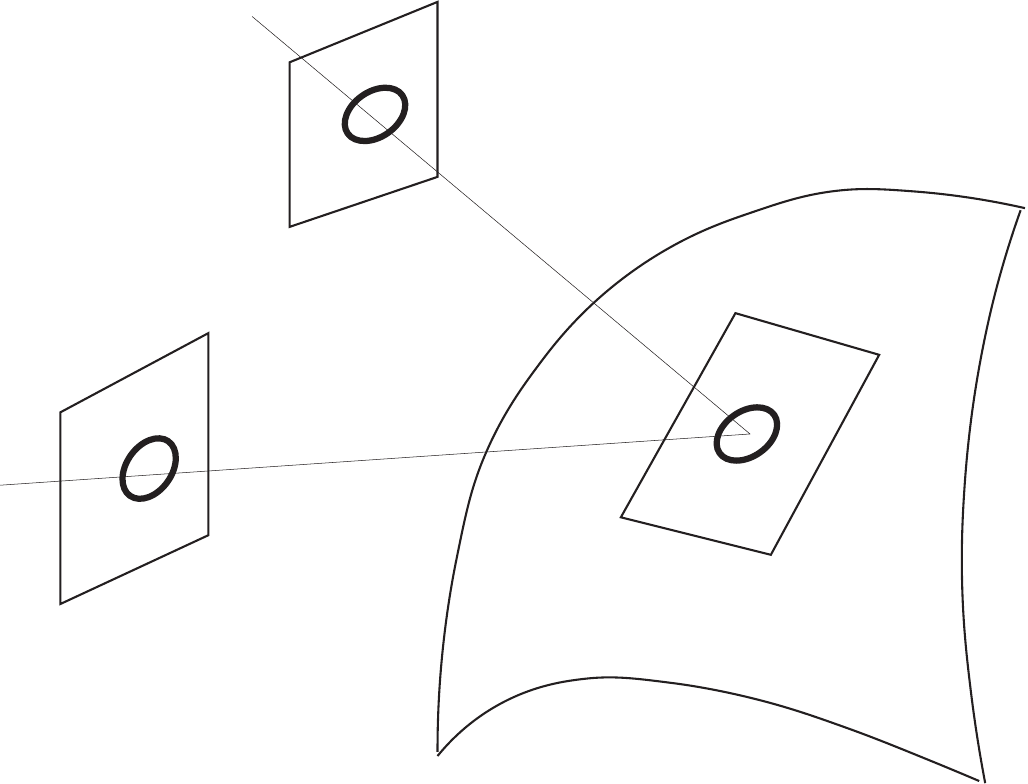}
  \end{center}  
  \caption{The covariance property
    (\ref{eq-equal-aff-scsp-repr-aff-cov-proof-hess-again}) of the
    scale-normalized affine Hessian operator
    (\ref{eq-def-sc-norm-aff-hess-op-aff-cov-main-result}) under
    general (non-singular) affine transformations means that, if we
    consider two cameras, that view the same local surface patch from
    general (non-degenerate) viewing conditions, then, to first order
    of approximation, the resulting affine Hessian responses for the
    different views, here illustrated as ellipses before the affine
    scale normalization, can, up
    to a combination of two (in this 2-D case equal)
    rotation transformations $\tilde{\rho}$
    and $\tilde{\rho}^T$, be perfectly
    matched, provided that the scale parameters and the covariance
    matrices of the receptive fields are properly matched according to
    (\ref{eq-match-sc-pars-cov-mats-cov-prop-sc-norm-aff-hess}).}
  \label{fig-geom-int-cov-prop-sc-norm-aff-hess}
\end{figure}

Interpreted geometrically,
this result means that if we interpret the spatial
affine transformation as a local linearization of either the perspective
mapping from the tangent plane of a local surface patch to the image
domain, or as a local linearization of the projective transformation
between two different views of the same local surface patch, then it holds
that the scale-normalized affine Hessian matrices
computed for matching image points and matching receptive field
parameters of the two domains, can to, first order of approximation,
be perfectly matched, provided that the scale parameters and the
covariance matrices in the two domains are matched according to
(\ref{eq-transf-prop-sc-par-spat-cov-mat-pure-aff-scsp-hess});
see Figure~\ref{fig-geom-int-cov-prop-sc-norm-aff-hess} for an illustration.

\subsection{Scale-normalized regular temporal derivative operators}
\label{sec-sc-norm-temp-ders}

Given any 1-D temporal signal $f \colon \bbbr \rightarrow \bbbr$,
consider its temporal scale-space representation
$L \colon \bbbr \times \bbbr_+ \rightarrow \bbbr$,
at temporal scale level $\tau \in \bbbr_+$:
\begin{equation}
  L(\cdot;\; \tau) = h(\cdot;\; \tau) * f(\cdot),
\end{equation}
obtained by convolution with either a non-causal
temporal Gaussian kernel (\ref{eq-def-temp-gauss-kern}),
the time-causal limit kernel (\ref{eq-time-caus-limit-kern}),
or, more generally, some other scale-covariant temporal kernel
$h \colon \bbbr \times \bbbr_+ \rightarrow \bbbr$,
that for any temporal scaling factor $S_t \in \bbbr_+$
obeys the temporal scaling property
(\ref{eq-temp-sc-cov-temp-kernel})
\begin{equation}
     h(t';\; \tau') = \frac{1}{S_t} \, h(t;\; \tau)
\end{equation}
under any temporal scaling transformation of the form
\begin{equation}
  t' = S_t \, t, \quad\quad\mbox{and}\quad\quad \tau' = S_t^2 \, \tau.
\end{equation}
Then, corresponding scale-normalized analogues
of the regular temporal derivative operators (\ref{eq-temp-der-def})
can be defined according to
Lindeberg (\citeyear{Lin17-JMIV}) 
\begin{equation}
  \label{eq-temp-der-def-sc-norm}
  \partial_{t,\norm}^n = \tau^{n/2} \, \partial_t^n.
\end{equation}
In analogy with the above scale-normalized spatial derivative
operators, the multiplication of the regular temporal derivative
operators by the temporal scale parameter raised to a power
proportional to the order of temporal differentiation, will compensate
for the otherwise general decrease in the magnitude of temporally
smoothed temporal derivatives with increasing temporal scales,
to enable truly scale-covariant temporal
derivative operators, whose magnitudes can be perfectly matched under
temporal scaling transformations, as will be described in the next section.

\subsection{Covariance property for scale-normalized regular
  temporal derivatives under temporal scaling transformations}
\label{sec-cov-prop-temp-sc-ders}

Given any 1-D temporal signal $f \colon \bbbr \rightarrow \bbbr$,
define a rescaled temporal signal
$f' \colon \bbbr \rightarrow \bbbr$
by a temporal scaling transformation of the form
\begin{equation}
  f'(t') = f(t)  \quad\quad\mbox{for}\quad\quad  t' = S_t \, t,
\end{equation}
for a temporal scaling factor $S_t \in \bbbr_+$,
and define purely temporal scale-space representations
$L \colon \bbbr \times \bbbr_+ \rightarrow \bbbr$
and
$L' \colon \bbbr \times \bbbr_+ \rightarrow \bbbr$
of $f$ and $f'$, respectively, according to
\begin{align}
  \begin{split}
    L(\cdot;\; \tau) = h(\cdot;\; \tau) * f(\cdot),
  \end{split}\\
  \begin{split}
    L'(\cdot;\; \tau') = h(\cdot;\; \tau') * f'(\cdot),
  \end{split}
\end{align}
that obey temporal scale covariance for the underlying
temporal smoothing transformation, such that
\begin{equation}
   L'(t';\; \tau') = L(t;\; \tau)
\end{equation}
holds for matching values of
the temporal scale parameters according to
\begin{equation}
  \tau' = S_t^2 \, \tau.
\end{equation}
This property does, for example, both hold for the non-causal temporal
scale-space representation, defined from convolutions with 1-D temporal
Gaussian kernels of the form (\ref{eq-def-temp-gauss-kern}),
and for the time-causal temporal scale-space
representation, defined from convolutions with the time-causal limit
kernel of the form (\ref{eq-time-caus-limit-kern})
(see Equations~(10) and~(104) in Lindeberg (\citeyear{Lin17-JMIV}) for
the temporal differentiation order $n = 0$ in that paper).

Let us next define corresponding scale-normalized temporal derivatives over the
transformed temporal domain according to
\begin{align}
  \begin{split}
     \partial_{t',\norm}^n & = {\tau'}^{n/2} \, \partial_{t'}^n.
  \end{split}
\end{align}
Then, since the scale-normalized temporal derivatives over the two
different domains will be related according to
(see Equations~(10) and~(104) in Lindeberg (\citeyear{Lin17-JMIV}) for
the scale normalization power $\gamma$ set to $\gamma = 1$ in that paper):
\begin{equation}
  \partial_{t',\norm} = \partial_{t,\norm},
\end{equation}
it follows that the scale-normalized temporal derivatives will be equal in the
two domains, such that
\begin{align}
  \begin{split}
    \label{eq-temp-sc-cov-property-pure-temp-ders}
     L'_{t',\norm}(t';\; \tau') & = L_{t,\norm}(t;\; \tau),
  \end{split}
\end{align}
which constitute covariance properties for
scale-normalized temporal derivatives of a purely
temporal scale-space representation.

\subsection{Scale-normalized velocity-adapted temporal derivative operators}
\label{sec-sc-norm-vel-adapt-temp-ders}

Consider any 2+1-D spatio-temporal video sequence or video stream
$f \colon \bbbr^2 \times \bbbr \rightarrow \bbbr$
of the form $f(x, t)$, where $x = (x_1, x_2) \in \bbbr^2$
denotes the spatial coordinates and $t \in \bbbr$ denotes the time variable.
To describe the properties of scale-normalized analogues of the velocity-adapted
derivative operators, which also involve the computation of spatial
derivatives, let us next consider a space-time separable
spatio-temp\-oral representation
$L \colon \bbbr^2 \times \bbbr \times \bbbr_+ \times \bbbr_+ \rightarrow \bbbr$
obtained by joint convolution with an isotropic Gaussian kernel
$g \colon \bbbr^2 \times \bbbr_+ \rightarrow \bbbr$
and a temporal smoothing kernel
$h \colon \bbbr \times \bbbr_+ \rightarrow \bbbr$
according to
\begin{equation}
  L(\cdot, \cdot;\; s, \tau)
  = g(\cdot;\; s) *_x h(\cdot;\; \tau) *_t f(\cdot, \cdot), 
\end{equation}
where the convolution with the spatial Gaussian kernel $g(\cdot;\; s)$ is
performed over the spatial domain only, and the convolution with the
temporal kernel $h(\cdot;\; \tau)$ is performed over the temporal
domain only, here indicated by the corresponding spatial and temporal
convolution operators $*_x$ and $*_t$, respectively.
Then, we can define scale-normalized analogues of the velocity-adapted
temporal derivative operators (\ref{eq-vel-adapt-der-def}) according
to  (as an extension of Lindeberg (\citeyear{Lin13-BICY}) Equation~(82))
\begin{equation}
  \label{eq-vel-adapt-der-def-sc-norm}
  \partial_{{\bar t},\norm}^n
  = \tau^{n/2} \,  (v^T \, \nabla_x + \partial_t)^n.
\end{equation}

\subsection{Covariance property for scale-normalized
  velocity-adapted temporal derivatives under temporal scaling transformations}
\label{sec-cov-prop-sc-norm-vel-adapt-temp-ders}

Given any 2+1-D spatio-temporal video sequence or video stream
$f \colon \bbbr^2 \times \bbbr \rightarrow \bbbr$,
a spatial scaling factor $S_x \in \bbbr_+$ and a temporal scaling
factor $S_t \in \bbbr_+$, let us next define a scaled video sequence
or video stream
$f' \colon \bbbr^2 \times \bbbr \rightarrow \bbbr$
under the composition of a spatial scaling transformation and
a temporal scaling transformation over the joint
space-time domain of the form
\begin{equation}
  f'(x', t') = f(x, t)  \quad\mbox{for}\quad  x' = S_x \, x 
  \quad\mbox{and}\quad t' = S_t \, t.
\end{equation}
Furthermore, let us define the space-time-separable spatio-temporal scale-space
representations
$L \colon \bbbr^2 \times \bbbr \times \bbbr_+ \times \bbbr_+ \rightarrow \bbbr$
and
$L' \colon \bbbr^2 \times \bbbr \times \bbbr_+ \times \bbbr_+ \rightarrow \bbbr$
of $f$ and $f'$, respectively, according to
\begin{align}
  \begin{split}
    L(\cdot, \cdot;\; s, \tau)
    = g(\cdot;\; s) *_x h(\cdot;\; \tau) *_t f(\cdot, \cdot),
  \end{split}\\
  \begin{split}
    L'(\cdot, \cdot;\; s', \tau')
    = g(\cdot;\; s') *_x h(\cdot;\; \tau') *_t f'(\cdot, \cdot),
  \end{split}
\end{align}
for matching values of the spatial and the
temporal scale parameters according to
\begin{equation}
  s' = S_x^2 \, s \quad\mbox{and}\quad \tau' = S_t^2 \, \tau.
\end{equation}
Let us also, in addition to the definition of scale-normalized
velocity-adapted temporal derivatives for the original video sequence
or video stream $f$
according to (\ref{eq-vel-adapt-der-def-sc-norm}),
define corresponding scale-normalized velocity-adapted
temporal derivatives over the jointly rescaled video sequence or video
stream $f'$ according to
\begin{align}
  \begin{split}
    \label{eq-def-sc-norm-vel-adapt-ders}
    \partial_{{\bar t}',\norm}^n
    = {\tau'}^{n/2} \, ({v'}^T \, \nabla_{x'} + \partial_{t'})^n.
  \end{split}
\end{align}
Then, provided that we define the transformed velocity vector $v'$
according to
\begin{equation}
  v' = \frac{S_x}{S_t} \, v,
\end{equation}
it follows that
the scale-normalized temporal derivatives of the
spatio-temporal scale-space representations in the two domains will be
equal
\begin{align}
  \begin{split}
    \label{eq-temp-sc-cov-property-veladapt-temp-ders}    
    L'_{{\bar t}',\norm}(x', t';\; s', \tau') &
    = L_{\bar t,\norm}(x, t;\; s, \tau).
  \end{split}
\end{align}
This constitutes the covariance property for
scale-normalized velocity-adapted temporal derivatives of a
space-time-separ\-able spatio-temporal scale-space representation,
under compositions of spatial scaling transformations and
temporal scaling transformations.

Both this temporal scale covariance property and the previously
treated temporal scale covariance property
(\ref{eq-temp-sc-cov-property-pure-temp-ders}) have the geometric
interpretation that we can, to first
order of approximation, perfectly match the temporal receptive
field responses between different views of a similar spatio-temporal
event, that occurs either faster or slower in relation to a previous
view of an otherwise similar event, with the other viewing parameters,
except the speed of the event, being different.

The additional degree of freedom introduced here, by also including an
arbitrary uniform spatial scaling transformation of the spatial domain, has been
introduced here, to demonstrate that the underlying space-time
separable spatio-temporal scale-space representation $L(x, t;\; s, \tau)$
is closed, also under arbitrary combinations of such variabilities,
however, then with the
important constraint that the image velocity must be adapted, as
determined by the spatial and temporal the temporal scaling factors
$S_x$ and $S_t$.
If we, on the other hand, would like to achieve closedness under
free variabilities of the velocity vector $v$, independent of the
spatial scaling factor $S_x$ and the temporal scaling factor $S_t$, then
a more complex joint spatio-temporal scale-space concept with
additional parameters for the receptive fields is needed, as
will be addressed in the next section.

\begin{table*}[hbt]
  \begin{tabular}{lll}
    \hline
    Section & Topic & Contribution \\
    \hline
    \ref{sec-transf-props-spat-temp-scsp-individ}
            & Covariance properties of the pure smoothing operation
              under individual image transformations
                    & Review of Lindeberg (\citeyear{Lin23-FrontCompNeuroSci}) \\
    \ref{sec-transf-props-spat-temp-ders-individ}
            & Covariance properties of spatio-temporal derivatives under individual 
              image transformations
                    & Extension of Lindeberg (\citeyear{Lin23-FrontCompNeuroSci}) \\
    \ref{sec-composed-img-transf}
            & Definition of the studied form of composed spatio-temporal image transformations
                    & New \\
    \ref{eq-transf-prop-spat-temp-smooth}
            & Covariance property of smoothing operation under composed
              spatio-temporal transformations
                    & New \\
    \ref{eq-transf-prop-spat-temp-ders}
            & Covariance properties of derivative operators under composed
              spatio-temporal transformations
                    & New \\
    \ref{sec-transf-props-geom-spat-temp-ders}
            & Covariance properties of geometric derivatives under composed
              spatio-temporal transformations
                    & New \\    
    \ref{sec-transf-prop-sc-norm-spat-temp-ders}
            & Covariance properties of scale-normalized spatio-temporal
              derivatives under compositions
                    & New \\
              \hline
  \end{tabular}
  \caption{Overview of the conceptual contributions regarding
    covariance properties over a {\em joint spatio-temporal domain\/} in the different
    subsections in Sections~\ref{sec-individ-cov-props}  and~\ref{sec-joint-cov-props},
    which then constitute the conceptual foundation for the 
    geometric interpretations and implications, that will follow in
    Sections~\ref{sec-geom-interpret}--\ref{sec-cues-3d-structure}.}
  \label{tab-contribs-sec45}
\end{table*}

\section{Covariance properties of the generalized Gaussian derivative
  model for spatio-temporal receptive fields}
\label{sec-individ-cov-props}

For processing 2+1-D spatio-temporal image data
$f \colon \bbbr^2 \times \bbbr \rightarrow \bbbr$, convolution of
the input video sequence or video stream $f(x, t)$, where $x = (x_1, x_2) \in \bbbr^2$ denotes the image
coordinates and $t \in \bbbr$ the time variable,
with the purely spatio-temporal smoothing kernel
$T \colon \bbbr^2 \times \bbbr \times \bbbr_+ \times \bbbs_+^2 \times \bbbr_+
\times \bbbr^2 \rightarrow \bbbr$
according to (\ref{eq-spat-temp-RF-model})
\begin{equation}
  \label{eq-spat-temp-RF-model-again-cov-props-basic}
  T(x, t;\; s, \Sigma, \tau, v) 
  = g(x - v \, t;\; s, \Sigma) \, h(t;\; \tau),
\end{equation}
defines spatio-temporal smoothed image data
$L \colon \bbbr^2 \times \bbbr \times \bbbr_+ \times \bbbs_+^2 \times \bbbr_+
\times \bbbr^2 \rightarrow \bbbr$
according to
\begin{equation}
  \label{eq-def-spat-temp-scsp}
  L(\cdot, \cdot;\; s, \Sigma, \tau, v) = T(\cdot, \cdot;\; s, \Sigma, \tau, v) * f(\cdot, \cdot),
\end{equation}
which is referred to as a spatio-temporal scale-space representation
of $f$ over the spatial scale parameter $s \in \bbbr_+$,
the spatial covariance matrix $\Sigma \in \bbbs_+^2$,
the temporal scale parameter $\tau \in \bbbr_+$ and
the velocity vector $v \in \bbbr^2$
(Lindeberg \citeyear{Lin10-JMIV}).

In this section, we will first in
Section~\ref{sec-transf-props-spat-temp-scsp-individ} restate
covariance properties of spatio-temporal receptive fields under
four {\em individual\/} classes of geometric image transformations,
as previously formulated in Lindeberg (\citeyear{Lin23-FrontCompNeuroSci}).
Then, we will combine these results with the transformation properties of
the regular (not scale-normalized) either purely spatial, the purely temporal or the joint
spatio-temporal derivative operators expressed for the restricted
subdomains in Section~\ref{sec-sc-norm-spat-temp-ders},
to in Section~\ref{sec-transf-props-spat-temp-ders-individ} express
transformation properties of spatio-temporal derivative operators over
a {\em joint\/} spatio-temporal domain, for
each class of individual geometric image transformations.
This conceptual background will then constitute a conceptual
foundation for formulating both covariance properties and
transformation properties of spatio-temporal receptive fields
in Section~\ref{sec-joint-cov-props}, for a
specific way of composing the four different classes of individual geometric
image transformations in cascade.

Table~\ref{tab-contribs-sec45} gives a comprehensive overview of the
different theoretical contributions that will follow in
Sections~\ref{sec-individ-cov-props}  and~\ref{sec-joint-cov-props}.

\subsection{Transformation properties of the spatio-temporal
  scale-space representation in the spatio-temporal receptive field
  model under geometric image transformations}
\label{sec-transf-props-spat-temp-scsp-individ}

In Lindeberg (\citeyear{Lin23-FrontCompNeuroSci}), covariance
properties of this generalized Gaussian derivative model for receptive
fields were studied in detail. It was specifically shown that:
\begin{itemize}
\item
  Under {\em purely spatial scaling transformations\/} of the form
  \begin{equation}
      f'(x', t') = f(x, t)
  \end{equation}
  for $t' = t$ and
   \begin{equation}
     x' = S_x \, x,
   \end{equation}
  where $S_x \in \bbbr_+$ denotes the spatial scaling factor,
  the spatio-temporal scale-space representations $L'$ and $L$, obtained by
  convolving the input signals $f'$ and $f$ with spatio-temporal
  convolution kernels of the form (\ref{eq-spat-temp-RF-model}),
  are related according to
  \begin{equation}
    L'(x', t';\; s', \Sigma', \tau', v')   = L(x, t;\; s, \Sigma, \tau, v),
  \end{equation}
  provided that the spatial scale parameters and the velocity vectors
  are related according to
  \begin{equation}
     \label{eq-transf-pars-spat-scal}
     s' = S_x^2 \, s \quad\quad\mbox{and}\quad\quad v' = S_x \, v.
   \end{equation}
\item
  Under {\em spatial affine transformations\/} of the form
  \begin{equation}
      f_R(x_R, t_R) = f_L(x_L, t_L)
  \end{equation}
  for $t_R = t_L$ and 
  \begin{equation}
     x_R = A \, x_L,
  \end{equation}
  where $x_L \in \bbbr^2$ and $x_R \in \bbbr^2$ denote the spatial
  image coordinates in the two domains and
  $A$ denotes a non-singular $2 \times 2$ affine transformation matrix, 
  the spatio-temporal scale-space representations $L_R$ and $L_L$, obtained by
  convolving the input signals $f_R \colon \bbbr^2 \times \bbbr \rightarrow \bbbr$
  and   $f_L \colon \bbbr^2 \times \bbbr \rightarrow \bbbr$ with spatio-temporal
  convolution kernels of the form (\ref{eq-spat-temp-RF-model}),
  are related according to
  \begin{multline}
    L_R(x_R, t_R;\; s_R, \Sigma_R, \tau_R, v_R) = \\
    = L_L(x_L, t_L;\; s_L, \Sigma_L, \tau_L, v_L) ,
  \end{multline}
  provided that the spatial covariance matrices
  $\Sigma_L \in \bbbs_+^2$ and $\Sigma_R \in \bbbr_+^2$ as well as the velocity
  vectors $v_L \in \bbbr^2$ and $v_R \in \bbbr^2$ are related according to
  \begin{equation}
    \label{eq-transf-pars-spat-aff}
     \Sigma_R = A \, \Sigma_L \, A^T \quad\mbox{and}\quad v_R = A \, v_L,
  \end{equation}
  and provided that the other parameters of the receptive fields are the same.
\item
  Under {\em purely temporal scaling transformations\/} of the form
  \begin{equation}
     f'(x', t') = f(x, t)
  \end{equation}
  for $x' = x$ and
  \begin{equation}
     t' = S_t^2 \, t,
   \end{equation}
  where $S_t \in \bbbr_+$ is the temporal scaling factor,
  the spatio-temporal scale-space representations $L'$ and $L$, obtained by
  convolving the input signals $f'$ and $f$ with spatio-temporal
  convolution kernels of the form (\ref{eq-spat-temp-RF-model}),
  are related according to
  \begin{equation}
    L'(x', t';\; s', \Sigma', \tau', v') = L(x, t;\; s, \Sigma, \tau, v),
  \end{equation}
  provided that the temporal scale parameters $\tau \in \bbbr_+$ and
  $\tau' \in \bbbr_+$ as well as the velocity
  vectors $v \in \bbbr^2$ and $v' \in \bbbr^2$ are related according to
  \begin{equation}
    \label{eq-transf-pars-temp-sc}
    \tau' = S_t^2 \, \tau  \quad\quad\mbox{and}\quad\quad v' = v/S_t,
  \end{equation}
 and provided that the other parameters of the receptive field are the same.
\item
  Under {\em Galilean transformations\/} of the form
  \begin{equation}
     f'(x', t') = f(x, t)
  \end{equation}
  for $t' = t'$ and
  \begin{equation}
     \label{eq-gal-trans-sec-cov-prop}
     x' = x + u \, t,
   \end{equation}
  where $u \in \bbbr^2$ is a velocity vector,
  the spatio-temporal scale-space representations $L'$ and $L$, obtained by
  convolving the input signals $f'$ and $f$ with spatio-temporal
  convolution kernels of the form (\ref{eq-spat-temp-RF-model}),
  are related according to
  \begin{equation}
     L'(x', t';\; s', \Sigma', \tau', v') = L(x, t;\; s, \Sigma, \tau, v),
  \end{equation}
  provided that the velocity parameters $v \in \bbbr^2$ and
  $v' \in \bbbr^2$
  for the two domains are related according to
  \begin{equation}
    \label{eq-transf-pars-galilean}    
    v' = v + u
  \end{equation}
  and provided that the other parameters of the receptive fields are the same.
\end{itemize}
A notable characteristics of the transformation properties of the
spatio-temporal
scale-space representations under these classes of natural image transformations,
is that the four classes of geometric image transformations are not
independent. Instead, for example, both the spatial scaling
transformations and the temporal scaling
transformations affect the image velocities, beyond the spatial and
temporal scale parameters, respectively. Furthermore, the spatial affine
transformations also affect the image velocities. For this reason, it
is of interest to additionally model all the four classes of image transformations
{\em jointly\/}, which we will do in Section~\ref{sec-joint-cov-props}.

Before that, let us, however, first also address the transformation
properties of the spatial and the temporal derivative operators, which
substantially extends the previous treatment of transformation properties of
spatio-temporal derivative operators over joint space-time in Lindeberg
(\citeyear{Lin23-FrontCompNeuroSci}), as here based on the in-depth
treatment of transformation properties of spatial and temporal
derivatives over spatial or temporal subdomains in
Section~\ref{sec-sc-norm-spat-temp-ders} in the current paper.

\subsection{Transformation properties of the spatio-temporal
  derivative operators  in the spatio-temporal receptive field
  model under geometric image transformations}
\label{sec-transf-props-spat-temp-ders-individ}

Beyond the spatio-temporal smoothing kernel
$T(x, t;\; s, \Sigma, \tau, v)$,
that defines the spatio-temporal scale-space representation
$L(x, t;\; s, \Sigma, \tau, v)$ in
(\ref{eq-def-spat-temp-scsp}),
the spatio-temporal receptive field model
(\ref{eq-spat-temp-RF-model-der})
\begin{multline}
  \label{eq-spat-temp-RF-model-der-again}
    T_{{\varphi}^{m} {\bar t}^n}(x, t;\; s, \Sigma, \tau, v) 
    =  \\ =
        \partial_{\varphi}^{m} \, \partial_{\bar t}^n 
           \left( g(x - v \, t;\; s, \Sigma) \, h(t;\; \tau) \right),
\end{multline}
that has been used for modelling the
the receptive fields of simple cells in the primary visual cortex,
does additionally comprise
spatial derivative operators in terms of directional derivative
operators $\partial_{\varphi}^{m}$ of the form (\ref{eq-dir-der-def}) as well as temporal
derivative operators $\partial_{\bar t}^n$ of the forms (\ref{eq-temp-der-def}) and
(\ref{eq-vel-adapt-der-def}).

In summary, under the four classes of spatio-temporal image
transformations studied in this work, the spatial and the temporal derivative operators
transform as follows:

\begin{itemize}
\item
  Under {\em purely spatial scaling transformations\/} of the form
  \begin{equation}
      f'(x', t') = f(x, t)
  \end{equation}
  for $t' = t$ and
   \begin{equation}
     x' = S_x \, x,
   \end{equation}
  where $S_x \in \bbbr_+$ denotes the spatial scaling factor, and the
  spatial gradient operator $\nabla_x = (\partial_{x_1}, \partial_{x_2})^T$
  transforms according to
  \begin{equation}
    \label{eq-transf-prop-nabla-pure-sc-transf-spat-temp}
     \nabla_x = S_x \, \nabla_{x'},
  \end{equation}
  with the transformed spatial gradient operator defined as
  $\nabla_{x'} = (\partial_{x'_1}, \partial_{x'_2})^T$.
  This means that also the directional derivative operator, defined according to
  (\ref{eq-dir-der-def}), transforms according to
  \begin{equation}
    \label{eq-transf-prop-dphi-pure-sc-transf-spat-temp}    
     \partial_{\varphi} = S_x \, \partial_{\varphi'}.
  \end{equation}
  Under a purely spatial scaling transformation, the
  regular temporal derivative operator is, however, unchanged
  \begin{equation}
     \partial_t = \partial_{t'}.
   \end{equation}
  Due to the transformation property (\ref{eq-transf-pars-spat-scal})
  of the velocity vector $v$ according to $v' = S_x \, v$, the
  velocity-adapted temporal derivative operator
  (\ref{eq-vel-adapt-der-def}) does also, with the transformed temporal
  derivative operator of the form
  \begin{equation}
  \partial_{{\bar t}'} = v'_1 \, \partial_{x'_1} + v'_2 \, \partial_{x'_2} + \partial_{t'},
   \end{equation}
  under a uniform spatial scaling transformation, transform according to
 \begin{equation}
     \partial_{\bar t} = \partial_{{\bar t}'}.
   \end{equation}
\item
  Under {\em spatial affine transformations\/} of the form
  \begin{equation}
      f_R(x_R, t_R) = f_L(x_L, t_L)
  \end{equation}
  for $t_R = t_L$ and 
  \begin{equation}
     x_R = A \, x_L,
  \end{equation}
  where $A$ denotes a $2 \times 2$ affine transformation matrix, the
  spatial gradient operator $\nabla_x$ transforms according to
  \begin{equation}
    \label{eq-transf-prop-nabla-spat-aff-transf-spat-temp}
     \nabla_x = A^T \, \nabla_{x'}.
   \end{equation}
   This implies that if we define directional derivative operator
   (\ref{eq-dir-der-def}) as
   \begin{equation}
     \partial_{\varphi} = e_{\varphi}^T \, \nabla_x
   \end{equation}
   with the unit vector $e_{\varphi}$ in the direction $\varphi$
   transforming according to
   \begin{equation}
     e_{\varphi'} = \frac{A \, e_{\varphi}}{\| A \, e_{\varphi} \|}
   \end{equation}
   to guarantee that also the transformed unit vector will be of unit
   length, then the
   corresponding transformed directional derivative operator
   $\partial_{\varphi'} = e_{\varphi'}^T \nabla_{x'}$ is
   \begin{equation}
      \partial_{\varphi} = \| A \, e_{\varphi} \| \, \partial_{\varphi'}.
   \end{equation}
   Both the regular and the velocity-adapted temporal
   derivative operators are, however, 
   unchanged under purely spatial affine transformations
   \begin{align}
     \begin{split}
       \label{eq-transf-prop-dt-temp-sc-transf-spat-temp}
       \partial_t = \partial_{t'},
     \end{split}\\
     \begin{split}
       \label{eq-transf-prop-dtbar-temp-sc-transf-spat-temp}
       \partial_{\bar t} = \partial_{{\bar t}'},
     \end{split}
   \end{align}
   when taking into account the
   transformation property $v' = A \, v$
   (\ref{eq-transf-pars-spat-aff}) of the velocity vector.
\item
  Under {\em purely temporal scaling transformations\/} of the form
  \begin{equation}
     f'(x', t') = f(x, t)
  \end{equation}
  for $x' = x$ and
  \begin{equation}
     t' = S_t^2 \, t,
   \end{equation}
  where $S_t \in \bbbr_+$ is a temporal scaling factor, both the
  regular spatial gradient operator and the spatial directional
  derivative operators are unchanged
  \begin{align}
    \begin{split}
      \nabla_{x} = \nabla_{x'},
    \end{split}\\
    \begin{split}
      \partial_{\varphi} = \partial_{\varphi'}.
    \end{split}
  \end{align}
  Both the regular and the velocity-adapted temporal derivative
  operators do, however, transform according to
    \begin{align}
     \begin{split}
       \partial_t = S_t \, \partial_{t'},
     \end{split}\\
     \begin{split}
        \partial_{\bar t} = S_t \, \partial_{{\bar t}'}
     \end{split}
    \end{align}
    when taking the transformation property
  $v' = v/S_t$ (\ref{eq-transf-pars-temp-sc})
  of the velocity vector into account.
\item
  Under {\em Galilean transformations\/} of the form
  \begin{equation}
     f'(x', t') = f(x, t)
  \end{equation}
  for $t' = t'$ and
  \begin{equation}
     \label{eq-gal-trans-sec-cov-prop-again}
     x' = x + u \, t,
   \end{equation}
   where $u \in \bbbr^2$ is a velocity vector, both the
  regular spatial gradient operator and the spatial directional
  derivative operators are unchanged
  \begin{align}
    \begin{split}
      \nabla_{x} = \nabla_{x'},
    \end{split}\\
    \begin{split}
      \partial_{\varphi} = \partial_{\varphi'}.
    \end{split}
  \end{align}
  Taking into account the transformation property $v' = v + u$
  (\ref{eq-transf-pars-galilean}) of the velocity vector, the regular
  temporal derivative transforms according to
  \begin{equation}
     \partial_t = u^T \, \nabla_{x'} + \partial_{t'},
   \end{equation}
   whereas the velocity-adapted derivatives are equal
   \begin{equation}
     \partial_{\bar t} = \partial_{{\bar t}'}.
   \end{equation}
\end{itemize}
Thus, we can from this summarizing overview
see how the spatial and the temporal derivative operators
are transformed in ways that interact strongly with the parameters of
the corresponding spatio-temporal image transformations.%
\footnote{Notably, several of these transformation properties become
  simpler, when expressed in terms of scale-normalized derivatives
  according to Section~\ref{sec-sc-norm-spat-temp-ders},
  where (i)~the scale-normalized spatial derivative operators
  $\partial_{\varphi,\norm}^m$ and $\nabla_{x,\norm}$
  according to (\ref{eq-dir-der-def-sc-norm-basic}) and (\ref{eq-nabla-op}) will
  absorb the spatial scaling factor $S_x$ in
  Equations~(\ref{eq-transf-prop-nabla-pure-sc-transf-spat-temp}) 
  and~(\ref{eq-transf-prop-dphi-pure-sc-transf-spat-temp})
  analogous to
  Equations~(\ref{eq-eq-sc-norm-dirders-spat-scsp-spat-sc-transf})
  and~(\ref{eq-eq-sc-norm-nabla-spat-scsp-spat-sc-transf}),
  (ii)~the scale-normalized affine gradient operator
  $\nabla_{x,\affnorm}$ according to
  (\ref{eq-def-sc-norm-aff-grad-op}) will absorb the
  affine transformation matrix $A$ in
  Equation~(\ref{eq-transf-prop-nabla-spat-aff-transf-spat-temp})
  analogous to
  Equation~(\ref{eq-equal-aff-scsp-repr-aff-cov-proof}),
  and (iii)~the scale-normalized temporal derivative operators
  $\partial_{t,\norm}^n$ and $\partial_{{\bar t},\norm}^n$
  according to (\ref{eq-temp-der-def-sc-norm}) and (\ref{eq-vel-adapt-der-def-sc-norm})
  will absorb temporal scaling factor $S_t$ in
  Equations~(\ref{eq-transf-prop-dt-temp-sc-transf-spat-temp})
  and~(\ref{eq-transf-prop-dtbar-temp-sc-transf-spat-temp})
  analogous to
  Equations~(\ref{eq-temp-sc-cov-property-pure-temp-ders})
  and~(\ref{eq-temp-sc-cov-property-veladapt-temp-ders}).
  To save space, we, however, postpone introducing these
  scale-normalized derivatives into the transformation properties of the
  composed spatio-temporal receptive fields, until also addressing the
  joint covariance properties of the spatio-temporal receptive fields in
  Section~\ref{sec-transf-prop-sc-norm-spat-temp-ders}.}
Of particular interest is therefore to also make explicit how these
transformation properties are composed, when coupling the different
types of primitive image transformations in cascade, which we will do in
the next section.

\section{Joint covariance property under spatial and temporal scaling
  transformations, spatial affine transformations and Galilean
  transformations}
\label{sec-joint-cov-props}

In this section, we will derive a set of joint covariance properties over the
composition of (i)~a spatial scaling transformation, (ii)~a spatial affine
transformation, (iii)~a Galilean transformation and (iv)~a temporal scaling
transformation.

For this purpose, we will first in
Section~\ref{sec-composed-img-transf}
define the composed
geometric transformation, and then in
Section~\ref{eq-transf-prop-spat-temp-smooth}
consider how different components
in the integral formulation of the convolution operation are
transformed under the corresponding change of variables, which,
when combined, leads to the desired transformation
property regarding the spatio-temporal smoothing components of the
spatio-temporal receptive fields.

Then, we will additionally in Sections~\ref{eq-transf-prop-spat-temp-ders}--\ref{sec-transf-prop-sc-norm-spat-temp-ders}
complement
with explicit transformation properties regarding the spatial and the
temporal derivative operators, underlying the formulation of the
joint spatio-temporal receptive fields, obtained by applying composed
spatio-temporal derivatives to the joint spatio-temporal smoothing
kernel, specifically including explicit formulations of joint
spatio-temporal covariance and transformation properties
in terms of scale-normalized derivatives.

The results presented in this section will: (i)~extend the individual
covariance properties of the spatio-temporal smoothing operation
for each one of the four primitive types geometric image transformations reviewed in
Section~\ref{sec-transf-props-spat-temp-scsp-individ} to a joint
spatio-temporal covariance property of the spatio-temporal smoothing operation,
(ii)~extend the transformation properties of the spatio-temporal
derivative operators for each one of the four primitive types
geometric image transformations described in
Section~\ref{sec-transf-props-spat-temp-ders-individ} to
transformation properties of spatio-temporal derivative operators
under a joint composition of the four primitive types of geometric
image transformations, and
(iii)~extend the transformation properties of the regular
spatio-temporal derivative operators under the composed geometric
transformation to algebraically much simpler covariance and transformation properties
in terms of scale-normalized derivatives.

In these ways, the presented results will show how spatio-temporal
receptive field responses in terms of spatio-temporal derivatives of
spatio-temporal smoothing operations can be matched under composed
geometric image transformations, provided that the parameters of the
receptive fields are properly matched between the domains before
{\em vs.\/}\ after the composed spatio-temporal image transformation.

\subsection{Composed geometric image transformation}
\label{sec-composed-img-transf}

Consider the composition of:
\begin{itemize}
\item
  a spatial scaling transformation with the spatial scaling factor $S_x
  \in \bbbr_+$,
\item
  a spatial affine transformation with the non-singular $2 \times 2$ affine transformation matrix $A$,
\item
  a Galilean transformation with the velocity vector $u \in \bbbr^2$, and
\item
  a temporal scaling transformation with the temporal scaling factor
  $S_t \in \bbbr_+$
\end{itemize}
of the form
\begin{align}
  \begin{split}
     \label{eq-x-transf}
     x' = S_x \, (A \,  x + u \, t),
   \end{split}\\
  \begin{split}
     \label{eq-t-transf}
     t' = S_t \, t,
   \end{split}
\end{align}
where $x \in \bbbr^2$, $x' \in \bbbr^2$, $t \in \bbbr$ and
$t' \in \bbbr$.

As will be described in more detail in
Section~\ref{sec-geom-interpret}, this way of composing the four
different types primitive image transformations,
geometrically corresponds to
interpreting:
\begin{itemize}
\item
  the $2 \times 2$ affine transformation matrix $A$ as an orthonormal projection of
  surface patterns from the tangent plane of a local surface patch
  to a plane, that is parallel with the image plane of the observer,
\item
  the velocity vector $u = (u_1, u_2)^T \in \bbbr^2$ as
  the projection of the 3-D motion vector
  $U = (U_1, U_2, U_3)^T$ of local surface patterns onto a plane, that is
  parallel to the image plane, by local orthonormal projection,
\item
  the spatial scaling factor $S_x \in \bbbr_+$ as corresponding to the perspective
  scaling factor proportional to the inverse depth $Z$, which will
  then affect both the projection of a spatial surface pattern and the
  magnitude of the perceived motion in the image plane, and
\item
  the temporal scaling factor $S_t \in \bbbr_+$ as capturing a variability of
  similar spatio-temporal events that may occur either faster or
  slower, when observing different instances of a similar event at
  different occasions.
\end{itemize}
In this way, the composed image transformation model captures the variabilities
of the scaled orthographic projection model, complemented with a
variability over projections of 3-D motions between an observed object
and the observer, including spatio-temporal
events that may occur faster or slower relative to a reference view.

In the following, we will derive a joint covariance property for the
spatio-temporal scale-space representation obtained by convolution
with the studied class of spatio-temp\-oral smoothing kernels,
under the above class of composed
spatio-temporal image transformations. By necessity, parts of this
treatment may be somewhat technical. For the reader, who may be
more interested in the final result, than the details of the
derivation, it should be possible to, without major loss of continuity,
skip the details, and then proceed to the below
boldface header ``Summary of main result'',
for a condensed summary of the resulting joint covariance property.

\subsection{Joint transformation property of purely spatio-temporally
  smoothed image data}
\label{eq-transf-prop-spat-temp-smooth}

{\bf Prerequisites:} Let us assume that we have two video sequences or
video streams
$f \colon \bbbr^2 \times \bbbr \rightarrow \bbbr$
and
$f' \colon \bbbr^2 \times \bbbr \rightarrow \bbbr$,
that are related according to (\ref{eq-x-transf}) and (\ref{eq-t-transf}), such that
\begin{equation}
  \label{eq-f-fprim-transf-proof}
  f'(x', t') = f(x, t)
\end{equation}
for all $x = (x_1, x_2)^T \in \bbbr^2$ and $t \in \bbbr$ at all
points $p = (x_1, x_2, t)^T \in \bbbr^3$,
with these coordinates interpreted as local coordinates in some
local region $\Omega$ in joint space-time,  around the origin
$O = (0, 0, 0)^T$, assumed to correspond to the image point
$x = (0, 0)$, and the temporal moment $t = 0$ corresponding to
the time moment when a particular
receptive field response is computed.

What we want to derive, is a relationship for how the scale-space
representations
$L \colon \bbbr^2 \times \bbbr \times \bbbr_+ \times \bbbs_+^2 \times \bbbr_+
\times \bbbr^2 \rightarrow \bbbr$
and
$L' \colon \bbbr^2 \times \bbbr \times \bbbr_+ \times \bbbs_+^2 \times \bbbr_+
\times \bbbr^2 \rightarrow \bbbr$
of $f'$ and $f$, respectively, around this point in joint
space-time are related, when they are
defined according to
\begin{align}
  \begin{split}
     & L(x, t;\; s, \Sigma, \tau, v) = 
  \end{split}\nonumber\\
  \begin{split}
    \label{eq-scsp-repr-L-in-proof}
     & = \int_{\xi \in \bbbr^2} \int_{\eta \in \bbbr}
               T(\xi, \eta;\; s, \Sigma, \tau, v) \, f(x - \xi, t - \eta) \,
               d\xi \, d\eta,
  \end{split}\\
  \begin{split}
     & L'(x', t';\; s', \Sigma', \tau', v') = \\
  \end{split}\nonumber\\
  \begin{split}
      & = \int_{\xi' \in \bbbr^2} \int_{\eta' \in \bbbr}
                 T(\xi', \eta';\; s', \Sigma', \tau', v') \times
 \end{split}\nonumber\\
  \begin{split}
    \label{eq-scsp-repr-Lprim-in-proof}    
    & \phantom{= \int_{\xi' \in \bbbr^2} \int_{\eta' \in \bbbr}} \quad
                 f'(x' - \xi', t' - \eta') \,
                d\xi' \, d\eta'.
  \end{split}
\end{align}
{\bf Step~I:\/} Let us first consider how the velocity-adapted Gaussian kernel
$g(x' - v' t';\; s' \, \Sigma')$ transforms in the expression for
the spatio-temporal receptive field
according to (\ref{eq-spat-temp-RF-model})
\begin{equation}
  \label{eq-spat-temp-RF-model-prim}
  T(x', t';\; s', \Sigma', \tau', v') 
  = g(x' - v' t';\; s' \, \Sigma') \, h(t';\; \tau').
\end{equation}
Expanding the expression for the function $g(x' - v' t';\; s' \, \Sigma')$
according to the definition (\ref{eq-gauss-fcn-2D}) of the 2-D affine
Gaussian kernel, and making use of the explicit expressions for the
spatio-temporal image transformation in (\ref{eq-x-transf}) and (\ref{eq-t-transf}), then gives
\begin{align}
   \begin{split}
     & g(x' - v' t';\; s' \, \Sigma') =
   \end{split}\nonumber\\
   \begin{split}
     & = \frac{1}{2 \pi s' \sqrt{\det{\Sigma'}}} \,
            e^{-(x' - v' t')^T {\Sigma'}^{-1} (x' - v' t')/2 s'}
   \end{split}\nonumber\\
   \begin{split}
     & = \frac{1}{2 \pi s' \sqrt{\det{\Sigma'}}} \times
   \end{split}\nonumber\\
  \begin{split}
      & \phantom{=} \quad
            e^{-(S_x (A x + u t) - v' S_t t)^T {\Sigma'}^{-1} (S_x (A x + u t) - v' S_t t)/2 s'}.
   \end{split}
\end{align}
If we, inspired by the transformation property
of the spatial scale parameters under a spatial
scaling transformation in (\ref{eq-transf-pars-spat-scal}),
introduce a similar relationship
\begin{equation}
    s' = S_x^2 \, s,
  \end{equation}
as well as inspired by the transformation property
of the spatial covariance matrices $\Sigma \in \bbbs_+^2$
and $\Sigma' \in \bbbs_+^2$ under a spatial affine
transformation in (\ref{eq-transf-pars-spat-aff}),
introduce a similar relationship
\begin{equation}
  \Sigma' = A \, \Sigma \, A^T,
\end{equation}
which gives
\begin{equation}
  {\Sigma'}^{-1} = (A \, \Sigma A^T)^{-1} = A^{-T} \, \Sigma^{-1} \, A^{-1},
\end{equation}
as well as
\begin{equation}
  \det \Sigma' = | \det A |^2 \, \det \Sigma,
\end{equation}
we then obtain
\begin{align}
   \begin{split}
     & g(x' - v' t';\; s' \, \Sigma') =
   \end{split}\nonumber\\
   \begin{split}
     & = \frac{1}{2 \pi \, S_x^2 \, s \, | \det A |\sqrt{\det{\Sigma}}} \times
   \end{split}\nonumber\\
  \begin{split}
      & \phantom{=}
            e^{-(A^{-1}(S_x (A x + u t) -v' S_t t))^T \Sigma^{-1}  (A^{-1}(S_x (A x + u t) -v' S_t t))/(2 S_x^2 s)}.
   \end{split}
\end{align}
Notably, this expression can be written as
\begin{align}
   \begin{split}
     & g(x' - v' t';\; s' \, \Sigma') =
   \end{split}\nonumber\\
  \begin{split}
     \label{eq-transf-gauss-over-prime}
     & = \frac{1}{2 \pi \, s \, S_x^2 \, | \det A |\sqrt{\det{\Sigma}}} 
            e^{-(x - v t)^T \Sigma^{-1} (x - v t)/2 s},
   \end{split}
\end{align}
provided that the velocity parameters $v \in \bbbr^2$ and
$v' \in \bbbr^2$ in the two domains are related according to
\begin{equation}
  - v = A^{-1} u - A^{-1} v' S_t/S_x,
\end{equation}
in other words if
\begin{equation}
  \frac{S_t}{S_x} A^{-1} \, v' = v + A^{-1} u,
\end{equation}
that is provided that the velocity parameters $v'$ and $v$ are related
according to
\begin{equation}
  v' = \frac{S_x}{S_t} (A \, v + u).
\end{equation}
{\bf Step~II:\/} Consider next the spatio-temporal scale-space representation $L'$ of $f'$
according to (\ref{eq-scsp-repr-Lprim-in-proof}), and perform the
change of variables
\begin{align}
  \begin{split}
     \label{eq-xi-transf}
     \xi' = S_x \, (A \,  \xi + u \, t),
   \end{split}\\
  \begin{split}
     \label{eq-eta-transf}
     \eta' = S_t \, \eta,
   \end{split}
\end{align}
which for $d\xi' = d\xi'_1 \, d\xi'_2$ and $d\xi = d\xi_1 \, d\xi_2$ 
gives
\begin{align}
  \begin{split}
     \label{eq-dxi-transf}
     d\xi' = S_x^2 \, | \det A | \, d\xi,
   \end{split}\\
  \begin{split}
     \label{eq-deta-transf}
     d\eta' = S_t \, d\eta.
   \end{split}
\end{align}
Let us additionally, in the convolution integral (\ref{eq-scsp-repr-Lprim-in-proof}),
transform the kernel $T(x', t';\; s', \Sigma',
\tau', v')$ according to (\ref{eq-spat-temp-RF-model-prim}), with
its components $g(x' - v' t';\; s' \, \Sigma')$ transforming according to
(\ref{eq-transf-gauss-over-prime}), and $h(t';\; \tau')$ transforming
according to (\ref{eq-temp-sc-cov-temp-kernel}), with the parameters
of the receptive fields transforming according to
\begin{align}
   \begin{split}
     s' & = S_x^2 \, s,
   \end{split}\\
    \begin{split}
     \Sigma' & = A \, \Sigma \, A^{T},
   \end{split}\\
   \begin{split}
     \tau' & = S_t^2 \, \tau,
   \end{split}\\
   \begin{split}
     v' & = \frac{S_x}{S_t} (A \, v + u),
   \end{split}
\end{align}
which then gives
\begin{equation}
  \label{eq-transf-T-in-proof}
  T(x', t';\; s', \Sigma', \tau', v')
  = \frac{1}{S_x^2 \, |\det A| \, S_t} \, T(x, t;\; s, \Sigma, \tau, v).
\end{equation}
{\bf Step~III:\/} Thus, given that the functions $f$ and $f'$
transform according to (\ref{eq-f-fprim-transf-proof}),
combined with the previous result, that $T$ transforms
according to (\ref{eq-transf-T-in-proof}),
as well as that $d\xi$, $d\xi'$, $d\eta$ and $d\eta'$ transform
according to (\ref{eq-dxi-transf}) and (\ref{eq-deta-transf}),
we obtain
\begin{align}
  \begin{split}  
     & L'(x', t';\; s', \Sigma', \tau', v') = \\
  \end{split}\nonumber\\
  \begin{split}
      & = \int_{\xi' \in \bbbr^2} \int_{\eta' \in \bbbr}
                 T(\xi', \eta';\; s', \Sigma', \tau', v') \times
 \end{split}\nonumber\\
  \begin{split}
    & \phantom{= \int_{\xi' \in \bbbr^2} \int_{\eta' \in \bbbr}} \quad
                 f'(x' - \xi', t' - \eta') \,
                d\xi' \, d\eta' =
  \end{split}\nonumber\\
  \begin{split}
     & = \int_{\xi \in \bbbr^2} \int_{\eta \in \bbbr}
               T(\xi, \eta;\; s, \Sigma, \tau, v) \, f(x - \xi, t - \eta) \,
               d\xi \, d\eta
  \end{split}\nonumber\\
  \begin{split}
     & = L(x, t;\; s, \Sigma, \tau, v).
  \end{split}
\end{align}

\begin{figure*}[hbt]
  \[
    \begin{CD}
       \hspace{0mm}L(x, t;\; s, \Sigma, \tau, v)
       @>{\footnotesize\begin{array}{c} x' = S_x (A \, x + u \, t)  \\ t' =
                         S_t \, t  \\ s' = S_x^2 \, s \\ \Sigma' = A \,
                         \Sigma \, A^{T} \\ \tau' = S_t^2 \, \tau \\
                         v' = \frac{S_x}{S_t} (A \, v + u) \end{array}}>> L'(x', t';\; s', \Sigma', \tau', v') \\
       \Big\uparrow\vcenter{\rlap{$\scriptstyle{{*T(x, t;\; s,
               \Sigma, \tau, v)}}$}} & &
       \Big\uparrow\vcenter{\rlap{$\scriptstyle{{*T(x', t';\; s',
               \Sigma', \tau', v')}}$}} \\
       f(x, t) @>{\footnotesize \begin{array}{c} x' = S_x (A \, x + u \, t)
                                  \\ t' = S_t \, t \end{array}}>> f'(x', t')
    \end{CD}
  \]
\caption{Commutative diagram for the joint spatio-temporal smoothing
  component (\ref{eq-spat-temp-RF-model}) in the joint spatio-temporal receptive field
  model (\ref{eq-spat-temp-RF-model-der}) under the composition of (i)~a
  spatial scaling transformation, (ii)~a spatial affine transformation, (iii)~a
  Galilean transformation and (iv)~a temporal scaling transformation
  according to (\ref{eq-x-transf}) and (\ref{eq-t-transf}). This
  commutative diagram, which should be read from the lower left corner to the
  upper right corner, means that irrespective of whether the input video 
  sequence or video stream $f(x, t)$ is first subject to the composed transformation
  $x' = S_x (A \,  x + u \, t)$ and $t' = S_t \, t$
  and then smoothed with a spatio-temporal kernel 
  $T(x', t';\; s', \Sigma', \tau', v')$, or instead directly convolved
  with the spatio-temporal smoothing kernel
  $T(x, t;\; s, \Sigma, \tau, v)$ and then
  subject to the same joint spatio-temporal transformation, we do then get the same
  result, provided that the parameters of the spatio-temporal
  smoothing kernels are related
  according to $s' = S_x^2 \, s$, $\Sigma' = A \, \Sigma \, A^{T}$,
  $\tau' = S_t^2 \, \tau$ and $v' = \frac{S_x}{S_t} (A \, v + u)$.}
\label{fig-comm-diag-spat-temp-smooth}
\end{figure*}

\noindent
{\bf Summary of main result:} To conclude, given two video sequences
or video streams $f'$ and $f$, that are
related according to
\begin{equation}
  f'(x', t') = f(x, t)
\end{equation}
under a composed image transformation of the form
\begin{align}
  \begin{split}
    \label{eq-composed-sapt-temp-transf-summ-xprime}
     x' = S_x \,  (A \,  x + u \, t),
   \end{split}\\
  \begin{split}
    \label{eq-composed-sapt-temp-transf-summ-tprime}    
     t' = S_t \, t,
   \end{split}
\end{align}
we have shown that the corresponding spatio-temporal scale-space
representations $L'$ and $L$ of $f'$ and $f$, respectively, are
related according to
\begin{equation}
  \label{eq-joint-cov-prop-result-of-proof}
  L'(x', t';\; s', \Sigma', \tau', v') = L(x, t;\; s, \Sigma, \tau, v),
\end{equation}
provided that the parameters of the receptive fields transform according to
\begin{align}
  \begin{split}
    \label{eq-s-transf-result}
    s' & = S_x^2 \, s,
  \end{split}\\
  \begin{split}
    \label{eq-Sigma-transf-result}
    \Sigma' & = A \, \Sigma \, A^{T},
  \end{split}\\
  \begin{split}
    \label{eq-tau-transf-result}
    \tau' & = S_t^2 \, \tau,
  \end{split}\\
  \begin{split}
    \label{eq-v-transf-result}    
    v' & = \frac{S_x}{S_t} (A \, v + u), 
  \end{split}
\end{align}
which proves the joint spatio-temporal covariance property, see
Figure~\ref{fig-comm-diag-spat-temp-smooth} for a commutative diagram
that illustrates this joint covariance property.

Notably, this result also serves as an explicit proof of all the individual
transformation properties in Lindeberg
(\citeyear{Lin23-FrontCompNeuroSci}), where the explicit proofs were
omitted there, because of lack of space.

\begin{figure*}[hbt]
  \[
    \begin{CD}
       \hspace{0mm}\nabla_x \partial_t L(x, t;\; s, \Sigma, \tau, v)
       @>{\footnotesize\begin{array}{c} x' = S_x (A \, x + u \, t)  \\ t' =
                         S_t \, t  \\ s' = S_x^2 \, s \\ \Sigma' = A \,
                         \Sigma \, A^{T} \\ \tau' = S_t^2 \, \tau \\
                         v' = \frac{S_x}{S_t} (A \, v + u) \\ \nabla_{x'} =
    \frac{1}{S_x} \, A^{-T} \, \nabla_{x} \\ \partial_{t'} = -
                         \frac{1}{S_x}  \, u^T A^{-T} \, \nabla_x +
                         \frac{1}{S_t} \, \partial_t \end{array}}>> \nabla_{x'} \partial_{t'} L'(x', t';\; s', \Sigma', \tau', v') \\
       \Big\uparrow\vcenter{\rlap{$\scriptstyle{{* (\nabla_x \partial_t T)(x, t;\; s,
               \Sigma, \tau, v)}}$}} & &
       \Big\uparrow\vcenter{\rlap{$\scriptstyle{{*( \nabla_{x'} \partial_{t'} T)(x', t';\; s',
               \Sigma', \tau', v')}}$}} \\
       f(x, t) @>{\footnotesize \begin{array}{c} x' = S_x (A \, x + u \, t)
                                  \\ t' = S_t \, t \end{array}}>> f'(x', t')
    \end{CD}
  \]
\caption{Commutative diagram for spatio-temporal derivative operators
  underlying the joint spatio-temporal receptive field
  model (\ref{eq-spat-temp-RF-model-der}) under the composition of (i)~a
  spatial scaling transformation, (ii)~a spatial affine transformation, (iii)~a
  Galilean transformation and (iv)~a temporal scaling transformation
  according to (\ref{eq-x-transf}) and (\ref{eq-t-transf}). This
  commutative diagram, which should be read from the lower left corner to the
  upper right corner, means that irrespective of whether the input video 
  sequence or video stream $f(x, t)$ is first subject to the composed transformation
  $x' = S_x (A \,  x + u \, t)$ and $t' = S_t \, t$
  and then filtered with a spatio-temporal derivative kernel
  $(\nabla_{x'} \partial_{t'} T)(x', t';\; s', \Sigma', \tau', v')$, or instead directly convolved
  with the spatio-temporal smoothing kernel
  $(\nabla_x \partial_t T)(x, t;\; s, \Sigma, \tau, v)$ and then
  subject to the same joint spatio-temporal transformation, we do then get the same
  result, provided that the spatial and the temporal derivative
  operators are transformed according to
  $\nabla_{x'} = \frac{1}{S_x} \, A^{-T} \, \nabla_{x}$ and
  $\partial_{t'} = - \frac{1}{S_x}  \, u^T A^{-T} \, \nabla_x + \frac{1}{S_t} \, \partial_t$
  and that the parameters of the spatio-temporal
  smoothing kernels are related
  according to $s' = S_x^2 \, s$, $\Sigma' = A \, \Sigma \, A^{T}$,
  $\tau' = S_t^2 \, \tau$ and $v' = \frac{S_x}{S_t} (A \, v + u)$.
  (In this commutative diagram, we have illustrated the general
  covariance properties of spatio-temporal derivatives for the
  particular choice of the
  composed spatio-temporal derivative operator $\nabla_x \partial_t T$
  in the spatio-temporal receptive field model (
  \ref{eq-spat-temp-RF-model-der}). Similar
covariance properties can, of course, also be obtained for
other combinations of the spatial and the temporal derivative
operators $\nabla_x$ and $\partial_t$, in a structurally similar manner.)}
\label{fig-comm-diag-spat-temp-ders}
\end{figure*}

\begin{figure*}[hbt]
  \[
    \begin{CD}
       \hspace{0mm}\nabla_{x,\affnormtiny} \partial_{{\bar t},\normtiny} L(x, t;\; s, \Sigma, \tau, v)
       @>{\footnotesize\begin{array}{c} x' = S_x (A \, x + u \, t)  \\ t' =
                         S_t \, t  \\ s' = S_x^2 \, s \\ \Sigma' = A \,
                         \Sigma \, A^{T} \\ \tau' = S_t^2 \, \tau \\
                         v' = \frac{S_x}{S_t} (A \, v + u) \\
                         \nabla_{x,\affnormtiny} = s^{1/2} \,
                         \Sigma^{1/2} \, \nabla_x \\ \nabla_{x',\affnormtiny} = {s'}^{1/2} \,
                         {\Sigma'}^{1/2} \, \nabla_{x'} \\ \nabla_{x',\affnormtiny} =
    \tilde{\rho} \, \nabla_{x,\affnormtiny} \\ \partial_{{\bar t}',\normtiny} = 
                         \partial_{{\bar t},\normtiny} \end{array}}>>
                     \nabla_{x',\affnormtiny} \partial_{{\bar t}',\normtiny} L'(x', t';\; s', \Sigma', \tau', v') \\
       \Big\uparrow\vcenter{\rlap{$\scriptstyle{{*
               (\nabla_{x,\affnormtiny} \partial_{{\bar t},\normtiny} T)(x, t;\; s,
               \Sigma, \tau, v)}}$}} & &
       \Big\uparrow\vcenter{\rlap{$\scriptstyle{{*(
               \nabla_{x',\affnormtiny} \partial_{{\bar t}',\normtiny} T)(x', t';\; s',
               \Sigma', \tau', v')}}$}} \\
       f(x, t) @>{\footnotesize \begin{array}{c} x' = S_x (A \, x + u \, t)
                                  \\ t' = S_t \, t \end{array}}>> f'(x', t')
    \end{CD}
  \]
\caption{Commutative diagram for scale-normalized spatio-temporal derivative operators
  defined from the joint spatio-temporal receptive field
  model (\ref{eq-spat-temp-RF-model-der}) under the composition of (i)~a
  spatial scaling transformation, (ii)~a spatial affine transformation, (iii)~a
  Galilean transformation and (iv)~a temporal scaling transformation
  according to (\ref{eq-x-transf}) and (\ref{eq-t-transf}). This
  commutative diagram, which should be read from the lower left corner to the
  upper right corner, means that irrespective of whether the input video 
  sequence or video stream $f(x, t)$ is first subject to the composed transformation
  $x' = S_x (A \,  x + u \, t)$ and $t' = S_t \, t$
  and then filtered with a scale-normalized spatio-temporal derivative kernel
  $(\nabla_{x',\affnorm} \partial_{t',\norm} T)(x', t';\; s', \Sigma', \tau', v')$,
  or instead directly convolved
  with the scale-normalized spatio-temporal smoothing kernel
  $(\nabla_{x,\affnorm} \partial_{t,\norm} T)(x, t;\; s, \Sigma, \tau, v)$ and then
  subject to the same joint spatio-temporal transformation, we do
  then, up to a possibly unknown rotation transformation $\tilde{\rho}$, get the same
  result, provided that the parameters of the spatio-temporal
  smoothing kernels are related
  according to $s' = S_x^2 \, s$, $\Sigma' = A \, \Sigma \, A^{T}$,
  $\tau' = S_t^2 \, \tau$ and $v' = \frac{S_x}{S_t} (A \, v + u)$.
  Note, in particular, the conceptual simplification in relation to
  the corresponding commutative diagram based on regular partial
  derivatives that have not been subject to scale normalization or
  velocity adaptation regarding the temporal derivatives, in that the
  scale-normalized spatio-temporal derivatives in this commutative
  diagram are essentially equal, up to a possibly unknown rotation transformation.
  (In this commutative diagram, we have illustrated the general
  covariance properties of spatio-temporal derivatives for the
  particular choice of the
  composed spatio-temporal derivative operator
  $\nabla_{x,\affnorm} \partial_{t,\norm} T$
  in the spatio-temporal receptive field model
  (\ref{eq-spat-temp-RF-model-der}). Similar
covariance properties can, of course, also be obtained for
other selections of the spatial and the temporal derivative
operators $\nabla_{x,\affnorm}$ and $\partial_{t,\norm}$
for which corresponding covariance properties hold.)}
\label{fig-comm-diag-spat-temp-ders-sc-norm}
\end{figure*}

\subsection{Joint transformation properties of spatio-temporal
  derivatives}
\label{eq-transf-prop-spat-temp-ders}

Let us denote the spatio-temporal coordinates for the original and the
transformed domains by $p = (x_1, x_2, t)^T \in \bbbr^3$ and
$p' = (x'_1, x'_2, t')^T \in \bbbr^3$,
respectively, and let us denote the components of the $2 \times 2$ affine
transformation matrix $A$ by $a_{ij}$ for $i$ and $j$ $\in \{ 1, 2\}$
and again let the velocity vector in the Galilean transformation
be $u = (u_1, u_2)^T$. Then, the composed image transformation according
to (\ref{eq-x-transf}) and (\ref{eq-t-transf}) can be written
\begin{align}
  \begin{split}
     p'
     & = \left( \begin{array}{c} x_1' \\ x_2' \\ t' \end{array} \right)
         = S_x \,
            \left(
               \begin{array}{ccc}
                   a_{11} & a_{12} & u_1 \\
                   a_{21} & a_{22} & u_2 \\
                   0        & 0        & S_t/S_x \\
               \end{array}
            \right)
            \left( \begin{array}{c} x_1 \\ x_2 \\ t \end{array} \right)
  \end{split}\nonumber\\
  \begin{split}
      & = Q \, p,
   \end{split}
\end{align}
where $Q$ is a $3 \times 3$ joint transformation matrix, that operates on the
spatio-temporal coordinates $p$.

According to the general transformation property of derivative
operators under a linear change of variables between the domains
$p = (x_1, x_2, t)^T$ and $p' = (x'_1, x'_2, t')^T$, which in terms of
explicit partial derivatives can be expressed on the form
\begin{align}
  \begin{split}
    \partial_{x_1}
    & = \frac{\partial x_1'}{\partial x_1} \, \partial_{x_1'}
           + \frac{\partial x_2'}{\partial x_1} \, \partial_{x_2'}
           + \frac{\partial t'}{\partial x_1} \, \partial_{t'}
   \end{split}\\
  \begin{split}
    \partial_{x_2}
    & = \frac{\partial x_1'}{\partial x_2} \, \partial_{x_1'}
           + \frac{\partial x_2'}{\partial x_2} \, \partial_{x_2'}
           + \frac{\partial t'}{\partial x_2} \, \partial_{t'}
    \end{split}\\
    \begin{split}
    \partial_{t}
    & = \frac{\partial x_1'}{\partial t} \, \partial_{x_1'}
           + \frac{\partial x_2'}{\partial t} \, \partial_{x_2'}
           + \frac{\partial t'}{\partial t} \, \partial_{t'},
   \end{split}
\end{align}
it then follows that
the spatio-temporal derivative operators in the original and the
transformed domains are related according to
\begin{align}
  \begin{split}
     \nabla_{p}
     & = \left( \begin{array}{c} \partial_{x_1} \\ \partial_{x_2} \\ \partial_t \end{array} \right)
      = S_x
        \left(
          \begin{array}{ccc}
             a_{11} & a_{21} & 0 \\
             a_{12} & a_{22} & 0 \\
             u_1    & u_2    & S_t/S_x \\
          \end{array}
       \right)
       \left( \begin{array}{c} \partial_{x_1'} \\ \partial_{x_2'} \\ \partial_{t'} \end{array} \right)
  \end{split}\nonumber\\
  \begin{split}
     \label{eq-transf-spat-temp-grad}
      & = Q^T \, \nabla_{p'},
    \end{split}
\end{align}
which, in turn, gives the following explicit transformation property for the
spatio-temporal derivative operator under the
inverse composed spatio-temporal transformation
\begin{equation}
    \label{eq-transf-spat-temp-grad-inv}
   \nabla_{p'} = Q^{-T} \, \nabla_{p}
\end{equation}
with
\begin{multline}
  Q^{-T}
  =
  \frac{1}{S_x\, \det A} \times \\
  \left(
    \begin{array}{ccc}
        a_{22}                    & -a_{21}                  & 0 \\
      -a_{12}                    & a_{11}                    & 0 \\
      \, - a_{22} \, u_1 + a_{12} \, u_2 \, & \, a_{21} \, u_1 - a_{11} \, u_2 \, & \, S_x \det A / S_t \, 
    \end{array}
  \right).
\end{multline}
In terms of vector notation, after introducing the purely spatial gradient operators
$\nabla_x = (\partial_{x_1}, \partial_{x_2})^T$ and
$\nabla_{x'} = (\partial_{x_1'}, \partial_{x_2'})^T$,
the transformation property (\ref{eq-transf-spat-temp-grad}) of
the spatio-temporal gradient operators can then be written as
\begin{align}
  \begin{split}
    \label{eq-joint-transf-prop-spat-grad}
    \nabla_{x} = S_x \, A^T \, \nabla_{x'},
  \end{split}\\
  \begin{split}
    \label{eq-joint-transf-prop-temp-der}  
    \partial_{t} = S_x \, u^T \, \nabla_{x'}  + S_t \, \partial_{t'},
  \end{split} 
\end{align}
whereas the corresponding inverse relationship
(\ref{eq-transf-spat-temp-grad-inv}) can be expressed as
\begin{align}
  \begin{split}
    \label{eq-joint-transf-prop-spat-grad-inv}
    \nabla_{x'} = \frac{1}{S_x} \, A^{-T} \, \nabla_{x},
  \end{split}\\
  \begin{split}
    \label{eq-joint-transf-prop-temp-der-inv}      
    \partial_{t'} = - \frac{1}{S_x}  \, u^T A^{-T} \, \nabla_x + \frac{1}{S_t} \, \partial_t.
  \end{split} 
\end{align}
Based on these relations, expressions for spatio-temporal derivatives
can be transformed between the two domains
under the composed image transformation, thus extending the
transformation property (\ref{eq-joint-cov-prop-result-of-proof})
of the spatio-temporal receptive fields to beyond the effect of purely
spatio-temporal smoothing operation in the spatio-temporal receptive
fields also cover the spatio-temporal derivative operators in the
composed spatio-temporal receptive field model of the form
(\ref{eq-spat-temp-RF-model-der}), see
Figure~\ref{fig-comm-diag-spat-temp-ders} for a commutative diagram
that illustrates this joint covariance property.

\subsection{Transformation properties of geometrically defined
  spatio-temporal derivative operators}
\label{sec-transf-props-geom-spat-temp-ders}

Beyond the above, essentially partial-derivative-based spatial and
temporal derivative operators, it can often be convenient to also introduce
more geometrically defined spatio-temporal derivative operators.

For example, given the vector notation for the derivative operators,
the velocity-adapted derivative operators
corresponding to (\ref{eq-vel-adapt-der-def}) are with
$v = (v_1, v_2)^T$ and $v' = (v_1', v_2')^T$ given by
\begin{equation}
  \label{eq-def-vel-adapt-ders-both-domains-repeated}
  \partial_{\bar t} = v^T \, \nabla_x + \partial_t
  \quad\mbox{and}\quad
  \partial_{\bar t'} = {v'}^T \, \nabla_{x'} + \partial_{t'},
\end{equation}
where the velocity vectors $v$ and $v'$ are related according to
(\ref{eq-v-transf-result}). Such velocity-adapted spatio-temporal
derivative operators are natural to use, when computing spatio-temporal
receptive responses from moving image structures. Specifically, there
are velocity-sensitive receptive fields in the primary cortex that can
be rather well modelled by such spatio-temporal derivatives; see
Figure~18 in Lindeberg (\citeyear{Lin21-Heliyon}).

By combining the transformation properties of the spatial and temporal
derivative operators according to
(\ref{eq-joint-transf-prop-spat-grad-inv})
and (\ref{eq-joint-transf-prop-temp-der-inv}) with the
transformation property (\ref{eq-v-transf-result}) of the velocity
parameters $v$ and $v'$ in the receptive fields, we can thus obtain
explicit expressions for how such velocity-adapted receptive fields
are transformed in a Galilean covariant way, under relative motions
between the objects in the world and the observer.
Specifically, inserting the expressions (\ref{eq-joint-transf-prop-spat-grad})
and (\ref{eq-joint-transf-prop-temp-der}) for the spatial
gradient operator $\nabla_x$ and the regular temporal derivative operator
$\partial_t$ as well as the velocity vector $v$ obtained by solving
for this velocity vector as function of the transformed velocity
vector $v'$ in the transformation property (\ref{eq-v-transf-result}) of the velocity
vector under the composed spatio-temporal transformation defined
by (\ref{eq-composed-sapt-temp-transf-summ-xprime})
and (\ref{eq-composed-sapt-temp-transf-summ-tprime}),
does
after simplification of this expression lead to the following simple relationship
\begin{equation}
  \label{eq-equal-veladapt-ders-composed-transf-main-result}
  \partial_{\bar t} = S_t \, \partial_{{\bar t}'}.
\end{equation}
In this way, the
velocity-adapted derivatives constitute a geometrically very meaningful way
to define spatio-temporal derivative responses on image observations
of a dynamic world.

Similarly, the directional derivative operators $\partial_{\varphi}$ and $\partial_{\varphi'}$
corresponding to (\ref{eq-dir-der-def}) are, with the unit vectors
\begin{equation}
  e_{\varphi} = (\cos \varphi, \sin \varphi)^T
  \quad\mbox{and}\quad
  e_{\varphi'} = (\cos \varphi', \sin \varphi')^T,
\end{equation}
given by 
\begin{equation}
  \label{eq-def-dir-ders-composed}
  \partial_{\varphi} = e_{\varphi}^T \, \nabla_x
  \quad\mbox{and}\quad
  \partial_{\varphi'} = e_{\varphi'}^T \, \nabla_{x'}.
\end{equation}
With regard to the composed spatio-temporal image transformation
(\ref{eq-composed-sapt-temp-transf-summ-xprime}), 
inserting the expression (\ref{eq-joint-transf-prop-spat-grad}) for
the spatial gradient operator $\nabla_x$ into the definition
(\ref{eq-def-dir-ders-composed}), and defining the transformed unit
vector $e_{\varphi'}$ as
\begin{equation}
  e_{\varphi'}
  = \frac{S_x \, A \, e_{\varphi}}{\| S_x \, A \,  e_{\varphi} \|}
  = \frac{A \, e_{\varphi}}{\| A \,  e_{\varphi} \|}   ,
\end{equation}
implies that the directional derivative operator transforms according to
\begin{equation}
  \partial_{\varphi} = \| S_x \, A \, e_{\varphi} \| \, \partial_{\varphi'}.
\end{equation}
In the special case, when the composed affine transformation matrix
$S_x \, A$ is a pure rotation matrix $S_x \, A = R_{\theta}$,
the eigenvectors of the spatial covariance
matrix $\Sigma$ in the spatio-temporal smoothing kernel
do also transform according to a rotation, according
to (\ref{eq-Sigma-transf-result}), implying that the rotational angles
$\varphi$ and $\varphi'$ will be related according to
\begin{equation}
  \varphi' = \varphi + \theta,
\end{equation}
which with regard to the unit vectors used for defining the
directional derivatives, can in terms of matrix operations be accomplished
according to
\begin{equation}
   e_{\varphi'} = R_{\theta} \, e_{\varphi}.
\end{equation}
In these ways, we can in a rotationally covariant way transform the
responses of the spatial components of the spatio-temporal receptive
field model (\ref{eq-spat-temp-RF-model-der}) under transformations within
the similarity group over the image domain.
For more general affine transformations over the image domain, the
corresponding relations are, however, more complex.

By using these transformation properties of spatio-temporal
gradient operators, we can thus in a geometric way transform all the
spatio-temporal derivative operators in the spatio-temporal receptive field models
described in
Sections~\ref{sec-spat-temp-ders-gen-gauss-model} and~\ref{sec-rel-biol-vision}.

\subsection{Transformation properties of scale-normalized 
  spatio-temporal derivative operators}
\label{sec-transf-prop-sc-norm-spat-temp-ders}

With the introduction of scale-normalized derivatives according to
Section~\ref{sec-sc-norm-spat-temp-ders}, the transformation
properties of the spatio-temporal receptive fields can be further
simplified:
\begin{itemize}
\item
  If we require the family of affine transformation matrices $A$ to be reduced
  to the group of rotation matrices $A = R_{\theta}$, such that the composed effect of the
  spatial scaling factor $S_x$ and the rotation matrix $A = R_{\theta}$ spans
  the variability of the similarity group, then, based on the
  theoretical results in Section~\ref{sec-cov-prop-aff-sc-norm-dir-ders},
  the affine
  scale-normalized directional derivative operators in the direction
  $e_{\varphi} = (\cos \varphi, \sin \varphi)$ according to
  (\ref{eq-aff-sc-norm-dir-der})
  \begin{equation}
    \partial_{\varphi,\norm}^m
    = s^{m/2} \, (e_{\varphi}^T \, \Sigma \, e_{\varphi})^{m/2} \, \partial_{\varphi}^m.
  \end{equation}
  are covariant under the resulting similarity
  group extended with the group of Galilean transformations and the
  group of temporal scaling transformations, such that
  \begin{multline}
    \label{eq-sc-norm-dir-der-joint-cov-prop-sc-norm-ders}
    \partial_{\varphi',\norm}^m L'(x', t';\; s', \Sigma', \tau', v') = \\
    = \partial_{\varphi,\norm}^m L(x, t;\; s, \Sigma, \tau, v),
  \end{multline}
  provided that the scale parameters $s$ and $s'$ are matched with the
  effect of the scaling transformation, the orientation angles
  $\varphi$ and $\varphi'$ are
  matched with the effect of the rotation matrix $R_{\theta}$, and
  provided that the other parameters of the receptive fields are
  matched according to
  (\ref{eq-s-transf-result})--(\ref{eq-v-transf-result})
  such that
  \begin{align}
    \begin{split}
      s' & = S_x^2 \, s,
    \end{split}\\
    \begin{split}    
      \varphi' & = \varphi + \theta,
    \end{split}\\
    \begin{split}
      \Sigma' & = R_{\theta} \, \Sigma \, R_{\theta}^T,
    \end{split}\\
  \begin{split}
    \tau' & = S_t^2 \, \tau,
  \end{split}\\
  \begin{split}
    v' & = \frac{S_x}{S_t} (R_{\theta} \, v + u).
  \end{split}    
  \end{align}
\item
  If we consider the group of general
  affine transformation matrices $A$, and
  define the scale-normalized affine gradient vector
  according to (\ref{eq-def-sc-norm-aff-grad-op})
  \begin{equation}
    \nabla_{x,\affnorm} = s^{1/2} \, \Sigma^{1/2} \, \nabla_x.
  \end{equation}  
  the scale-normalized affine Hessian matrix
  according to (\ref{eq-def-sc-norm-aff-hess-mat})
  \begin{equation}
    {\cal H}_{x,\affnorm} = s \, (\Sigma^{1/2}) \, {\cal H}_x \, (\Sigma^{1/2})^T,
  \end{equation}
  then, based on the results in Section~\ref{sec-cov-prop-sc-norm-aff-grad-op} and
  Section~\ref{sec-cov-prop-sc-norm-aff-hess-mat},
  these scale-normalized affine
  derivative-based entities will be equal up to 
  rotation matrices $\tilde{\rho}$ according to
  \begin{multline}
    \label{eq-cov-prop-sc-norm-aff-grad-summ-overview}
    (\nabla_{x',\affnorm}  L')(x', t';\; s', \Sigma', \tau', v') = \\ =
    \tilde{\rho} \,  (\nabla_{x,\affnorm} L)(x, t;\; s, \Sigma, \tau, v)
  \end{multline}
  and
  \begin{multline}
    \label{eq-cov-prop-sc-norm-aff-hess-summ-overview}
    ({\cal H}_{x',\affnorm} L')(x', t';\; s', \Sigma', \tau', v') = \\
    = \tilde{\rho} \, ({\cal H}_{x,\affnorm} L)(x, t;\; s, \Sigma, \tau, v) \, \tilde{\rho}^T,
  \end{multline}
  provided that the scale parameters $s$ and $s'$ as well as the
  spatial covariance matrices $\Sigma$ and $\Sigma'$ are matched
  according to (\ref{eq-transf-prop-sc-par-spat-cov-mat-pure-aff-scsp-main-result})
  \begin{equation}
     s' \, \Sigma' = s \, (S_x \, A) \, \Sigma \, (S_x A)^T = s \, S_x^2 \, A \, \Sigma \, A^T,
   \end{equation}
  and provided that the other parameters of the receptive fields are
  matched according to
  (\ref{eq-tau-transf-result})--(\ref{eq-v-transf-result})
  \begin{align}
    \begin{split}
      \tau' & = S_t^2 \, \tau,
    \end{split}\\
    \begin{split}
      v' & = \frac{S_x}{S_t} (A \, v + u).
    \end{split}
  \end{align}
\item
  Irrespective of any restrictions on the family of affine
  transformation matrices $A$, the velocity-adapted temporal
  derivative operators according to
  (\ref{eq-def-vel-adapt-ders-both-domains-repeated})
  \begin{equation}
    \partial_{\bar t} = v^T \, \nabla_x + \partial_t
    \quad\mbox{and}\quad
    \partial_{\bar t'} = {v'}^T \, \nabla_{x'} + \partial_{t'},
  \end{equation}
  extended to scale-normalized velocity-adapted temporal derivatives
  according to (\ref{eq-def-sc-norm-vel-adapt-ders})
  \begin{equation}
    \label{eq-def-vel-adapt-ders-both-domains-sc-norm}
    \partial_{\bar t,\norm}^n = \tau^{n/2} \, \partial_{\bar t}^n
    \quad\mbox{and}\quad
    \partial_{\bar t',\norm}^n = {\tau'}^{n/2} \, \partial_{\bar t'}^n,
  \end{equation}
  will, based on the result underlying
  Equation~(\ref{eq-equal-veladapt-ders-composed-transf-main-result}),
  be equal
  \begin{multline}
    \label{eq-equal-veladapt-ders-composed-transf-main-result-sc-norm}
    \partial_{{\bar t}',\norm}^n L' (x', t';\; s', \Sigma', \tau', v')  = \\
    =  \partial_{{\bar t},\norm}^n L(x, t;\; s, \Sigma, \tau, v),
  \end{multline}
  provided that the parameters $s$, $s'$, $\Sigma$, $\Sigma'$,
  $\tau$, $\tau'$, $v$ and $v'$ of the receptive fields are matched
  according to Equations~(\ref{eq-s-transf-result})--(\ref{eq-v-transf-result}).
\end{itemize}
In these ways%
\footnote{In addition to the above three main cases treated here, it is also possible to
extend the covariance property of the affine scale-normalized
directional operator in
(\ref{eq-sc-norm-dir-der-joint-cov-prop-sc-norm-ders}) to the
spatio-temporal extension of the case with coupled
eigendecompositions of the affine covariance matrix $A$ and the
spatial covariance matrix $\Sigma$ studied in
Section~\ref{sec-aff-anal-dir-der-coupled-eigen-decomp}.
To save space, we do, however, leave out those details to the
reader.},
the derived joint covariance properties for the
spatial-temporal derivatives assume much simpler forms, when expressed
in terms of scale-normalized derivatives, by being essentially equal,
up to a possibly unknown rotation transformation;
see Figure~\ref{fig-comm-diag-spat-temp-ders-sc-norm} for an
illustration in terms of a commutative diagram.

In this context, it should furthermore be specifically noted that the
covariance properties of the spatial derivative operators in
(\ref{eq-sc-norm-dir-der-joint-cov-prop-sc-norm-ders}),
(\ref{eq-cov-prop-sc-norm-aff-grad-summ-overview})
and (\ref{eq-cov-prop-sc-norm-aff-hess-summ-overview})
can also be combined with the transformation property of the
velocity-adapted temporal derivative operator
in (\ref{eq-equal-veladapt-ders-composed-transf-main-result-sc-norm}),
to also formulate the transformation properties for the
corresponding combined spatio-temporal
derivative operators, of the forms
\begin{align}
  \begin{split}
    \partial_{\varphi,\norm}^m \, \partial_{{\bar t},\norm}^n,
  \end{split}\\
  \begin{split}
    \nabla_{x,\affnorm} \, \partial_{{\bar t},\norm}^n,
  \end{split}\\
  \begin{split}
    {\cal H}_{x,\affnorm} \, \partial_{{\bar t},\norm}^n.
  \end{split}
\end{align}
Thus, we can, based on the presented framework, express the covariance
properties for a very rich family of geometric scale-normalized 
spatio-temporal derivative operators.

\begin{figure}[hbtp]
  \vspace{-35mm}
  \begin{center}
     \includegraphics[width=0.50\textwidth]{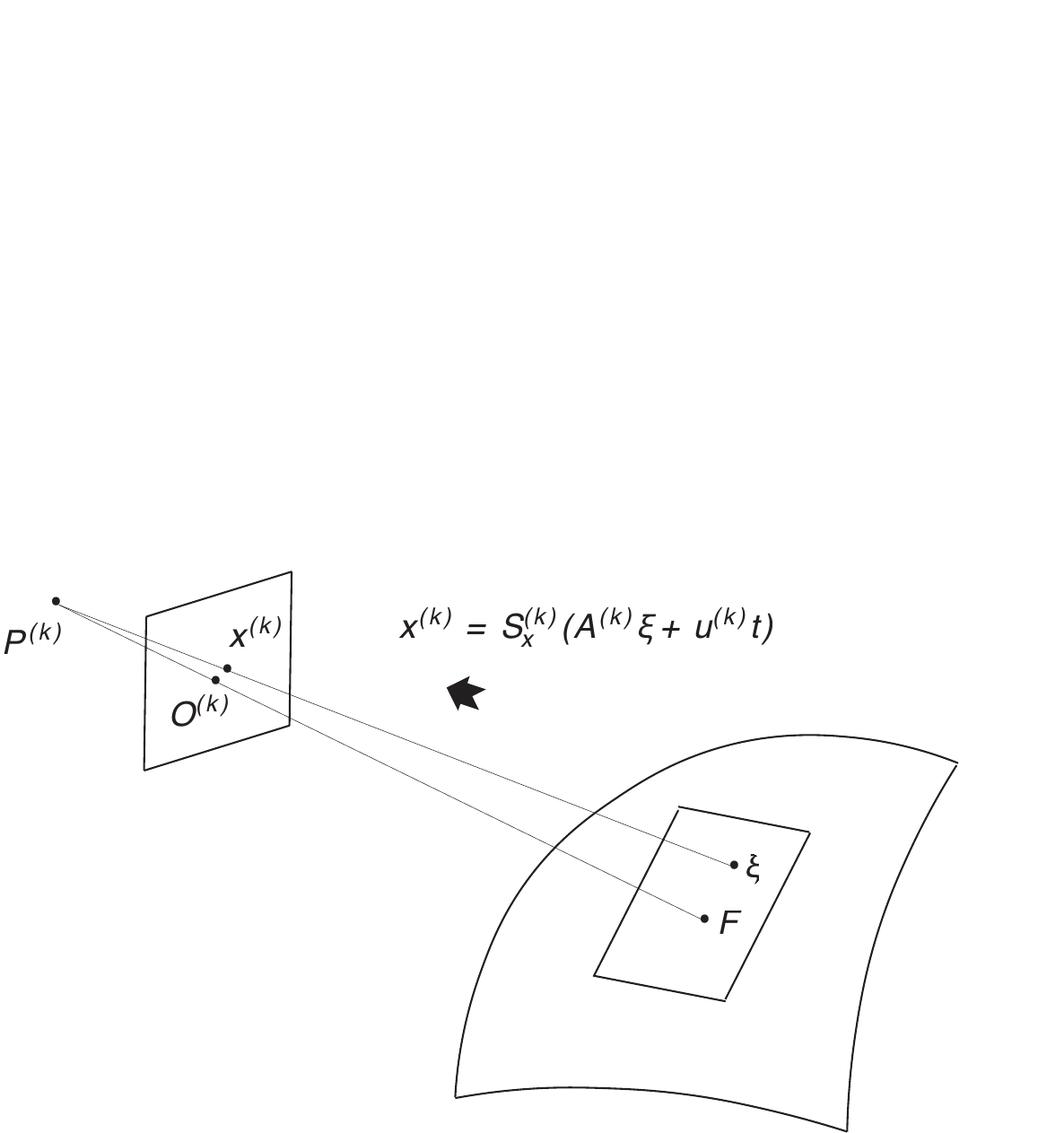} 
   \end{center}
   \caption{Illustration of the underlying geometric situtation for
     the locally linearized transformations from a local, possibly
     moving, surface patch to an arbitrary view indexed by $k$, with
     the fixation point $F$ on the surface mapped to the origin
     $O^{(k)} = 0$ in the image plane for the observer with the
     optic center $P^{(k)}$. Then, any point in the
     tangent plane to the surface at the fixation point, as
     parameterized by the local coordinates $\xi$ in a coordinate
     frame attached to the tangent plane of the surface with $\xi = 0$
     at the fixation point $F$, is by the
     local linearization mapped to the image point $x^{(k)}$.
     (Note, however, that this 3-D illustration is
     only intended to be schematic and not a fully quantitatively
     accurate representation,
     since the projection relations from the tangent
     plane to the surface have here been drawn according to a perspective
     projection model, whereas the algebraic model that we then will use for relating
     receptive field responses between the respective image domains are based
     on local linearizations of the underlying non-linear geometric
     transformations. This could in principle be accomplished by
     having different notation for the locally linearized projections
     {\em vs.\/} the true geometric projections. Here, we do, however,
     defer from making that distinction in the figure,
     in order to not overload the presentation with
     additional notation.)}
   \label{fig-singlegeom}
\end{figure}

\section{Geometric interpretation of the composed image transformation model}
\label{sec-geom-interpret}

In this section, we will interpret the above formulated
spatio-temporal covariance and transformation properties
geometrically, which will extend previous treatments of multi-view
geometry 
(see Hartley and Zisserman (\citeyear{HarZis04-Book})
and Faugeras (\citeyear{Faug-book}))
from (i)~multi-view observations of static scenes
for mainly point and line configurations to (ii)~multi-view
observations of dynamic scenes in terms of spatio-temporal receptive
field responses (with the extensions of this treatment to be performed in
Section~\ref{sec-cov-props-pairwise-views}),
where we have relative motions between the
objects and spatio-temporal events in the world and the observer.

With regard to a visual observer that observes 3-D objects in a dynamic
world, a geometric interpretation of the composed spatio-temporal image
transformation according to Equations~(\ref{eq-x-transf})
and~~(\ref{eq-t-transf}) can be obtained as follows:

\subsection{Transformations from the tangent plane of a local surface
  patch to the image domains of multiple views of the same scene or
  in similar scenes}
\label{sec-transf-from-tang-plane}

Consider a camera, alternatively an eye, that views a local surface
patch from different positions (optic centers)
  $P^{(k)} = (P_1^{(k)}, P_2^{(k)}, P_3^{(k)})^T \in \bbbr^3$ in the 3-D
world, relative to a global 3-D coordinate system.
For simplicity, with regard to
the following analysis that is to be performed, we will assume that the
fixation point is on the same point physical point
$F^{(k)} = (F_1^{(k)}, F_2^{(k)}, F_3^{(k)})^T \in \bbbr^3$
on the surface patch for each one of the observers, however, with the 3-D
coordinates for the fixation point now expressed relative to an individual
coordinate system for each observer (with index $k$), with the origin of
the individual 3-D coordinate system being at the optic center
$P^{(k)}$ of that observer.

For simplicity, we also assume that the image coordinates for each observer
are chosen such that the spatial image coordinate for the
fixation point being $x = (x_1, x_2)^T = (0, 0)^T$ at the time moment
$t = 0$, when the spatio-temporal receptive field response to the
studied is computed.

\subsubsection{Locally linearized static multi-view projection model}

Given the above multi-view viewing model, to first-order of approximation,
by approximating the non-linear perspective transformation for each
observer by its
first-order derivative, the transformation from a coordinate frame
with local coordinates $\xi = (\xi_1, \xi_2)^T$ in the tangent plane of the
surface patch, with the fixation point corresponding to
$\xi = (\xi_1, \xi_2)^T = (0, 0)^T$,
to the image coordinates $x^{(k)} = (x^{(k)}_1, x^{(k)}_2)^T$
in the image plane can be written on the form
\begin{equation}
  \label{eq-aff-transf-obs-model}
  x^{(k)} = A^{(k)} \, \xi,
\end{equation}
where $A^{(k)}$ represents a $2 \times 2$ affine transformation matrix
connected to the
viewing position $P^{(k)}$, and we can specifically choose a preferred
reference
view $P^{(0)}$ perpendicular to the tangent plane of the surface. We
can also decide to choose that preferred reference observation point, such that it
corresponds to orthonormal projection, with
$A^{(0)} = I$, where $I$ is the $2 \times 2$ unit matrix.

By introducing an additional explicit parameterization for the observation
points $P^{(k)}$, that are at different distances from the fixation point
on the local surface patch, and thus lead to different spatial scaling
factors in the underlying perspective transformation, that is locally
approximated by a local affine transformation, we can extend the model
(\ref{eq-aff-transf-obs-model}) to a model of the form
\begin{equation}
  \label{eq-sc-aff-transf-obs-model}
  x^{(k)} = S_x^{(k)} \, A^{(k)} \, \xi,
\end{equation}
where $S_x^{(k)}$ represents the additional scaling factor that arises by
changing the viewing distance relative to the observation point $P^{(0)}$
used as the main reference. For the scaled orthographic projection
model, the spatial scaling factor $S_x^{(k)}$ will thus correspond to
the inverse depth, such that $S_x^{(k)} = 1/Z^{(k)}$,
with the depth $Z^{(k)}$ for each observer measured relative to its
observation point $P^{(k)}$.

 \subsubsection{Locally linearized dynamic multi-view projection model}
 
If in addition, the local surface patch moves over time, with a 3-D
motion vector $U^{(k)} = (U_1^{(k)}, U_2^{(k)}, U_3^{(k)})^T$ relative
to the observation point $P^{(k)}$,
that is then mapped to the 2-D motion
vector $u^{(k)} = (u_1^{(k)}, u_2^{(k)})^T$ under the same orthonormal projection model,
as underlying the definition of the affine transformation matrix
$A^{(k)}$ above, then the motions of the spatio-temporal surface
patterns projected to the image plane can, with these scaled orthographic
projection models, to first-order of approximation, in the view labelled
by the index $k$, be modelled as a
motion over over time of the form
(see Figure~\ref{fig-singlegeom} for an illustration)
\begin{equation}
  \label{eq-sc-aff-vel-transf-obs-model}
  x^{(k)} = S_x^{(k)} (A^{(k)} \xi + u^{(k)} \, t),
\end{equation}
where
\begin{itemize}
\item
  $x^{(k)} \in \bbbr^2$ is the locally linearized projection of the physical point
  on the surface pattern, in the view from the observer with index $k$,
  at time $t$,
\item
  $\xi \in \bbbr^2$ constitute time-independent local coordinates in the tangent
  plane of the local surface patch,
\item
  $S_x^{(k)} \in \bbbr_+$ is a spatial scaling factor for the observer with index $k$,
\item
  $A^{(k)}$ is a non-singular $2 \times 2$ affine projection matrix, that represents
  an orthographic projection from the tangent plane of the surface
  to a plane parallel with the image plane,
  for the observer with index $k$,
\item
  $u^{(k)} \in \bbbr^2$ is a 2-D motion vector, that represents an orthographic
  projection of the 3-D motion vector $U^{(k)}$
  of the physical fixation point on the surface, to a plane parallel
  with the image plane, for the observer with index $k$.
\end{itemize}
  The role of the temporal scaling transformation according to
Equation~(\ref{eq-t-transf}) in this context
\begin{equation}
   \label{eq-t-transf-geom}
     t' = S_t \, t,
\end{equation}
is to additionally
account for making it possible to relate spatio-temporal events that occur either faster
or slower in relation to the spatio-temporal variations relative to
a reference view, for multiple observations at different time
moments of otherwise qualitatively similar types of motion patterns
or spatio-temporal events.

\begin{figure}[hbtp]
  \vspace{-25mm}
  \begin{center}
     \includegraphics[width=0.50\textwidth]{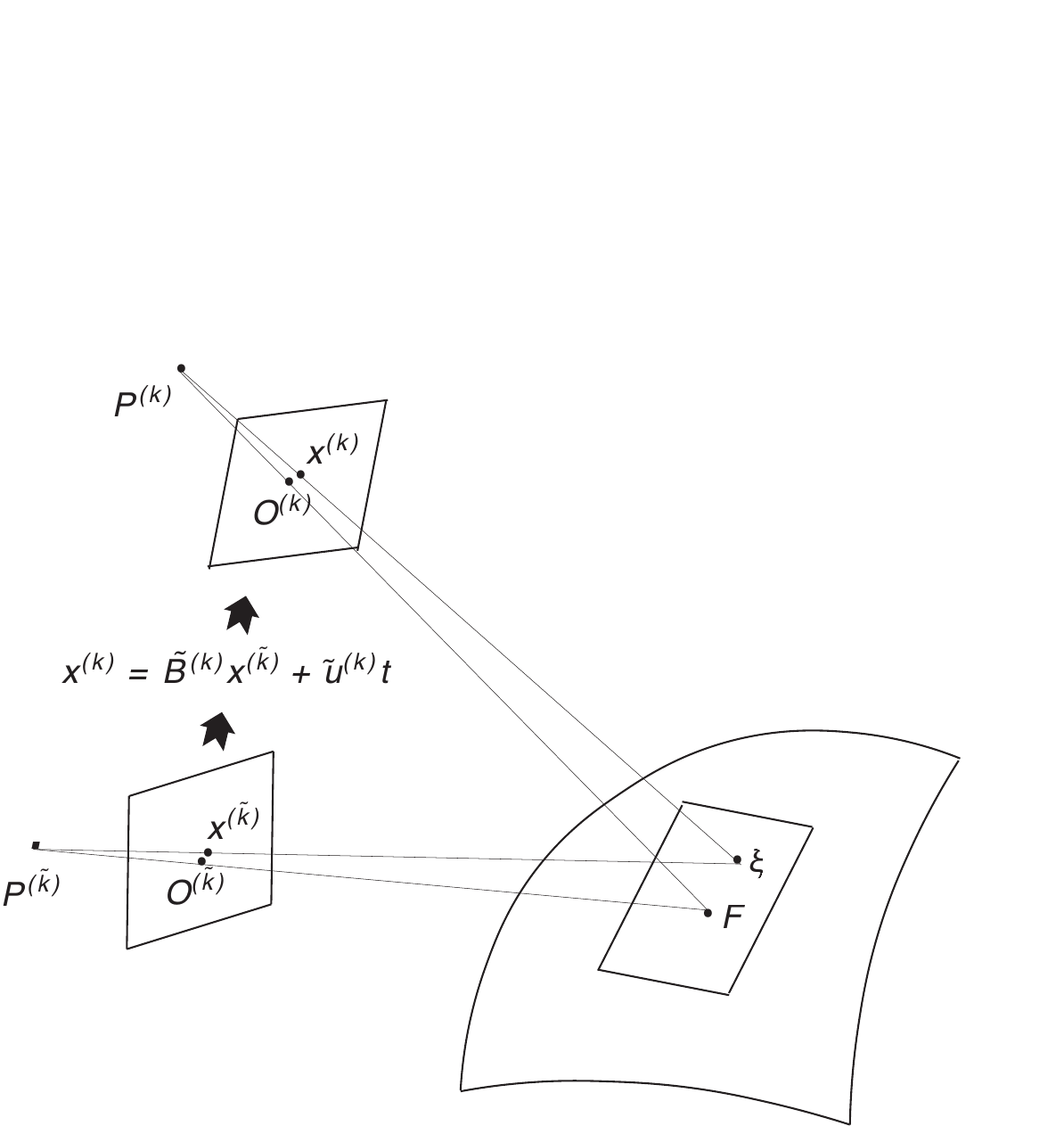} 
   \end{center}
   \caption{Illustration of the underlying geometric situtation for
     the locally linearized transformations between pairwise views of
     the same, possibly moving, local surface patch, with the view
     indexed by $\tilde{k}$ constituting the reference view and the
     view indexed by $k$ constituting an arbitrary view. Here, the
     fixation point $F$ on the surface is mapped to the origin
     $O^{(\tilde{k})} = 0$ in the reference view by the observer with the
     optic center $P^{(\tilde{k})}$, while the fixation point $F$ is
     mapped to the origin $O^{(k)} = 0$ in the other view by the observer with the
     optic center $P^{(k)}$. Then, in turn any other point in the
     tangent plane to the surface at the fixation point, as
     parameterized by the local coordinates $\xi$ in a coordinate
     frame attached to the tangent plane of the surface with $\xi = 0$
     at $F$, is by the
     local linearization mapped to the image point $x^{(\tilde{k})}$
     in the reference view indexed by $\tilde{k}$ and and by a corresponding other local
     linearization mapped mapped to the point $x^{(k)}$ in the other
     view indexed by $k$. (Note, however, that this 3-D illustration is
     only intended to be schematic and not a fully quantitatively
     accurate representation,
     since the projection relations from the tangent
     plane to the surface have here been drawn according to a perspective
     projection model, whereas the algebraic model that we then will use for relating
     receptive field responses between the respective image domains are based
     on local linearizations of the underlying non-linear geometric
     transformations. This could in principle be accomplished by
     having different notation for the locally linearized projections
     {\em vs.\/} the true geometric projections. Here, we do, however,
     defer from making that distinction in the figure,
     in order to not overload the presentation with
     additional notation.)}
   \label{fig-pairgeom}   
\end{figure}

\subsection{Transformations between different image domains of
  multiple views of the same local surface patch}
\label{sec-geom-interpret-mono}

While we above, for simplicity, decided to chose a normal view to the tangent
plane as the reference view, such an assumption is, however, not in
any way necessary, to be able to apply this joint spatio-temporal
covariance model for relating spatio-temporal receptive field
responses under composed spatio-temporal image transformations.

If we instead decide to choose some other particular observation
point $P^{(\tilde{k})}$, with its associated spatial scaling
factor $S_x^{(\tilde{k})}$, affine transformation matrix
$A^{(\tilde{k})}$ and image velocity $u^{(\tilde{k})}$ as the reference view,
then within the algebra of locally linearized approximations of the
underlying projective image transformation model between the
pairwise views, we obtain that 
Equation~(\ref{eq-sc-aff-vel-transf-obs-model}) will instead assume
the form (see Figure~\ref{fig-pairgeom} for an illustration)
\begin{equation}
  \label{eq-sc-aff-vel-transf-alt-obs-model}
  x^{(k)} = \tilde{B}^{(k)} \, x^{(\tilde{k})} + \tilde{u}^{(k)} \, t,
\end{equation}
where
\begin{itemize}
\item
  $x^{(k)} \in \bbbr^2$ is the locally linearized projection of the physical point
  on the surface pattern in the view from the observer with index $k$
  at time $t$,
\item
  $x^{(\tilde{k})} \in \bbbr^2$ is the locally linearized projection of the physical point
  on the surface pattern in the view from the observer with index
  $\tilde{k}$ at time $t$,
\item
  $\tilde{B}^{(k)}$ is a non-singular $2 \times 2$ affine projection matrix
  for the observer with index $k$ in relation to an observation from 
  a reference view with index $\tilde{k}$, and
\item
  $\tilde{u}^{(k)} \in \bbbr^2$ is a corresponding
  2-D relative motion vector for the observer with index $k$ in
  relation to an observation from a reference view with index $\tilde{k}$.
\end{itemize}
Here, we have thus, first of all, replaced the previous local coordinates $\xi$ in the tangent
plane of the surface patch by the image coordinates $x^{(k)}$
for a particular observation frame, to be used as the new reference.
Additionally, since the transformations from the new reference frame
will no longer correspond to interpretations according a scaled
orthographic model complemented by motion, we have removed the degree
of freedom for the separation of the linear approximation of the
perspective transformation in terms of a separate scaling factor and a
separate orthonormal projection, such that in the new frame, it should
instead hold that
\begin{align}
  \begin{split}
    \tilde{B}^{(\tilde{k})} = I,
  \end{split}\\
  \begin{split}
    \tilde{u}^{(\tilde{k})} = 0.
   \end{split}
\end{align}
Then, for $k = \tilde{k}$ the model
(\ref{eq-sc-aff-vel-transf-alt-obs-model}) reduces to the mere
identity
\begin{equation}
  x^{(\tilde{k})} = x^{(\tilde{k})}.
\end{equation}
By furthermore inserting Equation~(\ref{eq-sc-aff-vel-transf-obs-model}) for
$k = \tilde{k}$
\begin{equation}
  x^{(\tilde{k})} S_x^{(\tilde{k})} \, A^{(\tilde{k})} (\xi + u^{(\tilde{k})} \, t).
\end{equation}
into Equation~(\ref{eq-sc-aff-vel-transf-alt-obs-model}), we obtain
\begin{equation}
  x^{(k)} =
  \tilde{B}^{(k)} (S_x^{(\tilde{k})} \, A^{(\tilde{k})} \xi +
  u^{(\tilde{k})} \, t) + \tilde{u}^{(k)} \, t,
\end{equation}
which can be expanded to
\begin{equation}
  x^{(k)} =
  \tilde{B}^{(k)} \, S_x^{(\tilde{k})} \, A^{(\tilde{k})} \xi
  + (\tilde{B}^{(k)} \, S_x^{(\tilde{k})} \, A^{(\tilde{k})} \, u^{(\tilde{k})} + \tilde{u}^{(k)}) \, t.
\end{equation}
By identifying this expression with
Equation~(\ref{eq-sc-aff-vel-transf-alt-obs-model}), we get
\begin{align}
  \begin{split}
    S_x^{(k)} \, A^{(k)} = \tilde{B}^{(k)} \, S_x^{(\tilde{k})} \, A^{(\tilde{k})},
  \end{split}\\
  \begin{split}
    S_x^{(k)} \, A^{(k)} \, u^{(k)}
    = \tilde{B}^{(k)} \, S_x^{(\tilde{k})} \, A^{(\tilde{k})} \, u^{(\tilde{k})} + \tilde{u}^{(k)}.
  \end{split}
\end{align}
Thus, we have that the parameters $\tilde{B}^{(k)}$ and
$\tilde{u}^{(k)}$ of the spatio-temporal transformation model
(\ref{eq-sc-aff-vel-transf-alt-obs-model}) relative
to the particular reference view for $k = \tilde{k}$ are related to
the parameters $S_x^{(k)}$, $A^{(k)}$and $u^{(k)}$ for the
spatio-temporal transformation model
(\ref{eq-sc-aff-vel-transf-obs-model}) relative to the canonical frame
in the tangent plane of the surface patch according to
\begin{equation}
  \tilde{B}^{(k)}
  = \frac{S_x^{(k)}}{S_x^{(\tilde{k})}} \, A^{(k)} (A^{(\tilde{k})})^{-1}
\end{equation}
and
\begin{equation}
  \tilde{u}^{(k)} = S_x^{(k)} \, A^{(k)} \left( u^{(k)} - u^{(\tilde{k})} \right).
\end{equation}
In this way, we can derive the transformation parameters for the
transformation (\ref{eq-sc-aff-vel-transf-alt-obs-model}) between multiple
pairwise views, from the transformation parameters for the
monocular transformation (\ref{eq-sc-aff-vel-transf-obs-model})
from the tangent plane of the surface patch to any one of the
multiple single views.

\subsection{Transformation property of pairwise viewing parameters
  between different reference views}
\label{sec-geom-interpret-pair}

Let us next choose some other view for $k = \bar{k}$ as the
reference view, such the spatio-temporal image transformations between
the multiple pairwise views are instead of the form
\begin{equation}
  \label{eq-sc-aff-vel-transf-alt-obs-model-alt}
  x^{(k)} = \bar{B}^{(k)} \, x^{(\bar{k})} + \bar{u}^{(k)} \, t,
\end{equation}
where
\begin{itemize}
\item
  $x^{(k)} \in \bbbr^2$ is the locally linearized projection of the physical point
  on the surface pattern in the view from the observer with index $k$ at time $t$,
\item
  $x^{(\bar{k})} \in \bbbr^2$ is the locally linearized projection of the physical point
  on the surface pattern in the view from the observer with index
  $\bar{k}$ at time $t$,
\item
  $\bar{B}^{(k)}$ is a non-singular $2 \times 2$ affine projection matrix
  for the observer with index $k$ in relation to an observation from 
  a reference view with index $\bar{k}$ and
\item
  $\bar{u}^{(k)} \in \bbbr^2$ is a corresponding 2-D relative motion vector
  for the observer with index $k$ in relation to an observation from 
  a reference view with index $\bar{k}$.
\end{itemize}
To relate the parameters $\bar{B}^{(k)}$ and $\bar{u}^{(k)}$ in this
latter transformation model to the parameters $\tilde{B}^{(k)}$ and
$\tilde{u}^{(k)}$ in the previous transformation model
(\ref{eq-sc-aff-vel-transf-alt-obs-model}), let us insert
the expression for $x^{(\bar{k})}$ obtained by setting
$k = \bar{k}$ in (\ref{eq-sc-aff-vel-transf-alt-obs-model})
\begin{equation}
  \label{eq-sc-aff-vel-transf-alt-obs-model-for-kbar}
  x^{(\bar{k})} = \tilde{B}^{(\bar{k})} \, x^{(\tilde{k})} + \tilde{u}^{(\bar{k})} \, t,
\end{equation}
into (\ref{eq-sc-aff-vel-transf-alt-obs-model-alt}), which gives
\begin{equation}
  \label{eq-sc-aff-vel-transf-alt-obs-model-alt-inserted-xbar}
  x^{(k)} =
  \bar{B}^{(k)} (\tilde{B}^{(\bar{k})} \, x^{(\tilde{k})} + \tilde{u}^{(\bar{k})} \, t)
  + \bar{u}^{(k)} \, t,
\end{equation}
and
\begin{equation}
  \label{eq-sc-aff-vel-transf-alt-obs-model-alt-inserted-xbar-2}
  x^{(k)} =
  \bar{B}^{(k)} \, \tilde{B}^{(\bar{k})} \, x^{(\tilde{k})}
  + (\bar{B}^{(k)} \tilde{u}^{(\bar{k})}  + \bar{u}^{(k)}) \, t.
\end{equation}
Identifying the coefficients for $x^{(\tilde{k})}$ and $t$ with
the general expression (\ref{eq-sc-aff-vel-transf-alt-obs-model})
for the transformation between the views $\tilde{k}$ and $k$, then
gives that the transformation parameters $\bar{B}^{(k)}$ and
$\bar{u}^{(k)}$ for the corresponding transformation model based on
the view $\bar{k}$ have to be related to the parameters $\tilde{B}^{(k)}$ and
$\tilde{u}^{(k)}$ for the reference view based on $k = \tilde{k}$
according to
\begin{align}
  \begin{split}
    \tilde{B}^{(k)} = \bar{B}^{(k)} \, \tilde{B}^{(\bar{k})},
  \end{split}\\
  \begin{split}
    \tilde{u}^{(k)} = \bar{B}^{(k)} \, \tilde{u}^{(\bar{k})} + \bar{u}^{(k)}.
  \end{split}
\end{align}
Let us also insert the the expression for $x^{(\tilde{k})}$ obtained by setting
$k = \tilde{k}$ in (\ref{eq-sc-aff-vel-transf-alt-obs-model-alt})
\begin{equation}
  \label{eq-sc-aff-vel-transf-alt-obs-model-alt-for-ktilde}
  x^{(\tilde{k})} = \bar{B}^{(\tilde{k})} \, x^{(\bar{k})} + \bar{u}^{(\tilde{k})} \, t,
\end{equation}
into (\ref{eq-sc-aff-vel-transf-alt-obs-model}), which gives
\begin{equation}
  \label{eq-sc-aff-vel-transf-alt-obs-model-inserted-xtilde}
  x^{(k)}
  = \tilde{B}^{(k)} (\bar{B}^{(\tilde{k})} \, x^{(\bar{k})} + \bar{u}^{(\tilde{k})} \, t)
  + \tilde{u}^{(k)} \, t
\end{equation}
and
\begin{equation}
  \label{eq-sc-aff-vel-transf-alt-obs-model-inserted-xtilde-2}
  x^{(k)}
  = \tilde{B}^{(k)} \, \bar{B}^{(\tilde{k})} \, x^{(\bar{k})}
      + (\tilde{B}^{(k)} \, \bar{u}^{(\tilde{k})} + \tilde{u}^{(k)}) \, t.
\end{equation}
Identifying the coefficients for $x^{(\bar{k})}$ and $t$ with
the general expression (\ref{eq-sc-aff-vel-transf-alt-obs-model-alt})
for the transformation between the views $\bar{k}$ and $k$, then
gives that the transformation parameters $\tilde{B}^{(k)}$ and
$\tilde{u}^{(k)}$ for the corresponding transformation model based on
the view $\tilde{k}$ have to be related to the parameters $\bar{B}^{(k)}$ and
$\bar{u}^{(k)}$ for the reference view based on $k = \bar{k}$
according to
\begin{align}
  \begin{split}
    \bar{B}^{(k)} = \tilde{B}^{(k)} \, \bar{B}^{(\tilde{k})},
  \end{split}\\
   \begin{split}
    \bar{u}^{(k)} = \tilde{B}^{(k)} \, \bar{u}^{(\tilde{k})} + \tilde{u}^{(k)}.
  \end{split}
\end{align}
Furthermore, by setting the transformation
(\ref{eq-sc-aff-vel-transf-alt-obs-model-for-kbar}) between
$x^{(\tilde{k})} $ and $x^{(\bar{k})}$ into the transformation
(\ref{eq-sc-aff-vel-transf-alt-obs-model-alt-for-ktilde})
between $x^{(\bar{k})} $ and $x^{(\tilde{k})}$, we obtain
\begin{equation}
  x^{(\tilde{k})}
  = \bar{B}^{(\tilde{k})}
      (\tilde{B}^{(\bar{k})} \,  x^{(\bar{k})} + \tilde{u}^{(\bar{k})} \, t)
      + \bar{u}^{(\tilde{k})} \, t
\end{equation}
and
\begin{equation}
  x^{(\tilde{k})}
  = \bar{B}^{(\tilde{k})} \, \tilde{B}^{(\bar{k})} \,  x^{(\bar{k})}
     + (\bar{B}^{(\tilde{k})} \, \tilde{u}^{(\bar{k})} + \bar{u}^{(\tilde{k})}) \, t.
\end{equation}
Identifying the coefficients for $x^{(\bar{k})}$ and $t$, then gives
\begin{align}
  \begin{split}
     \bar{B}^{(\tilde{k})} \, \tilde{B}^{(\bar{k})} = I,
   \end{split}\\
 \begin{split}
     \bar{B}^{(\tilde{k})} \, \tilde{u}^{(\bar{k})} + \bar{u}^{(\tilde{k})} = 0,
   \end{split}
\end{align}
which can be rewritten into the following specific consistency
relations between the parameters in the mutually pairwise views based
on either of the reference views $\tilde{k}$ or $\bar{k}$
\begin{align}
  \begin{split}
     \bar{B}^{(\tilde{k})} = (\tilde{B}^{(\bar{k})})^{-1},
   \end{split}\\
 \begin{split}
     \bar{u}^{(\tilde{k})} = - \bar{B}^{(\tilde{k})} \, \tilde{u}^{(\bar{k})}.
   \end{split}
\end{align}
Due to the linearity of all the components of the first-order
approximations of these composed spatio-temporal image
transformations, the algebra for modelling the receptive field
responses is therefore closed under the considered family of 
spatio-temporal image transformations.

With regard to receptive field responses, this closedness property between
any set of locally linearized pairwise views of the same, possibly
moving, surface patch
will, in turn, imply that we can model the
spatio-temporal responses computed at matching points in space-time between
different pairwise views of the same local surface patch,
using a joint covariance property under a corresponding
class of composed spatio-temporal image transformations,
as will be developed more explicitly in the next section.


\begin{figure*}[hbtp]
  \[
    \begin{CD}
       \hspace{0mm}L(x, t;\; \tilde{\Sigma}, \tilde{\tau}, \tilde{v})
       @>{\footnotesize\begin{array}{c} x' = \tilde{B} \, x +
                         \tilde{u} \, t  \\ t' =
                         S_t \, t  \\ \tilde{\Sigma}' = \tilde{B} \,
                         \tilde{\Sigma} \, \tilde{B}^{T} \\ \tilde{\tau}' = S_t^2 \, \tilde{\tau} \\
                         \tilde{v}' = \frac{1}{S_t} (\tilde{B} \, \tilde{v} + \tilde{u}) \end{array}}>> L'(x', t';\; \tilde{\Sigma}', \tilde{\tau}', \tilde{v}') \\
       \Big\uparrow\vcenter{\rlap{$\scriptstyle{{*T(x, t;\; 
               \tilde{\Sigma}, \tilde{\tau}, \tilde{v})}}$}} & &
       \Big\uparrow\vcenter{\rlap{$\scriptstyle{{*T(x', t';\; 
               \tilde{\Sigma}', \tilde{\tau}', \tilde{v}')}}$}} \\
       f(x, t) @>{\footnotesize \begin{array}{c} x' = \tilde{B} \, x +
                                  \tilde{u} \, t
                                  \\ t' = S_t \, t \end{array}}>> f'(x', t')
    \end{CD}
  \]
\caption{Commutative diagram for the joint spatio-temporal smoothing
  component (\ref{eq-spat-temp-RF-model-mod})
  in the joint spatio-temporal receptive field
  model 
  under the composition of a spatial affine transformation, a
  Galilean transformation and a temporal scaling transformation
  according to (\ref{eq-transf-pairwise-views}) and
  (\ref{eq-t-transf-geom}), for relating the spatio-temporal receptive
  field responses between pairwise views of a local surface patch. This
  commutative diagram, which should be read from the lower left corner to the
  upper right corner, means that irrespective of whether the input video 
  sequence or video stream $f(x, t)$ is first subject to the composed transformation
  $x' = \tilde{B} \,  x + \tilde{u} \, t$ and $t' = S_t \, t$
  and then smoothed with a spatio-temporal kernel 
  $T(x', t';\; \tilde{\Sigma}', \tilde{\tau}', \tilde{v}')$, or instead directly convolved
  with the spatio-temporal smoothing kernel
  $T(x, t;\; \tilde{\Sigma}, \tilde{\tau}, \tilde{v})$ and then
  subject to the same joint spatio-temporal transformation, we do then get the same
  result, provided that the parameters of the spatio-temporal
  smoothing kernels are related
  according to $\tilde{\Sigma}' = \tilde{B} \, \tilde{\Sigma} \, \tilde{B}^{T}$,
  $\tilde{\tau}' = S_t^2 \, \tilde{\tau}$ and
  $\tilde{v}' = \frac{1}{S_t} (\tilde{B} \, \tilde{v} + \tilde{u})$.}
\label{fig-comm-diag-spat-temp-smooth-mod}
\end{figure*}

\section{Explicit joint spatio-temporal covariance properties between
  pairwise views of the same local surface patch}
\label{sec-cov-props-pairwise-views}

In this section, we will geometrically derive and analyze the
transformation properties of the spatio-temporal receptive field responses
that arise if we, instead of (i)~choosing a virtual normal view with its associated affine
transformation matrix $A$ in relation to the coordinates $\xi$ in the
tangent plane of the surface as the main
reference for expressing the composed geometric transformation,
according to the treatments in Sections~\ref{sec-composed-img-transf}
and~\ref{sec-transf-from-tang-plane}, (ii)~choosing
a particular observation view with its associated affine
transformation matrix $\tilde{B}$ in relation to the actual image
coordinates $x$ as the reference, for expressing the 
transformations properties between the spatio-temporal receptive field
responses computed from different views, according to the
treatment in Section~\ref{sec-geom-interpret-mono}.

In this way, we will obtain more explicit transformation properties
between the spatio-temporal receptive field responses computed
from different pairwise views of the same surface patch, compared to the
previous treatments of joint covariance properties in
Sections~\ref{eq-transf-prop-spat-temp-smooth}--\ref{sec-transf-prop-sc-norm-spat-temp-ders},
thus deriving the results of
integrating the results from the geometric analysis in Section~\ref{sec-geom-interpret}
into the transformation properties of the spatio-temporal receptive
fields according to Section~\ref{sec-joint-cov-props}.

In Section~\ref{sec-joint-cov-prop-spat-temp-smooth}, we derive such
explicit transformation properties for the underlying
spatio-temporal smoothing transformation.
In Section~\ref{sec-joint-cov-prop-spat-temp-spat-temp-ders}, we then
derive corresponding explicit transformation properties for regular (not
scale-normal\-ized) spatio-temporal derivatives, as well as
algebraically much simpler forms of transformation properties
in terms of scale-normalized derivatives. The latter results
clearly demonstrate
the advantage of using scale-normalized derivatives,
as described in Section~\ref{sec-sc-norm-spat-temp-ders}, as opposed to
regular spatio-temp\-oral derivatives, since the scale-normalized
spatio-temporal derivatives become essentially equal
(up to a possibly unknown rotation transformation over the elements
in the either vector-valued or matrix-valued image features in the case of affine-extended
scale-normalized derivatives) under the composed geometric image
transformation, provided that the parameters of the spatio-temporal receptive fields
can be properly matched in relation to the parameters of the composed geometric
image transformation.

\subsection{Prerequisites}

By comparing the joint transformation between pairwise views according
to (\ref{eq-sc-aff-vel-transf-alt-obs-model}), rewritten to the form
\begin{equation}
  \label{eq-transf-pairwise-views}
  x' = \tilde{B} \, x + \tilde{u} \, t,
\end{equation}
with the joint transformation property from the tangent plane of the
surface according to (\ref{eq-sc-aff-vel-transf-obs-model}),
rewritten to the form (\ref{eq-x-transf})
\begin{equation}
  \label{eq-transf-single-view}
  x' = S_x \, (A \,  x + u \, t),
\end{equation}
we can see that these transformations merely correspond to different
parameterizations of the same underlying algebraic structure, with the
parameters in the two different domains related according to
\begin{align}
  \begin{split}
    \tilde{B} = S_x \, A,
  \end{split}\\
 \begin{split}
    \tilde{u} = S_x \, u.
  \end{split}
\end{align}
Therefore, corresponding joint covariance properties for
spatio-temporal receptive fields can be stated for the
locally linearized transformations between pairwise views according to
(\ref{eq-transf-pairwise-views}) as for the locally linearized
transformation from the tangent plane of the surface to the image
plane according to (\ref{eq-transf-single-view}).

For clarity of presentation, we will in the following describe these
joint covariance properties for the spatio-temporal smoothing
operation and the spatio-temporal derivatives explicitly.
Since this involves removing the degree of freedom corresponding to
the parameter $S_x$ in the treatment in
Section~\ref{eq-transf-prop-spat-temp-smooth}, we will start by also
removing the degree of freedom corresponding to the spatial scale
parameter $s$ in the model (\ref{eq-spat-temp-RF-model}) for the
purely spatio-temporal smoothing operation of the spatio-temporal
receptive fields.

\begin{figure*}[hbtp]
  \[
    \begin{CD}
       \hspace{0mm}\nabla_x \partial_t L(x, t;\; \tilde{\Sigma}, \tilde{\tau}, \tilde{v})
       @>{\footnotesize\begin{array}{c} x' = \tilde{B} \, x +
                         \tilde{u} \, t  \\ t' =
                         S_t \, t  \\ \tilde{\Sigma}' = \tilde{B} \,
                         \tilde{\Sigma} \, \tilde{B}^{T} \\ \tilde{\tau}' = S_t^2 \, \tilde{\tau} \\
                         \tilde{v}' = \frac{1}{S_t} (\tilde{B} \, \tilde{v} + \tilde{u}) \\ \nabla_{x'} =
    \tilde{B}^{-T} \, \nabla_{x} \\ \partial_{t'} = -
                          u^T \tilde{B}^{-T} \, \nabla_x +
                         \frac{1}{S_t} \, \partial_t \end{array}}>> \nabla_{x'} \partial_{t'} L'(x', t';\;  \tilde{\Sigma}', \tilde{\tau}', \tilde{v}') \\
       \Big\uparrow\vcenter{\rlap{$\scriptstyle{{* (\nabla_x \partial_t T)(x, t;\; 
               \tilde{\Sigma}, \tilde{\tau}, \tilde{v})}}$}} & &
       \Big\uparrow\vcenter{\rlap{$\scriptstyle{{*( \nabla_{x'} \partial_{t'} T)(x', t';\; 
               \tilde{\Sigma}', \tilde{\tau}', \tilde{v}')}}$}} \\
       f(x, t) @>{\footnotesize \begin{array}{c} x' = \tilde{B} \, x +
                                  u \, t
                                  \\ t' = S_t \, t \end{array}}>> f'(x', t')
    \end{CD}
  \]
\caption{Commutative diagram for spatio-temporal derivative operators
  underlying the joint spatio-temporal receptive field
  model 
  under the composition of a spatial affine transformation, a
  Galilean transformation and a temporal scaling transformation
  according to (\ref{eq-transf-pairwise-views}) and
  (\ref{eq-t-transf-geom}), between different pairwise views of the same
  local surface patch. This
  commutative diagram, which should be read from the lower left corner to the
  upper right corner, means that irrespective of whether the input video 
  sequence or video stream $f(x, t)$ is first subject to the composed transformation
  $x' = \tilde{B} \,  x + \tilde{u} \, t$ and $t' = S_t \, t$
  and then filtered with a spatio-temporal derivative kernel
  $(\nabla_{x'} \partial_{t'} T)(x', t';\; \tilde{\Sigma}', \tilde{\tau}', \tilde{v}')$, or instead directly convolved
  with the spatio-temporal smoothing kernel
  $(\nabla_x \partial_t T)(x, t;\; \tilde{\Sigma}, \tilde{\tau}, \tilde{v})$ and then
  subject to the same joint spatio-temporal transformation, we do then get the same
  result, provided that the spatial and the temporal derivative
  operators are transformed according to
  $\nabla_{x'} = \tilde{B}^{-T} \, \nabla_{x}$ and
  $\partial_{t'} = - u^T \tilde{B}^{-T} \, \nabla_x + \frac{1}{S_t} \, \partial_t$
  and that the parameters of the spatio-temporal
  smoothing kernels are related
  according to $\tilde{\Sigma}' = \tilde{B} \, \tilde{\Sigma} \, \tilde{B}^{T}$,
  $\tau' = S_t^2 \, \tau$ and $\tilde{v}' = \frac{1}{S_t} (\tilde{B} \, \tilde{v} + \tilde{u})$.
  (In this commutative diagram, we have illustrated the general
  covariance properties of spatio-temporal derivatives in terms of the
  composed spatio-temporal derivative $\nabla_x \partial_t T$. Similar
covariance properties can, of course, also be obtained for
other combinations of the spatial and the temporal derivative
operators $\nabla_x$ and $\partial_t$.)}
\label{fig-comm-diag-spat-temp-ders-mod}
\end{figure*}

\begin{figure*}[hbtp]
  \[
    \begin{CD}
       \hspace{0mm}\nabla_{x,\affnormtiny} \partial_{{\bar t},\normtiny} L(x, t;\; \tilde{\Sigma}, \tilde{\tau}, \tilde{v})
       @>{\footnotesize\begin{array}{c} x' = \tilde{B} \, x + \tilde{u} \, t  \\ t' =
                         S_t \, t  \\ \tilde{\Sigma}' = \tilde{B} \,
                         \Sigma \, \tilde{B}^{T} \\ \tilde{\tau}' = S_t^2 \, \tilde{\tau} \\
                         v' = \frac{1}{S_t} (\tilde{B} \, \tilde{v} +
                         \tilde{u}) \\
                         \nabla_{x,\affnormtiny} = \tilde{\Sigma}^{1/2} \, \nabla_x \\
                         \nabla_{x',\affnormtiny} = \tilde{\Sigma'}^{1/2} \, \nabla_{x'} \\    \nabla_{x',\affnormtiny} =
    \tilde{\rho} \, \nabla_{x,\affnormtiny} \\ \partial_{{\bar t}',\normtiny} = 
                         \partial_{{\bar t},\normtiny} \end{array}}>>
                     \nabla_{x',\affnormtiny} \partial_{{\bar t}',\normtiny} L'(x', t';\; \tilde{\Sigma}', \tilde{\tau}', \tilde{v'}) \\
       \Big\uparrow\vcenter{\rlap{$\scriptstyle{{*
               (\nabla_{x,\affnormtiny} \partial_{{\bar t},\normtiny} T)(x, t;\; 
               \tilde{\Sigma}, \tilde{\tau}, \tilde{v})}}$}} & &
       \Big\uparrow\vcenter{\rlap{$\scriptstyle{{*(
               \nabla_{x',\affnormtiny} \partial_{{\bar t}',\normtiny} T)(x', t';\; 
               \tilde{\Sigma}', \tilde{\tau}', \tilde{v}')}}$}} \\
       f(x, t) @>{\footnotesize \begin{array}{c} x' = \tilde{B} \, x + \tilde{u} \, t
                                  \\ t' = S_t \, t \end{array}}>> f'(x', t')
    \end{CD}
  \]
\caption{Commutative diagram for scale-normalized spatio-temporal derivative operators
  defined from the joint spatio-temporal receptive field
  model (\ref{eq-spat-temp-RF-model-der}) under the composition of a spatial affine transformation, a
  Galilean transformation and a temporal scaling transformation
  according to (\ref{eq-transf-pairwise-views}) and
  (\ref{eq-t-transf-geom}), between different pairwise views of the same
  local surface patch. This
  commutative diagram, which should be read from the lower left corner to the
  upper right corner, means that irrespective of whether the input video 
  sequence or video stream $f(x, t)$ is first subject to the composed transformation
  $x' = \tilde{B} \,  x + \tilde{u} \, t$ and $t' = S_t \, t$
  and then filtered with a scale-normalized spatio-temporal derivative kernel
  $(\nabla_{x',\affnormtiny} \partial_{t',\normtiny} T)(x', t';\; \tilde{\Sigma}', \tilde{\tau}', \tilde{v}')$,
  or instead directly convolved
  with the scale-normalized spatio-temporal smoothing kernel
  $(\nabla_{x,\affnormtiny} \partial_{t,\normtiny} T)(x, t;\; \tilde{\Sigma}, \tilde{\tau}, \tilde{v})$ and then
  subject to the same joint spatio-temporal transformation, we do
  then, up to a possibly unknown rotation transformation, get the same
  result, provided that the parameters of the spatio-temporal
  smoothing kernels are related
  according to $\tilde{\Sigma}' = \tilde{B} \, \tilde{\Sigma} \, \tilde{B}^{T}$,
  $\tau' = S_t^2 \, \tau$ and $\tilde{v}' = \frac{1}{S_t} (\tilde{B} \, \tilde{v} + \tilde{u})$.
  Note, in particular, the conceptual simplification in relation to
  the corresponding commutative diagram based on regular partial
  derivatives that have not been subject to scale normalization or
  velocity adaptation regarding the temporal derivatives, in that the
  scale-normalized spatio-temporal derivatives in this commutative
  diagram are essentially equal, up to a possibly unknown rotation transformation.
  (In this commutative diagram, we have illustrated the general
  covariance properties of spatio-temporal derivatives for the
  particular choice of the
  composed spatio-temporal derivative operator
  $\nabla_{x,\affnorm} \partial_{t,\norm} T$
  in the spatio-temporal receptive field model
  (\ref{eq-spat-temp-RF-model-mod}). Similar
covariance properties can, of course, also be obtained for
other combinations of the spatial and the temporal derivative
operators $\nabla_{x,\affnorm}$ and $\partial_{t,\norm}$
for which corresponding covariance properties hold.)}
\label{fig-comm-diag-spat-temp-ders-sc-norm-pairwise-views}
\end{figure*}

\subsection{Joint covariance property for the purely spatio-temporal
  smoothing component of the spatio-temporal receptive fields}
\label{sec-joint-cov-prop-spat-temp-smooth}

If we merge the degrees of freedom in the spatial scale parameter $s \in \bbbr_+$
and the spatial covariance matrix $\Sigma \in \bbbs_+^2$ in the purely
spatio-temporal smoothing component of the receptive fields according
to (\ref{eq-spat-temp-RF-model}) into the joint parameter
\begin{equation}
  \tilde{\Sigma} = s^2 \Sigma,
\end{equation}
then we can express the purely spatio-temporal smoothing component of
the receptive fields according to
\begin{equation}
  \label{eq-spat-temp-RF-model-mod}
  T(x, t;\; \tilde{\Sigma}, \tilde{\tau}, \tilde{v}) 
  = \tilde{g}(x - \tilde{v} \, t;\; \tilde{\Sigma}) \, h(t;\; \tilde{\tau}),
\end{equation}
where we also redefine the 2-D affine Gaussian kernel
(\ref{eq-gauss-fcn-2D}) according to
 \begin{equation}
   \label{eq-gauss-fcn-2D-mod}
   \tilde{g}(x;\; \tilde{\Sigma})
     = \frac{1}{2 \pi \sqrt{\det \tilde{\Sigma}}} \, e^{-x^T  \tilde{\Sigma}^{-1} x/2}.
\end{equation}
If we correspondingly to (\ref{eq-f-fprim-transf-proof}) consider two
video sequences or video streams $f'(x', t')$ and $f(x, t)$, that are related according
to (\ref{eq-transf-pairwise-views}) and (\ref{eq-t-transf-geom}) such that
\begin{equation}
  f'(x', t') = f(x, t),
\end{equation}
and correspondingly to (\ref{eq-scsp-repr-L-in-proof}) and
(\ref{eq-scsp-repr-Lprim-in-proof}) define spatio-temporal scale-space
representations of these video sequences or video streams according to
\begin{align}
  \begin{split}
     & L(x, t;\; \tilde{\Sigma}, \tilde{\tau}, \tilde{v}) = 
  \end{split}\nonumber\\
  \begin{split}
    \label{eq-scsp-repr-L-in-proof-mod}
     & = \int_{\xi \in \bbbr^2} \int_{\eta \in \bbbr}
               T(\xi, \eta;\; \tilde{\Sigma}, \tilde{\tau}, \tilde{v}) \, f(x - \xi, t - \eta) \,
               d\xi \, d\eta,
  \end{split}\\
  \begin{split}
     & L'(x', t';\; \tilde{\Sigma}', \tilde{\tau}', \tilde{v}') = \\
  \end{split}\nonumber\\
  \begin{split}
      & = \int_{\xi' \in \bbbr^2} \int_{\eta' \in \bbbr}
                 T(\xi', \eta';\; \tilde{\Sigma}', \tilde{\tau}', \tilde{v}') \times
 \end{split}\nonumber\\
  \begin{split}
    \label{eq-scsp-repr-Lprim-in-proof-mod}    
    & \phantom{= \int_{\xi' \in \bbbr^2} \int_{\eta' \in \bbbr}} \quad
                 f'(x' - \xi', t' - \eta') \,
                d\xi' \, d\eta',
  \end{split}
\end{align}
it then follows, from similar calculations as lead to the
transformation properties
(\ref{eq-joint-cov-prop-result-of-proof})--(\ref{eq-v-transf-result}),
that the spatio-temporal scale-space representations
$L(x, t;\; \tilde{\Sigma}, \tilde{\tau}, \tilde{v})$ and
$L'(x', t';\; \tilde{\Sigma}', \tilde{\tau}', \tilde{v}')$ of the video
sequences or video streams $f(x, t)$ and $f'(x', t')$ are related according to
\begin{equation}
  \label{eq-joint-cov-prop-result-of-proof-mod}
  L'(x', t';\; \tilde{\Sigma}', \tilde{\tau}', \tilde{v}')
  = L(x, t;\; \tilde{\Sigma}, \tilde{\tau}, \tilde{v}),
\end{equation}
provided that the parameters of the receptive fields transform according to
\begin{align}
  \begin{split}
    \label{eq-Sigma-transf-result-mod}
    \tilde{\Sigma}' & = \tilde{B} \, \tilde{\Sigma} \, \tilde{B}^{T},
  \end{split}\\
  \begin{split}
    \label{eq-tau-transf-result-mod}
    \tilde{\tau}' & = S_t^2 \, \tilde{\tau},
  \end{split}\\
  \begin{split}
    \label{eq-v-transf-result-mod}    
    \tilde{v}' & = \frac{1}{S_t} (\tilde{B} \, \tilde{v} + \tilde{u}).
  \end{split}
\end{align}
This follows from similar calculations as done in
Section~\ref{eq-transf-prop-spat-temp-smooth}, by replacing the
previous affine transformation matrix $A$ by the new affine
transformation matrix $B$, while simultaneously replacing the spatial
scaling factor $S_x$ by $1$;
see Figure~\ref{fig-comm-diag-spat-temp-smooth-mod}
for an illustration in terms of a commutative diagram.

\subsection{Joint covariance properties for the spatial and the temporal derivative operator components of the spatio-temporal receptive fields}
\label{sec-joint-cov-prop-spat-temp-spat-temp-ders}

By similarly replacing the affine transformation matrix $A$ by the
affine transformation matrix $\tilde{B}$, while simultaneously
replacing the spatial scaling factor $S_x$ by $1$ in the
transformation properties
(\ref{eq-joint-transf-prop-spat-grad})--(\ref{eq-joint-transf-prop-temp-der})
and
(\ref{eq-joint-transf-prop-spat-grad-inv})--(\ref{eq-joint-transf-prop-temp-der-inv})
of the spatial and temporal derivative operators under the composed
spatio-temporal transformation defined by (\ref{eq-x-transf}) and (\ref{eq-t-transf}),
we obtain that the spatial and the
temporal derivative operators in the two domains under the composed
spatio-temporal transformation defined by
(\ref{eq-transf-pairwise-views}) and (\ref{eq-t-transf-geom}) are related according to
\begin{align}
  \begin{split}
    \label{eq-joint-transf-prop-spat-grad-mod}
    \nabla_{x} = \tilde{B}^T \, \nabla_{x'},
  \end{split}\\
  \begin{split}
    \label{eq-joint-transf-prop-temp-der-mod}  
    \partial_{t} = u^T \, \nabla_{x'}  + S_t \, \partial_{t'},
  \end{split} 
\end{align}
and
\begin{align}
  \begin{split}
    \label{eq-joint-transf-prop-spat-grad-inv-mod}
    \nabla_{x'} = \tilde{B}^{-T} \, \nabla_{x},
  \end{split}\\
  \begin{split}
    \label{eq-joint-transf-prop-temp-der-inv-mod}      
    \partial_{t'} = - \, u^T \tilde{B}^{-T} \, \nabla_x + \frac{1}{S_t} \, \partial_t,
  \end{split} 
\end{align}
see Figure~\ref{fig-comm-diag-spat-temp-ders-mod} for a commutative
diagram that illustrates these joint covariance properties.

In analogy with the previous treatment of the transformation properties of
scale-normalized derivatives in
Section~\ref{sec-transf-prop-sc-norm-spat-temp-ders}, also these
transformation properties will be simplified, if instead expressing them in
terms of scale-normalized derivatives, and also if replacing the
partial temporal derivative operators by velocity-adapted temporal
derivatives:
\begin{itemize}
\item
  If we consider the group of general%
\footnote{For the purpose of modelling the transformations between
  pairwise views, we do here not explicitly consider the special case
  of the similarity group, since the geometric viewing
  configurations that lead to such a restricted form of variability are
  very degenerate, in relation to the case of multiple views of the
  same object from different viewing directions.}
  affine transformation matrices $\tilde{B}$, and
  define the scale-normalized affine gradient vector
  and the scale-normalized affine Hessian matrix
  according to (\ref{eq-def-sc-norm-aff-grad-op})
  and (\ref{eq-def-sc-norm-aff-hess-mat}), with the spatial scale
  parameter set to $s = 1$ and the spatial covariance matrix $\Sigma$
  replaced by $\tilde{\Sigma}$, 
  then, based on the results in Section~\ref{sec-cov-prop-sc-norm-aff-grad-op} and
  Section~\ref{sec-cov-prop-sc-norm-aff-hess-mat},
  these scale-normalized affine
  derivative-based entities will be equal up to 
  rotation matrices $\tilde{\rho}$ according to
  \begin{multline}
    (\nabla_{x',\affnorm}  L')(x', t';\; \tilde{\Sigma}', \tilde{\tau}', \tilde{v}') = \\ =
    \tilde{\rho} \,  (\nabla_{x,\affnorm} L)(x, t;\; \tilde{\Sigma}, \tilde{\tau}, \tilde{v})
  \end{multline}
  and
  \begin{multline}
    ({\cal H}_{x',\affnorm} L')(x', t';\; \tilde{\Sigma}', \tilde{\tau}', \tilde{v}') = \\
    = \tilde{\rho} \, ({\cal H}_{x,\affnorm} L)(x, t;\; s, \tilde{\Sigma}, \tilde{\tau}, \tilde{v}) \, \tilde{\rho}^T,
  \end{multline}
  provided that the parameters $\tilde{\Sigma}$, $\tilde{\Sigma}'$,
  $\tilde{\tau}$, $\tilde{\tau}'$, $\tilde{v}$ and $\tilde{v}'$ of the
  receptive fields are matched according to
  Equations~(\ref{eq-Sigma-transf-result-mod})--(\ref{eq-v-transf-result-mod}).
\item
  The velocity-adapted spatio-temporal
  derivative operators according to
  (\ref{eq-def-vel-adapt-ders-both-domains-repeated}),
  with $v$ replaced by $\tilde{v}$, and
  extended to scale-normalized derivatives
  with $\tau$ replaced by $\tilde{\tau}$, will, based on the result underlying
  Equation~(\ref{eq-equal-veladapt-ders-composed-transf-main-result}),
  be equal
  \begin{multline}
    \label{eq-equal-veladapt-ders-composed-transf-main-result-sc-norm-again}
    \partial_{{\bar t}',\norm}L' (x', t';\; \tilde{\Sigma}', \tilde{\tau}', \tilde{v}')  = \\
    =  \partial_{{\bar t},\norm}L(x, t;\; \tilde{\Sigma}, \tilde{\tau}, \tilde{v}),
  \end{multline}
  provided that the parameters $\tilde{\Sigma}$, $\tilde{\Sigma}'$,
  $\tilde{\tau}$, $\tilde{\tau}'$, $\tilde{v}$ and $\tilde{v}'$ of the
  receptive fields are matched according to
  Equations~(\ref{eq-Sigma-transf-result-mod})--(\ref{eq-v-transf-result-mod}).
\end{itemize}
Figure~\ref{fig-comm-diag-spat-temp-ders-sc-norm-pairwise-views}
illustrates the combined effects of these covariance
properties in a joint commutative diagram.

\subsection{Transformations of receptive field responses under varying geometric viewing conditions}

In this way, if we consider a vision system, either biological or
based on computer vision operations, that records spatial and
spatio-temporal image structures observed by viewing local surface
patches, in either a static or dynamic world, in terms of receptive
field responses, then the above geometric analysis in combination with
the the previously derived joint transformation properties according to
Equations~(\ref{eq-joint-cov-prop-result-of-proof})--(\ref{eq-v-transf-result})
of the underlying spatial or spatio-temporal smoothing operations in the
either spatial or spatio-temporal receptive fields, together with the
corresponding explicit transformation properties of the spatial and
temporal derivative operators according to
Equations~(\ref{eq-joint-transf-prop-spat-grad})--(\ref{eq-joint-transf-prop-temp-der-inv})
do therefore, beyond a trivial
usually unknown spatial translation between the origins of the
coordinate systems between the different image domains,
fully describe how the spatial and
spatio-temporal receptive field responses can be related or matched, when viewing
either the same physical scene from multiple views.

When complemented by
temporal scaling transformations, this matching property does furthermore
extend to relating or matching the receptive field responses between
different views of similarly looking motion patterns or
spatio-temporal events that may occur
either faster or slower between different instances of the same event.

In this context it should be remarked, however, that due to the
modelling of the spatial or spatio-temporal image transformations in
terms of local linearizations only, the matches between the receptive
field responses obtained according to the joint covariance property
will not be fully perfect, in situations when the spatial or spatio-temporal
support regions of the receptive field cover larger regions in image
space or space-time than cannot be compactly modelled by local
linearizations. Compared to not attempting to compensate for the
effect of the spatial or spatio-temporal image transformations on the
receptive field responses, the positive effects of incorporating
covariance properties of the receptive field responses with respect to
local linearizations of the underlying non-linear perspective or
projective image transformations should, however, be expected
to lead to substantial improvements. Handling the locally linear approximations
of the underlying non-linear perspective or projective image
transformations can in this context also be expected to be be
conceptually much simpler, than aiming at
compensating for more complex non-linear image deformation models.

With regard to observations of more complex scenes, containing
multiple local image structures, based on different characteristics in
terms of {\em e.g.\/} local surface geometry, it should be noted that
linearized transformations of receptive field responses could also be
computed regionally, over larger regions of image space than could be
well modelled by a single locally linearized image transformation.
Then, if regional statistics of receptive field responses are to be
computed, for {\em e.g.\/} purposes with regard to spatial or
spatio-temporal recognition, then an overall compensation of the
receptive field responses with respect to gross geometric and motion
effects of the entire region could also be performed, thus with the
parameters of the spatio-temporal image transformation not determined
by the local spatio-temporal geometry and motion, but by instead
determined by a coarser-scale regional geometry and motion.

\section{Interpretation in terms of the variability of image
  structures under natural image transformations in relation to the
  degrees of freedom spanned by the parameters in the spatio-temporal receptive field model}
\label{sec-interpret-geom-biol}

With regard to the axiomatically%
\footnote{For the axiomatically formulated theory of visual receptive
  fields, that leads to the principled model for spatio-temporal receptive
  fields that underlies this treatment, see Lindeberg
  (\citeyear{Lin10-JMIV}) concerning the foundations and
  Lindeberg (\citeyear{Lin21-Heliyon}) Appendix~B for a complement.}
derived model for
spatio-temporal receptive fields (\ref{eq-spat-temp-RF-model}), that we
build the analysis in this treatment on, the geometric analysis that we have presented
in Section~\ref{sec-geom-interpret}
shows that the degrees of freedom in this spatio-temporal receptive
field model
(the parameters $s$, $\Sigma$, $\tau$ and $v$
in (\ref{eq-spat-temp-RF-model})) span the degrees of
freedom in the locally linearized scaled orthographic model,
complemented with a Galilean motion to account for relative motions
between objects in the world and the observer, as well as a temporal scaling
transformation to account for spatio-temporal events that may occur
either faster or slower relative to a reference view
(the parameters $S_x$, $A$, $S_t$ and $u$ in
(\ref{eq-sc-aff-vel-transf-obs-model}) and (\ref{eq-t-transf-geom})).

The degrees of freedom in the slightly modified spatio-temporal
receptive field model (\ref{eq-spat-temp-RF-model-mod})
(the parameters $\tilde{\Sigma}$, $\tilde{\tau}$ and $\tilde{v}$) do also span the degrees of
freedom in the locally linearized projective projection model
between pairwise views according to (\ref{eq-transf-pairwise-views})
and (\ref{eq-t-transf-geom})
(the parameters $\tilde{B}$, $S_t$ and $\tilde{u}$).

In this respect, these spatio-temporal receptive field models
make it possible to perfectly capture
the first-order linearized approximations of the variabilities
generated by observing the surfaces of smooth objects in the world,
that move in relation to the observer in a dynamic 3-D environment.
This does specifically imply that with regard to modelling the
first-order linearizations of receptive field responses under
the perspective or projective transformations in either single-view or
multi-view observations of 3-D scenes, we can isomorphically perform
these operations as joint spatio-temporal image transformations in
image space only. Thus, the algebra of the interaction between the
receptive fields and the first-order linearized geometric
transformations constitute a sufficient%
\footnote{Note, however, that complementary mechanisms may be needed
  to handle discontinuities in depth or surface
  orientation, as well as for handling the effects of illumination variations. With
regard to a subset of the space of variability spanned by illumination
variations, it should, however, be noted that if the studied idealized
receptive field model is applied over a logarithmic brightness scale, then the
receptive field responses will be automatically invariant under local
multiplicative illumination variations and exposure mechanisms,
see Lindeberg (\citeyear{Lin13-BICY}) Section~2.3 and
Lindeberg (\citeyear{Lin21-Heliyon}) Section~3.4.}
algebra to handle either
single-view or multi-view observations of smooth surface patches in a
dynamic world.

In this way, it is not really necessary to make use of explicit models
of 3-D scene geometry or 3-D object motion, when to operate on the
spatio-temporal image data that originate from different views.
Instead, it is sufficient to just make use of the composed
spatio-temporal image transformations between the multiple views of
the same scene, which in that way constitute a minimal type of model
of the world with respect to the image-based observer's view.

As we previously described in Section~\ref{sec-rel-biol-vision},
comparisons with biological receptive fields obtained by
neurophysiological recordings of neurons in the primary visual cortex
(V1), have shown that the receptive fields of simple cells can be
qualitatively rather well modelled by idealized receptive fields of
the form,
see Lindeberg (\citeyear{Lin21-Heliyon}) Section~4 for explicit
comparisons between biological receptive fields and these idealized
receptive fields of the form
\begin{multline}
  \label{eq-spat-temp-RF-model-der-again-again}
    T_{{\varphi}^{m} {\bar t}^n}(x, t;\; s, \Sigma, \tau, v) 
    =  \\ =
        \partial_{\varphi}^{m} \, \partial_{\bar t}^n 
           \left( g(x - v \, t;\; s, \Sigma) \, h(t;\; \tau) \right).
 \end{multline}
By extending this definition with  the
affine scale-normalized directional derivative
operator $\partial_{\varphi,\norm}^{m}$ according to
(\ref{eq-dir-der-def-sc-norm-basic}), again in one of the
eigendirections $\varphi$ of the spatial covariance matrix $\Sigma$,
as well as complementing with scale-normalized
velocity-adapted temporal derivatives
$\partial_{{\bar t},\norm}^n$  in the direction $v$
according to (\ref{eq-temp-der-def-sc-norm}), we can thus
also express a corresponding scale-normalized model of the
spatio-temporal receptive fields according to (\ref{eq-spat-temp-RF-model-der-again}) as
\begin{multline}
  \label{eq-spat-temp-RF-model-der-norm}
    T_{{\varphi}^{m} {\bar t}^n,\norm}(x, t;\; s, \Sigma, \tau, v) 
    =  \\ =
        \partial_{\varphi,\norm}^{m} \, 
        \partial_{{\bar t},\norm}^n 
           \left( g(x - v \, t;\; s, \Sigma) \, h(t;\; \tau) \right),
\end{multline}
which then extends the applicability of the previous model
(\ref{eq-spat-temp-RF-model-der-again})  to
provable covariance properties under compositions
of spatial similarity transformations

If we additionally, would extend the interpretation of those modelling
results, corresponding to spatial derivatives of orders 1 and 2, to
replacing the interpretation of the spatial derivative operators as
components of the scale-normalized affine gradient vector according to
(\ref{eq-def-sc-norm-aff-grad-op}) or
as components of the scale-normalized affine Hessian matrix according
to (\ref{eq-def-sc-norm-aff-hess-mat})
\begin{align}
  \begin{split}
      & T_{\nabla_x {\bar t}^n,\norm}(x, t;\; s, \Sigma, \tau, v) 
      =  \\ 
        & \quad\quad = \nabla_{x,\affnorm} \, 
        \partial_{{\bar t},\norm}^n 
           \left( g(x - v \, t;\; s, \Sigma) \, h(t;\; \tau) \right),
  \end{split}\\
 \begin{split}
      & T_{{\cal H}_x {\bar t}^n,\norm}(x, t;\; s, \Sigma, \tau, v) 
      =  \\ 
       & \quad\quad = {\cal H}_{x,\affnorm} \, 
        \partial_{{\bar t},\norm}^n 
           \left( g(x - v \, t;\; s, \Sigma) \, h(t;\; \tau) \right),
  \end{split}
\end{align}
then such a model would additionally allow for provable covariance
properties under arbitrary combinations of spatial affine
transformations and Galilean transformations, with clear biological
relevance for a biological visual agent, to be able to handle the variability of
image structures under natural image transformations.

Furthermore, considering that the receptive fields of simple cells in
the primary visual cortex can be qualitatively very well modelled by such
spatio-temporal receptive fields,
these results
can be taken as further support for the working hypothesis that the receptive
fields in the primary visual cortex may be regarded as being very well adapted to the structure
of our environment, as also previously proposed in connection with the formulation of
the normative theory of visual receptive fields that underlies the
definition of the idealized spatio-temporal receptive field model that
we have used as a basis for this theoretical
treatment, see Lindeberg (\citeyear{Lin21-Heliyon}) Section~6, for a
condensed summary of such conceptual theoretical arguments
and Lindeberg
(\citeyear{Lin23-FrontCompNeuroSci,Lin25-JCompNeurSci-spanelong})
for formulations of
more explicit hypotheses regarding possible affine covariance and
Galilean covariance for the receptive fields in biological vision.

\section{Using 2-D image deformation parameters from the matching of receptive field responses for obtaining direct cues to the 3-D
  structure of the environment}
\label{sec-cues-3d-structure}

With respect to the inference of cues to the 3-D structure of the
world, it should furthermore be noted that:
\begin{itemize}
\item
  knowledge about the affine
  transformation $A^{(k)}$, in the locally linearized perspective projection
  model (\ref{eq-sc-aff-vel-transf-obs-model}), provides direct cues
  to the local surface orientation of the surface patch, according to the
  theoretical analysis in G{\aa}rding and Lindeberg
  (\citeyear{GL94-IJCV}) Section~5.2,
\item
  provided that the affine transformation $A^{(k)}$ in the locally
  linearized perspective projection
  model (\ref{eq-sc-aff-vel-transf-obs-model})
  is normalized such that it constitutes a pure orthographic
  projection, then knowledge about the spatial scaling factor $S_x^{(k)}$ provides
  direct cues to the depth $Z^{(k)} = 1/S_x^{(k)}$,
\item
  knowledge about the 
  affine transformation matrix
  $\tilde{B}^{(k)}$, in the locally linearized projective transformation
  (\ref{eq-sc-aff-vel-transf-alt-obs-model}) between pairwise views,
  provides direct cues to the local surface orientation,
  according to the theoretical analysis in G{\aa}rding and Lindeberg
  (\citeyear{GL94-IJCV}) Section~6.1.
\end{itemize}
In these ways, the parameters of the joint spatio-temporal
transformation models are therefore directly related to the 3-D structure of the
scene, provided that appropriate matching of the positions in image
space for the receptive field
responses can be obtained, to compute the image deformation parameters.

Of particular importance in this context is to really adapt the shapes
of the receptive fields according to the covariance property of
the actual image deformation. In Lindeberg
and G{\aa}rding (\citeyear{LG96-IVC}), it was specifically shown that
such shape-adaptation of the receptive fields can improve the accuracy
of surface orientation estimates by typically an order of magnitude,
compared to not adapting the shapes of the receptive fields to the
actual image deformation,
see Tables~1--4 in Lindeberg and G{\aa}rding (\citeyear{LG96-IVC}).

For more extensive treatments of the topic of deriving cues to 3-D scene
structure by combination of information from multiple views, see
the monographs by
Hartley and Zisserman (\citeyear{HarZis04-Book}) and
Faugeras (\citeyear{Faug-book}) and the references therein.

\section{Summary and conclusions}
\label{sec-summ-disc}

We have presented an in-depth unified theory for covariance
properties and transformation properties
of the spatio-temporal receptive fields according to the
generalized Gaussian model for spatio-temporal receptive fields,
which extends the previous work on this topic to both joint
compositions of multiple types of geometric image transformation, as
well as to the basic types of spatio-temporal differentiation
operators, that occur in the models of the spatio-temporal receptive
fields, including the extension of the covariance and transformation
properties to algebraically much simpler forms in terms of
scale-normalized derivatives.

After first in Section~\ref{sec-gen-gauss-der-model} giving
an overview of the spatial-temporal receptive
model, that we base this work on, as well as its biological relevance,
we have in
Section~\ref{sec-sc-norm-spat-temp-ders} described a
general theoretical foundation for obtaining provable covariance
properties for spatial and temporal scale derivatives at multiple
spatial and temporal scales, by formulating
scale-normalized spatial or temporal derivative operators
over {\em lower-dimensional\/} spatial or temporal domains.

Specifically, we have in
Sections~\ref{sec-aff-sc-norm-dir-ders}--\ref{sec-cov-prop-sc-norm-aff-hess-mat}
both formulated and analyzed a set of new notions of affine scale-normal\-ized
directional derivative operators as well as scale-normal\-ized affine gradient
and affine Hessian operators, to be applied to affine Gaussian scale-space
representations, obtained by convolution with anisotropic affine
Gaussian kernels, and shown that these concepts leads to provable
covariance properties, for the notion affine scale-normalized directional
derivatives with respect to two important subgroups of the group of general
spatial affine transformations, while for the notions of
scale-normalized affine gradients and for the notion of the
scale-normalized affine Hessian matrix, the covariance properties
hold over the full group of non-singular spatial affine transformations.

Then, we have in Section~\ref{sec-individ-cov-props} described extensions of
such transformation properties and covariance properties to 
higher-dimensional {\em joint\/} spatio-temporal receptive field models,
for the four classes of {\em single\/} image transformations, in terms
of either (i)~a pure spatial scaling transformation, (ii)~a pure spatial affine
transformation, (iii)~a pure temporal scaling transformation, or (iv)~a pure
Galilean transformation.

To handle more general geometric configurations, where variabilities
due to different types of image transformations may occur together, we
have then in Section~\ref{sec-joint-cov-props}
derived a set of {\em joint\/} covariance properties for the
{\em composition\/} of a spatial scaling transformation, a spatial affine
transformation, a Galilean transformation and a temporal scaling
transformation, with explicit expressions for how the receptive field
parameters should be transformed under the composed image
transformation, to make it possible to perfectly match the receptive
field responses under convolutions with spatio-temporal receptive
fields according to the generalized Gaussian derivative model.
This analysis has been performed with regard to both the underlying
joint spatio-temporal smoothing transformation and with regard to the
both regular and the scale-normalized spatio-temporal derivative
operators, that are applied to the
output of the pure smoothing transformation, to produce the receptive
field responses for different combinations of spatio-temporal
derivative operators.

Specifically, we have shown that when using the
notion of scale-normalized spatio-temporal derivative operators,
the resulting spatio-temporal derivative responses become essentially
equal, up to a possibly unknown rotation transformation for the
case of affine-extended scale-normalized derivatives,
under the composed spatio-temporal image transformation,
provided that the parameters of the spatio-temporal receptive fields
can be properly matched to the actual form of the geometric
image transformation.

To interpret the class of studied joint covariance properties
geometrically, we have then in Section~\ref{sec-geom-interpret} performed a
geometric analysis of locally linearized projections from the 3+1-D
spatio-temporal world to 2+1-D spatio-temporal image domains,
to interpret the studied class of composed
spatio-temporal image transformations as locally linearized scaled
orthographic projections of a local surface patch, from the tangent
plane of the surface at the fixation point to the image planes for
different viewers, and
also complemented with Galilean motions to represent the possibly
{\em a priori\/} unknown relative motions between the observed object
and the observers, as well as complemented with temporal scaling
transformations, to represent spatio-temporal motion patterns and events that may occur
either faster or slower relative to previous observations of
similarly looking motion patterns or spatio-temporal events.

In this context, we have
also shown how a slight reformulation of that model can be used for
modelling the locally linearized projective transformations between
pairwise views of the same surface patch, including an explicit
derivation of how the corresponding algebra of locally linearized
spatio-temporal transformations will then be closed between different reference
views, with accompanying explicit transformation properties for the
parameters of those locally linearized projection models,
when the reference view is changed between
different visual observations.

For the modified composed spatio-temporal transformation model between
pairwise views of the same local surface patch, we
have also in Section~\ref{sec-cov-props-pairwise-views}
presented explicit expressions for the corresponding joint
spatio-temporal covariance properties, regarding both the underlying
spatio-temporal smoothing transformation as well as its associated
spatio-temporal derivative operators that form the receptive fields.

With regard to biological interpretations of these results, we have
then in Section~\ref{sec-interpret-geom-biol}
described how the degrees of freedom spanned by the free parameters in
the spatio-temporal receptive field model span the same degrees of
freedom as spanned by the free parameters in the locally linearized
scaled orthographic projection model complemented by a local Gali\-lean
motion and a temporal scaling transformation, to account for motion patterns
and spatio-temporal events that may occur either faster or slower relative
to a previous observation of a similarly looking motion pattern or
event.

In view of previously obtained biological modelling results, that the receptive fields of
simple cells in the primary visual cortex can be qualitatively rather
well modelled by idealized receptive fields according to the
theoretical model of visual receptive fields used in this treatment,
we have in this way obtained complementary support for a previously
formulated working hypothesis that the shapes of the receptive fields
found in the primary visual cortex may be regarded as very well
adapted to the structure of the environment.

Finally, we have in Section~\ref{sec-cues-3d-structure} described how
direct cues to the structure of 3-D scenes can be obtained from the parameters
in the locally linearized perspective or projective image formation models
according to
(\ref{eq-sc-aff-vel-transf-obs-model})  and (\ref{eq-sc-aff-vel-transf-alt-obs-model}).

While previous work with the generalized Gaussian derivative model for
spatio-temporal receptive fields have primarily focused on using
either the non-causal 1-D Gaussian kernel or the time-causal limit
kernel for temporal smoothing in the spatio-temporal smoothing
process, the derivations in this paper have been made under a weaker
assumption of only requiring temporal scale covariance
(according to (\ref{eq-temp-sc-cov-temp-kernel})) for the temporal
smoothing kernels. Thus, the results presented in this article do
immediately generalize to the use of other temporal smoothing kernels,
provided that the kernels are covariant under temporal scaling
transformations.


\subsection{Outlook}

In relation to the presented geometric interpretations of the joint
covariance properties in Section~\ref{sec-geom-interpret}, the derived explicit
transformation properties for receptive field responses in
Section~\ref{sec-cov-props-pairwise-views}, defined in
terms of spatio-temporal derivatives of the underlying covariant
spatio-temporal smoothing kernels, do notably show how to both interpret
and relate spatio-temporal receptive field responses,
when viewing dynamic scenes under
different composed geometric viewing conditions.

Specifically, we propose that this theoretical analysis should have direct relevance,
when interpreting the functional properties of biological receptive
fields, both computationally and with regard to how the simple cells
in the primary visual cortex, whose functional properties we here
model with an
idealized axiomatically derived spatio-temporal receptive field
model. From the viewpoint of the here presented theory, in combination
with previous biological modelling results, that demonstrate a
very good qualitative agreement between idealized
receptive field models according to this theory and neurophysiological
recordings of actual biological receptive fields in the primary visual
cortex of higher mammals,
the shapes of these joint spatio-temporal receptive fields can, from
this viewpoint,
be regarded as very well adapted to 
the structural properties of the environment.

The theoretical results derived in this treatment are more generally intended as a
theoretical foundation for computer vision modules, that make use of
populations of spatio-temp\-oral receptive field responses as the first processing
layers in the visual hierarchy, as well as for formulating 
models of biological vision and interpreting the functional properties of
biological vision from a computational viewpoint, as well as with
regard to constraints from the environment, that may strongly influence
the formation of the receptive fields from a combination of learning
and evolution mechanisms over time.

\section*{Acknowledgements}

I would like to thank Jens Pedersen for valuable interactions
concerning an earlier form of joint covariance property of the
spatio-temporal smoothing transformation.

I would also like to thank the Editor and the Reviewer for valuable comments that
improved the presentation.

{\footnotesize
\bibliographystyle{abbrvnat}
\bibliography{bib/defs,bib/tlmac}}

\end{document}